\documentclass[11pt, logo, acopyright, a4paper]{googledeepmind}

\usepackage[authoryear, sort&compress, round]{natbib}

\usepackage{pdfpages}
\usepackage{cleveref}
\usepackage{url}
\usepackage{wrapfig,booktabs}
\usepackage{xcolor}         % colors
\usepackage{tikz}
\usetikzlibrary{patterns}
\usepackage{colortbl}
\usepackage{arydshln}
\usepackage{adjustbox}
\usepackage{multirow}
\usepackage{enumitem}
\usepackage{subfigure}
\usepackage{gensymb}
\usepackage{xspace} % for algorithm names
\usepackage{graphicx}
\usepackage{makecell}  % Include this in the preamble
\usepackage{minitoc}             % For creating the appendix TOC
\usepackage{makeidx}
\usepackage{booktabs}
\usepackage{bbm}
\usepackage{algorithm}
\usepackage{algorithmic}
\usepackage{colortbl}
\usepackage{psfrag}
\usepackage{adjustbox}
\usepackage{wrapfig}
\usepackage{xspace}
\usepackage{amssymb}
\usepackage{multirow}
\usepackage{afterpage}
\usepackage{float}
\usepackage[toc,page,header]{appendix}
\usepackage{minitoc}
\usepackage{placeins} % For FloatBarrier
\usepackage{titletoc} % Required for partial ToCs
\usepackage{subcaption}

\hypersetup{colorlinks=true, linkcolor=blue, citecolor=blue, urlcolor=blue}

\colorlet{darkgreen}{green!65!black}
\colorlet{darkblue}{blue!75!black}
\colorlet{darkred}{red!80!black}
\definecolor{lightblue}{HTML}{0071bc}
\definecolor{lightgreen}{HTML}{39b54a}
\definecolor{manyshot}{HTML}{6969ff}
\definecolor{medshot}{HTML}{f7c600}
\definecolor{fewshot}{HTML}{ff6969}
\definecolor{mypurple}{HTML}{412F8A}
\definecolor{myorange}{HTML}{fc8e62}

\definecolor{deemph}{gray}{0.55}

\definecolor{textgreen}{RGB}{57, 172, 57}
\definecolor{textred}{RGB}{200, 10, 10}
\definecolor{textgray}{RGB}{100, 100, 100}
\definecolor{visiongold}{RGB}{230, 184, 0}
\definecolor{speechpurple}{RGB}{204, 0, 255}
\definecolor{dataprep}{RGB}{38, 189, 128}
\definecolor{modeltraining}{RGB}{38, 189, 128}
\definecolor{backgroundcol}{RGB}{232, 230, 230}
\definecolor{gold}{rgb}{225, 215, 200} % Gold color
\definecolor{navyblue}{RGB}{40, 66, 200} % Navy blue color
\definecolor{orange}{RGB}{255,127,80} % Navy blue color
\definecolor{pink}{RGB}{219,112,147} % Navy blue color

\definecolor{baselinecolor}{gray}{.95}

\usepackage{pifont}% http://ctan.org/pkg/pifont

\usepackage{makecell, tabularx}
\newcolumntype{L}{>{\RaggedRight}X}
\usepackage{siunitx}

\title{An AI Co-Data-Scientist for Prioritizing Candidate Biomarkers from Wearable Sensor Data}

\correspondingauthor{$\{$ybkim, natviv, hamidpalangi, dmcduff$\}$@google.com}

\reportnumber{} % Leave blank if n/a

\renewcommand{\today}
\author[$\dagger$,1,3]{Yubin Kim}
\author[1]{Salman Rahman}
\author[2]{Samuel Schmidgall}
\author[2]{Chunjong Park}
\author[1]{A. Ali Heydari}
\author[1]{Ahmed A. Metwally}
\author[1]{Hong Yu}
\author[1]{Xin Liu}
\author[1]{Xuhai Xu}
\author[1]{Yuzhe Yang}
\author[5]{Hyeonhoon Lee}
\author[3]{Hyewon Jeong}
\author[7]{Kyungho Lim}
\author[6]{MingYu Lu}
\author[8]{Dongjae Lee}
\author[9]{Theodora Pappa}
\author[10]{Hanseul Cho}
\author[1]{Maxwell A. Xu}
\author[1]{Zhihan Zhang}
\author[3]{Cynthia Breazeal}
\author[1]{Tim Althoff}
\author[4]{Petar Sirkovic}
\author[4]{Ivor Rendulic}
\author[1]{Annalisa Pawlosky}
\author[4]{Nicolas Stroppa}
\author[4]{Juraj Gottweis}
\author[2]{Elahe Vedadi}
\author[2]{Alan Karthikesalingam}
\author[2]{Pushmeet Kohli}
\author[1]{Mark Malhotra}
\author[1]{Shwetak Patel}
\author[10]{Samir Tulebaev}
\author[3]{Hae Won Park}
\author[$\dagger$,2]{Vivek Natarajan} 
\author[$\dagger$,1]{Hamid Palangi}
\author[$\dagger$,1]{Daniel McDuff}

\affil[$\dagger$]{Corresponding Authors}
\affil[1]{Google Research}
\affil[2]{Google DeepMind}
\affil[3]{Massachusetts Institute of Technology}
\affil[4]{Google Cloud AI}
\affil[5]{Seoul National University Hospital}
\affil[6]{University of Washington}
\affil[7]{Yonsei University College of Medicine}
\affil[8]{Korea University Guro Hospital}
\affil[9]{Mass General Brigham}
\affil[10]{Brigham and Women's Hospital}

\begin{abstract}
Wearable devices generate continuous physiological and behavioral data, but converting these signals into clinically reviewable biomarker hypotheses remains labor-intensive. We introduce CoDaS, an AI co-data-scientist that integrates multi-agent hypothesis generation, deterministic statistical analysis, adversarial validation and literature-grounded interpretation under human oversight. Across three wearable cohorts comprising 9,279 participant-observations, CoDaS prioritized candidate associations for mental-health and metabolic endpoints after internal checks for replication, stability, robustness and leakage. The system identified related circadian-instability signals associated with depression, including sleep-duration variability in DWB ($\rho = 0.252$, $p < 0.001$) and sleep-onset variability in GLOBEM ($\rho = 0.126$, $p < 0.001$), and derived a wearable cardiovascular-fitness index associated with insulin resistance (steps/resting heart rate; $\rho = -0.374$, $p < 0.001$). Adding these features to demographic models produced modest gains ($\Delta R^{2}=0.040$ for depression, 0.021 for insulin resistance). In a 12-clinician review totaling approximately 25 active hours, clinician validity judgments aligned with CoDaS confidence tiers ($\rho = 0.67$, $p = 0.005$), whereas added clinical value and confidence to act were rated lower. CoDaS supports traceable, hypothesis-generating prioritization of wearable candidate biomarkers.
\end{abstract}

\begin{document}

\maketitle

\newenvironment{Itemize}{
    \begin{itemize}[leftmargin=*]
    \setlength{\itemsep}{0pt}
    \setlength{\topsep}{0pt}
    \setlength{\partopsep}{0pt}
    \setlength{\parskip}{0pt}}
{\end{itemize}}
\setlength{\leftmargini}{9pt}

\section{Introduction}

Consumer wearables, including smartwatches, continuous ECG patches, and temperature sensors provide continuous longitudinal measurements of human physiology beyond traditional clinical environments \citep{dunn2021wearable, daniore2024wearables, brasier2024nextgen}. These devices generate high-dimensional data streams capturing heartbeat dynamics, activity patterns, sleep architecture, and thermoregulation, enabling continuous detection of early physiological deviations preceding clinical presentation \citep{topol2019deep, goldsack2021evaluation, li2026hearts}. However, translating these signals into clinically useful candidate digital measures at scale remains challenging \citep{vasudevan2022convergence, definitionsdigital2024, coravos2019developing, babrak2019traditional}. Existing approaches rely on handcrafted features within narrowly defined disease settings, limiting generalizability across populations, devices, and clinical contexts.

\begin{figure}[t!]
\centering
\includegraphics[width=\textwidth]{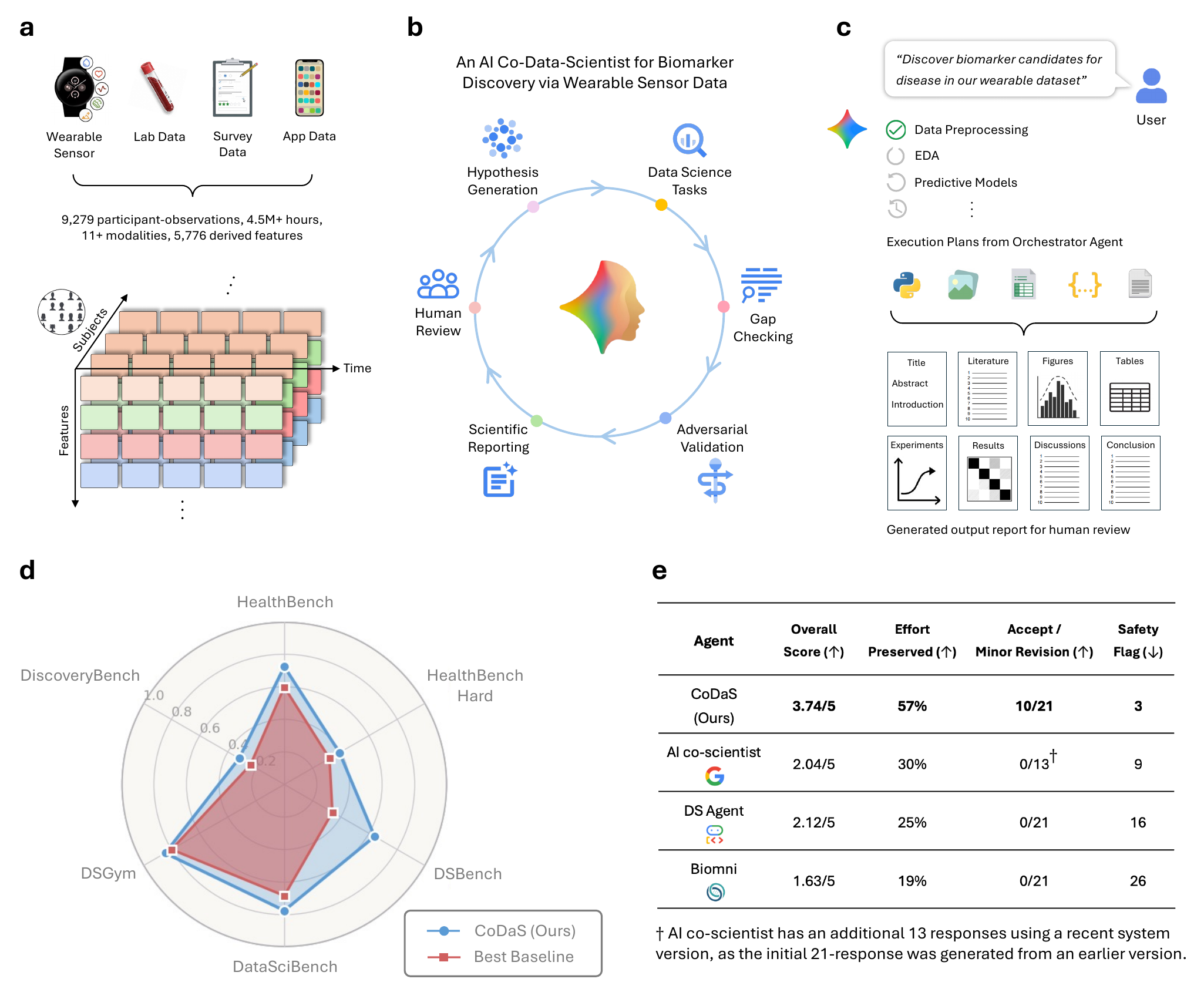}
\centering
\caption{\textbf{Overview.}
\textbf{(a)}~CoDaS takes continuous physiological time series data from wearable sensors, laboratory results, surveys, and third-party applications.
\textbf{(b)}~A closed-loop architecture orchestrates six phases mirroring the biomarker discovery lifecycle, enabling iterative refinement of candidate biomarkers for statistical robustness and physiological interpretability.
\textbf{(c)}~Given a natural-language research goal, CoDaS decomposes the objective into a phased execution plan, and sub-agents carry out each step through iterative analysis and deterministic code execution, producing a report draft with literature reviews, statistical summaries, mechanistic hypotheses, and assessments for human review.
\textbf{(d)}~CoDaS was additionally evaluated on data-science and health benchmarks, achieving competitive performance against per-benchmark strongest baselines.
\textbf{(e)}~In a blinded human expert evaluation ($n = 15$), CoDaS received the highest mean scores across quality dimensions among compared systems.}
\label{fig:teaser}
\end{figure}

Digital biomarkers have shown clinical utility across several domains, from smartphone-based cognitive assessments for neurodegeneration \citep{dagum2018digital} and nocturnal movement signatures for Parkinson's disease \citep{yang2022ai} to activity-derived indices for cardiometabolic risk \citep{guan2024walk, chapman2022impact}. However, these approaches remain confined to single disease contexts, lacking a scalable framework for cross-domain biomarker discovery \citep{fagherazzi2020digital, guan2024walk}.

At the same time, recent advances in large language models (LLMs) have enabled partial automation of scientific workflows, including literature synthesis \citep{lu2024aiscientist, zheng2025automation}, hypothesis generation \citep{zhou2024hypothesis, tong2024automating}, and experimental planning \citep{clusmann2023future}. In clinical settings, LLMs have approached expert-level performance in controlled settings such as in medical question answering \citep{singhal2025expert, nori2023capabilities} and clinical decision support \citep{tu2024generalist, savage2024diagnostic}, with growing applications in safety-aware diagnostics \citep{kim2025vocalagent} and personalized health monitoring \citep{kim2024health, cosentino2024towards, heydari2025anatomy}. Multi-agent frameworks extend these capabilities to autonomous research workflows, with demonstrations in chemistry \citep{boiko2023autonomous}, nanobody design \citep{swanson2025virtual}, single-cell analysis \citep{xiao2024cellagent}, and end-to-end manuscript generation \citep{lu2024aiscientist, lu2024aiscientistv2}. In these systems, agents such as hypothesis generators, statistical analysts, and critics collaborate through shared state to mirror the division of labor in human research teams \citep{gao2024empowering, li2023camel}.

However, the majority of these systems operate in symbolic, textual, or algorithmic domains rather than on real world, high-dimensional physiological time series data with clinical or validated survey endpoints. This creates a practical tension between exploratory capacity and scientific rigor, particularly in high-dimensional physiological data where spurious correlations and feature leakage are prevalent. For candidate biomarkers to translate into clinical impact, it is insufficient to demonstrate statistical significance alone. They must also exhibit physiological interpretability, mechanistic plausibility, and consistency across independent cohorts \citep{coravos2019developing, babrak2019traditional, goldsack2021evaluation}. Single agent or monolithic systems may struggle to balance exploratory breadth with scientific rigor, risking spurious associations or uninterpretable feature composites \citep{rotem2024visual, jiang2024interpretable}. Moreover, while autonomous AI systems excel at pattern recognition, translating discovered patterns into clinically reviewable hypotheses requires integrating domain knowledge, evaluating mechanistic coherence, and maintaining expert oversight throughout the discovery process \citep{daniore2024wearables, vasudevan2022convergence}.

To this end, we introduce \textbf{CoDaS} (AI \underline{Co}-\underline{Da}ta-\underline{S}cientist), a multi-agent system for biomarker discovery from wearable time-series data. The system organizes discovery as an iterative six-phase loop spanning data profiling, hypothesis generation, statistical and machine learning analysis, adversarial validation, mechanistic reasoning, and report synthesis (Figure \ref{fig:teaser}b). Specialized agents operate over shared state to enforce separation between exploration, validation, and critique, reducing the risk of uninterpretable findings (Figure \ref{fig:pipeline_teaser}). Unlike fully autonomous systems that operate in isolation, CoDaS is designed to preserve human oversight where the framework supports optional human feedback on intermediate findings and mechanistic interpretation at predefined checkpoints. 

% In the experiments reported here, however, discovery was run autonomously and human input was restricted to post-discovery review.

We evaluated CoDaS on three large-scale wearable datasets (Figure \ref{fig:teaser}a) spanning mental health and metabolic disease contexts, selected to stress test the pipeline across complementary axes of analytical difficulty: high-frequency multi-modal sensing with rich behavioral signals, sparse longitudinal data with severe missingness, and cross-sectional wearable--clinical linkage requiring mechanistic integration of noninvasive and laboratory features. Specifically, we use the Digital Wellbeing (DWB) dataset (7{,}497 participants; hourly multimodal sensing of sleep, activity, heart rate, and smartphone usage) \citep{mcduff2023research}, the GLOBEM dataset (704 participant-wave observations from 497 unique individuals; longitudinal passive sensing from smartphones and wearables) \citep{xu2022globem}, and the WEAR-ME dataset (1{,}078 participants from an original cohort of 1{,}165; physiological aggregates linked to comprehensive clinical panels) \citep{metwally2026insulin}. Across 9{,}279 total participant-observations, CoDaS generated and prioritized candidate digital biomarkers, including sleep-timing variability signatures and nocturnal digital behaviour indices associated with depression severity, and a wearable-derived cardiovascular fitness index associated with insulin resistance (details in Section \ref{sec:experiments_and_results}). Principal candidates were evaluated with an internal validation battery operationalizing four complementary dimensions: replication, stability, robustness, and discriminative signal via 11 checks (including permutation testing, bootstrap confidence intervals, subgroup consistency, and methodological triangulation), consistent with emerging digital biomarker standards. Effect sizes are modest in magnitude, and all findings should be interpreted as hypothesis-generating signals requiring prospective validation.

\noindent
Our primary contributions are as follows:
\begin{enumerate}
    \item \textbf{Wearable-specific multi-agent pipeline.} We introduce CoDaS, a multi-agent system designed to support biomarker discovery from wearable sensor data. Its six-phase pipeline, spanning data profiling, hypothesis generation, parallel statistical and machine learning exploration, adversarial validation, deep literature research, and structured report generation forms an iterative loop that mirrors the human biomarker discovery lifecycle.

    \item \textbf{Population-scale hypothesis generation across disease domains.} We demonstrate candidate biomarker prioritization across 9{,}279 participant-observations and three datasets, subjecting principal candidates to a structured internal validation battery and using an internal audit that operationalizes held-out confirmation, stability, robustness and discriminative signal via 11 checks. 

    \item \textbf{Human oversight and auditability.} CoDaS operates the discovery phase with human supervision through a feedback module for post-discovery review, interpretation, and optional follow-up guidance.
\end{enumerate}

\section{An AI Co-Data-Scientist for Candidate Biomarker Prioritization from Wearable Sensors}

In this section, we introduce \textbf{CoDaS} (AI Co-Data-Scientist), a multi-agent system that implements a structured, hybrid discovery pipeline. CoDaS structures candidate-biomarker prioritization by systematically exploring, generating, and prioritizing candidate hypotheses at a scale that would be challenging to be achieved manually.

\subsection{System Architecture and Agent Specialization}
CoDaS operates as a set of AI agents coordinated through a shared workflow, using Gemini-3.1 Pro Preview for research-intensive reasoning and code generation and Gemini-3 Flash Preview for repeated lower-latency tasks. To support the full discovery process, CoDaS replaces monolithic reasoning with specialized personas, dedicated tool sets, and distinct mandates within a six-phase workflow. This specialization separates empirical analysis, theoretical grounding, and strategic oversight.

\begin{figure}[t!]
\centering
\includegraphics[width=\textwidth]{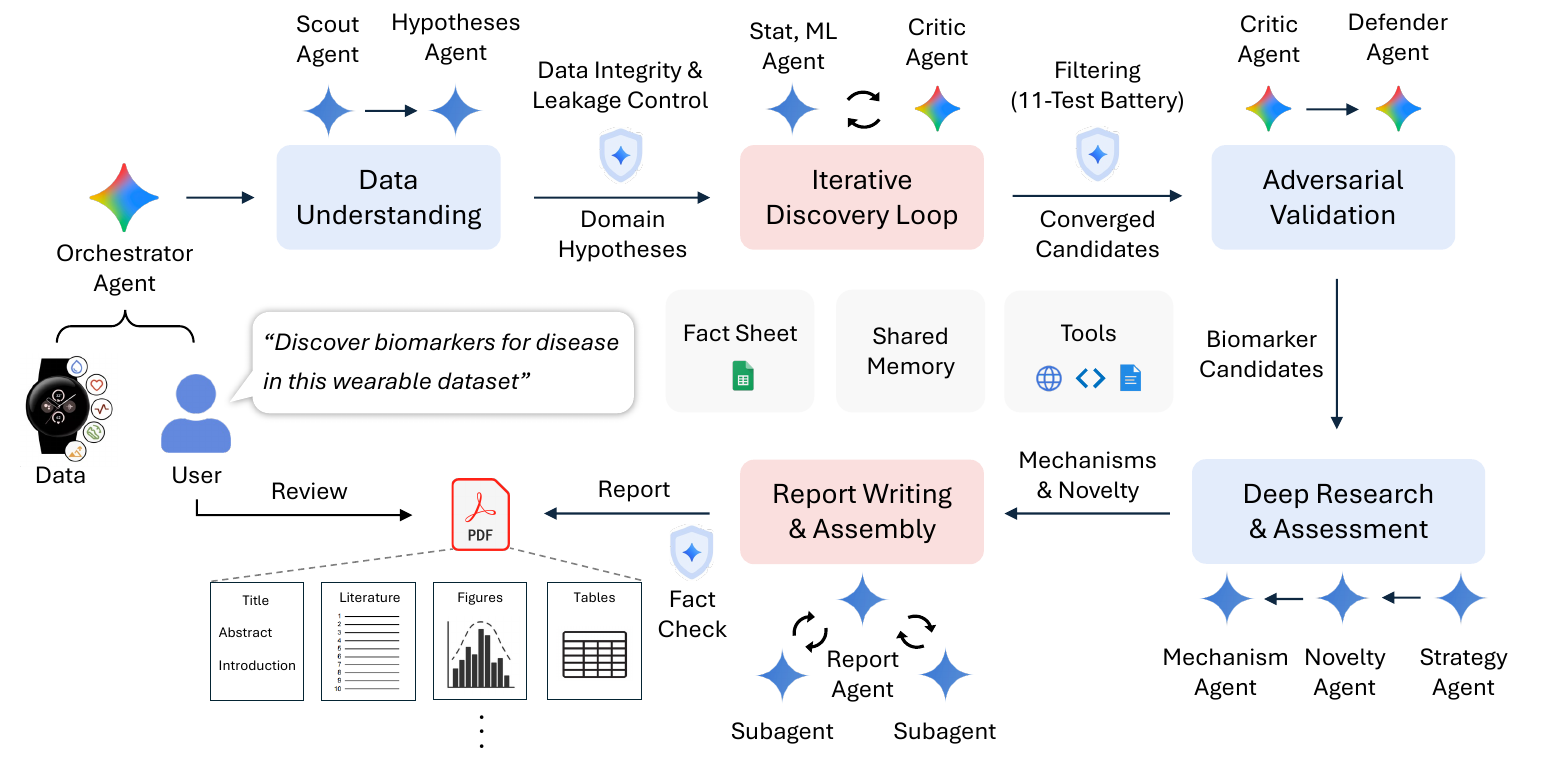}
\centering
\caption{\textbf{CoDaS Architecture.}
Given a research goal and wearable sensor data, an Orchestrator Agent coordinates sequential stages through specialized sub-agents.
\textbf{(1)}~\textit{Data Understanding}: Scout and Hypotheses Agents profile the dataset and generate domain-grounded hypotheses.
\textbf{(2)}~\textit{Iterative Discovery Loop}: Statistical/ML and Critic Agents iteratively refine candidate biomarkers until convergence.
\textbf{(3)}~\textit{Adversarial Validation}: Critic and Defender Agents debate each candidate to filter for statistical robustness.
\textbf{(4)}~\textit{Deep Research \& Assessment}: Mechanism, Novelty, and Strategy Agents evaluate biological plausibility, literature novelty, and translational potential of statistically prioritized candidates.
\textbf{(5)}~\textit{Report Writing \& Assembly}: a Report Agent with dedicated sub-agents compiles a draft manuscript for expert review.
All agents share a memory, fact sheet and tool sets. CoDaS enforces safety mechanisms to support statistical validity and reduce the risk of spurious reporting.
A \emph{leakage-prevention} module separates feature construction from target signals, while candidate associations are filtered through statistical tests with FDR correction and assigned explicit reporting labels, including screened, conditional, exploratory, rejected or unstable.
An \emph{adversarial validation step} further audits each candidate for target leakage, overfitting, construct overlap, confounding sensitivity and physiological implausibility. A detailed architectural diagram is provided in Figure \ref{fig:architecture}.}
\label{fig:pipeline_teaser}
\end{figure}

\paragraph{Researcher Ensemble.}
The Researcher ensemble compiles a structured summary of the relevant clinical literature to guide subsequent biomarker discovery.
\begin{itemize}[leftmargin=*, itemsep=2pt]
    \item \textbf{Inputs:} A high level research query (e.g., ``discover predictive signatures for depression severity'') and access to scientific literature databases.
    \item \textbf{Process:} The ensemble utilizes a set of sub-agents. A \textit{Literature searcher} executes targeted queries to retrieve a corpus of relevant scientific abstracts and full text articles. Simultaneously, a \textit{BibTeX Validator} and a \textit{Paper Verifier} deterministically crosscheck retrieved claims to reduce hallucination. Finally, specialized \textit{Novelty} and \textit{Mechanism} agents summarize candidate mechanistic links reported in the literature.
    \item \textbf{Outputs:} A structured biological prior containing: (1) a list of established biomarkers and their reported physiological pathways; (2) assessments determining the true novelty of generated candidates; and (3) mechanistic rationales linking the data-driven findings to plausible biological pathways.
\end{itemize}

% \begin{figure*}[htb]
% \centering
% \includegraphics[width=0.95\textwidth]{imgs/dsa_teaser.pdf}
% \caption{\textbf{Data Science Engine.}
% The Data Science engine automates computational analysis through a hybrid architecture coupling deterministic Python execution runners with language model interpreters. A \textit{Scout} sets the clinical target and dataset taxonomy. \textit{Statistical Runners} and \textit{Machine Learning Runners} execute code in a sandboxed environment to compute significance, effect sizes, and predictive power. Intermediate results return to explicitly paired language model interpreters that translate statistical outputs into scientific insights. This dual track exploration continues iteratively, allowing the system to optimize parameter selection and refine analytical models based on empirical ground truth.}
% \label{fig:dsa}
% \end{figure*}

\paragraph{Data Science Engine.}
The Data Science engine represents the analytical core of CoDaS, responsible for all direct interactions with the wearable sensor data. It abandons purely generated code in favor of a hybrid deterministic and generative approach.
\begin{itemize}[leftmargin=*, itemsep=2pt]
    \item \textbf{Inputs:} Raw, high-dimensional time series datasets intersecting with the structured knowledge base provided by the \emph{Researcher} ensemble.
    \item \textbf{Process:} The engine comprises paired deterministic code runners and language model interpreters. A \textit{Scout} agent first maps the dataset schema, defining the clinical target variable. A \textit{DataLoader} and \textit{Exploratory Data Analysis (EDA)} runner then profile the data structure. Following this, \textit{Statistical runners} implement robust univariate testing (e.g., correlation, effect size), while \textit{Machine Learning runners} execute multivariate predictive modeling with cross validation (e.g., Ridge regression, ensemble trees) in parallel. Paired interpreters parse the respective outputs to form coherent analytical narratives.
    \item \textbf{Outputs:} A comprehensive suite of empirical results including: (1) ranked lists of novel biomarker candidates (ordered by composite effect size, validation pass rate, and clinical plausibility); (2) robust statistical metrics including significance levels adjusted for multiple comparisons; (3) out-of-sample predictive performance metrics; and (4) generated correlation heatmaps and feature importance distributions.
\end{itemize}

\paragraph{Orchestrator and Critical Evaluators.}
The Orchestrator coordinates the CoDaS pipeline, providing strategic direction, managing state transitions across the six phases, and ensuring rigorous internal review.
\begin{itemize}[leftmargin=*, itemsep=2pt]
    \item \textbf{Inputs:} The overarching research objective, real time outputs from specialist agents, and periodic interactive feedback from human domain experts.
    \item \textbf{Process:} The Orchestrator manages the pipeline trajectory. It implements a \textit{GapChecker} to identify unresolved questions following iterative empirical analysis, deciding whether to pursue deeper feature engineering or move to validation. 
    Crucially, it initiates an adversarial debate phase involving \textit{Critic} and \textit{Defender} agents, which actively stress test the proposed biomarkers for confounding variables, statistical leakage, and physiological implausibility before finalization. This hierarchical oversight structure aligns with recent paradigms emphasizing tiered agentic architectures to ensure AI safety and rigorous validation in clinical contexts \citep{kim2025tiered}. A worked Critic and Defender exchange, covering the rejection of a leakage feature and a redundant proxy and the handling of a confounded behavioral candidate, is provided in Section~\ref{sec:ablation}.
    \item \textbf{Outputs:} The coordinated sequence of agent invocations, automated manuscript generation via paired \textit{Writer} and \textit{Reviewer} agents, and structured interactive prompts requesting domain expert feedback at critical junctures.
\end{itemize}

\subsection{The Hybrid Discovery Pipeline}
CoDaS implements a structured six-phase pipeline that narrows the search space from thousands of raw sensor permutations to a curated set of statistically prioritized, mechanistically grounded biomarker candidates (see Figure \ref{fig:pipeline_teaser} for an overview and Figure \ref{fig:architecture} for the full details).

\paragraph{Phase A: Automated Data Profiling and Literature Grounding}
The objective of the initial phase is to build both a conceptual and empirical map of the clinical task. 
\begin{itemize}[leftmargin=*, itemsep=2pt]
    \item \textbf{Empirical Contextualization:} Deterministic loaders and the EDA runner survey the raw data. The Scout agent synthesizes these statistical profiles to establish an analytical baseline, understanding data sparsity, longitudinal variance, and demographic distributions.
    \item \textbf{Biological Anchoring:} The Orchestrator tasks the Researcher ensemble to construct the theoretical foundation. By identifying established clinical predictors, CoDaS seeds the search space with literature-derived priors, encouraging subsequent exploration to remain anchored to biological plausibility rather than spurious correlations.
\end{itemize}

\paragraph{Phases B \& C: Parallel Agentic Search and Adversarial Validation}
The core discovery engine actively synthesizes raw features into composite physiological parameters over iterative loops.
\begin{itemize}[leftmargin=*, itemsep=2pt]
    \item \textbf{Dual Track Parallel Exploration:} To balance interpretability with maximal predictive power, the Orchestrator runs parallel statistical and machine learning iterations. Generative interpreters propose physiologically rational transformations (e.g., standard deviations of resting heart rates, or ratios between activity profiles), which deterministic runners immediately evaluate. A GapChecker module monitors marginal gains in model performance, candidate novelty, and validation yield, and uses these signals to decide whether further refinement is warranted.
    \item \textbf{Adversarial Stress Testing:} Candidate biomarkers prioritized for Phase C undergo an additional review stage designed to identify potential sources of spurious association. A Critic agent evaluates alternative explanations, including statistical artifacts, data leakage, confounding and inconsistencies with the literature, whereas a Defender agent assesses whether the available empirical evidence supports retention of the candidate. Biomarkers that cannot be supported following this review are removed from further consideration. A representative example illustrating how this process identifies leakage, redundant proxies and confounding-prone candidates is provided in Section \ref{sec:ablation}.
\end{itemize}

\paragraph{Phases D, E, \& F: Mechanistic Reasoning and Automated Reporting}
The concluding phases reconstruct the surviving empirical signals into a structured draft manuscript.
\begin{itemize}[leftmargin=*, itemsep=2pt]
    \item \textbf{Novelty and Mechanism Extraction (Phase D):} The Researcher ensemble executes deep secondary literature sweeps focused exclusively on the surviving biomarker candidates. It formally assesses literature novelty and formulates hypothesis-generating physiological rationales linking the digital signature to plausible pathways described in the literature.
    \item \textbf{Drafting and Interactive Feedback (Phases E \& F):} Writer and Reviewer agents format the findings, encompassing statistical summaries, machine learning benchmarks, and mechanistic rationales, into structured draft reports suitable for expert review. The system then enters the final collaborative stage, presenting the comprehensive draft to the human expert. 
\end{itemize}

\subsection{Coordination and Interactive Feedback}
The Orchestrator agent facilitates communication through a shared state, allowing downstream agents to access and build upon preceding qualitative evaluations. Most importantly, the framework embraces collaborative discovery with human practitioners. The Orchestrator is programmed to identify complex scenarios demanding biological intuition and immediately solicit expert input. These interactive triggers include:
\begin{itemize}
    \item \textbf{Plausibility Gaps:} A generated biomarker achieves higher-than-expected predictive performance but lacks a clear analogue in the retrieved literature. The Orchestrator pauses progression, querying the medical professional for interpretative guidance.
    \item \textbf{Conflicting Evidence Paradigm:} Substantial data-driven findings fundamentally contradict established paradigms identified by the Researcher ensemble.
    \item \textbf{Search Stagnation:} The generative exploration fails to surpass predefined performance baselines over sequential iterations, prompting the expert to explicitly suggest customized feature transformations.
\end{itemize}
This mechanism is intended to preserve expert interpretive authority while allowing the system to surface candidate findings more efficiently.

\subsection{Multi-axis candidate assessment}
Optimizing solely for machine learning accuracy frequently yields noninterpretable blackbox algorithms unsuited for medical application. To mitigate this, the Orchestrator evaluates candidate features across a multidimensional framework spanning both quantitative validity and qualitative clinical assessment.

\begin{enumerate}
    \item \textbf{Statistical Validity:} Evaluates signal strength across robust metrics including correlation effect sizes, multiple hypothesis adjusted significance levels, and out-of-sample predictive performance using extensive cross validation procedures.
    \item \textbf{Clinical Plausibility:} Ensures alignment with biological plausibility. Clinical plausibility is assessed by asking the mechanism agents to link each candidate feature to a plausible physiological pathway supported by retrieved primary literature.
    \item \textbf{Originality:} Quantifies the conceptual distance between the proposed biomarker and established indicators curated during the initial literature seeding phase, surfacing potentially novel candidate metrics.
    \item \textbf{Generalizability:} Assesses robustness by running available subgroup analyses, examining biomarker performance consistency across differing demographic cohorts, sensor platforms, or distinct disease severity stratifications.
    \item \textbf{Interpretability:} Penalizes extreme mathematical complexity. The system preferentially weights intuitive, physiologically meaningful composites (e.g., activity recovery gradients or nocturnal variance indices) over highly abstract, untethered neural embeddings.
    \item \textbf{Clinical Utility:} Separates candidate discovery from clinical actionability. Candidates are later assessed by clinicians for practical measurability, added value over existing biomarkers, likelihood of influencing advice and confidence to act. Candidates with low actionability are retained as hypotheses for prospective evaluation.
\end{enumerate}

These axes deliberately separate statistical validity from clinical value, so that a candidate can be statistically significant, biologically plausible and novel yet still rank low when its incremental clinical utility is weak. Measured confounding is partially assessed within the validity axis through a causal-robustness check, which residualizes each candidate against demographic covariates and previously validated biomarkers and requires the association to survive partial-correlation control (Section~\ref{sec:validation_integrity}). This guards against candidates whose association is carried by an obvious common cause, for example physical activity that influences both step count and mood. Because the data are observational, this control cannot exclude unmeasured confounding, so a high multi-axis score marks a candidate as worth confounder-controlled prospective study rather than as an established causal biomarker.

\subsection{Data Integrity and Leakage Guardrail}
\label{sec:leakage_firewall}
Information leakage and confounding are critical threats to automated biomarker discovery, where the combinatorial feature space can inadvertently include transformations of the outcome variable. CoDaS enforces the following procedural safeguards throughout the pipeline, designed to ensure that reviewers can audit the boundary between discovery inputs and evaluation targets.

\begin{enumerate}
    \item \textbf{Raw variable exclusion.} The target variable (e.g., PHQ-8, HOMA-IR) and its direct clinical proxies (e.g., fasting glucose and fasting insulin for HOMA-IR, BDI-II for PHQ-4) are excluded from the candidate feature pool at data loading time, before any feature engineering begins.
    \item \textbf{Transformation prohibition.} Monotonic transformations of excluded variables (squared, log, rank-transformed) are detected by the Critic agent's construct overlap analysis, which computes the Spearman correlation of every candidate against all excluded variables. Features exceeding $|\rho| > 0.85$ with any excluded variable are flagged and removed from the candidate pool. We adopt 0.85 as the operational threshold for ``very strong'' monotonic association in biomedical research; in practice, tautological transformations (e.g., \texttt{glucose\_sq}) typically exhibit $|\rho| > 0.95$ with the target, and no candidate in any cohort fell in the 0.80--0.90 range with an excluded variable, so moderate variations in this threshold would not alter the reported results.
    \item \textbf{Discovery-evaluation separation.} All predictive performance is reported exclusively via 5-fold cross-validation with stratified participant-level splits. No participant appears in both training and validation folds within any round. Hyperparameter selection occurs within each training fold only.
    \item \textbf{Label isolation.} Target labels are never exposed to the hypothesis generation, feature engineering, or literature grounding agents. Labels are loaded only by the deterministic statistical and ML runners, which execute in isolated subprocesses. The LLM-based agents observe only summary statistics (e.g., correlation direction, $p$-value ranges) returned by these runners, not the underlying label data.
    \item \textbf{Human review boundary.} Human feedback during the interactive phase is restricted to mechanistic interpretation, plausibility assessment, and high-level analytical guidance (e.g., suggesting domain-informed feature transformations when the pipeline stagnates). Crucially, domain experts do not have access to raw data, model performance metrics, or fold-level predictions during feedback sessions, preventing target-informed feature selection.
    \item \textbf{Construct overlap gating.} Every candidate surviving statistical screening undergoes a construct overlap test measuring its independence from existing validated clinical instruments and from other candidates. Features with high intra-cluster correlation ($|\rho| > 0.85$) are reported but not double-counted in the final validated set.
\end{enumerate}

% The effectiveness of this firewall is demonstrated empirically: disabling the construct overlap analysis in an ablation study permitted tautological features (e.g., \texttt{glucose\_sq}, a monotonic proxy of the HOMA-IR target) to pass 10 of 11 validation tests, inflating the CV $R^2$ from 0.551 to 0.963. The full pipeline correctly identified and rejected this candidate.
% A complete feature exclusion registry listing all excluded variables per dataset is provided in the Appendix (Table~\ref{tab:feature_exclusion}).

\subsection{Statistical Internal Validation Battery and Reporting Integrity}
\label{sec:validation_integrity}

Autonomous AI systems operating in biomedical research face inherent risks of hallucination, spurious discovery, and unverified reporting~\citep{tang2025risks, luo2025more, cornelio2025need, zhu2025safescientist}.
CoDaS addresses these through two complementary mechanisms: a multi-stage statistical internal validation battery that every candidate must pass, and a deterministic reporting framework that reduces LLM-generated prose from diverging from empirical results.

\paragraph{Internal Validation battery: four dimensions, eleven checks.}
Every candidate biomarker that passes the statistical filtering stage must survive an internal validation battery organized around four complementary dimensions: replication, stability, robustness, and discriminative power, operationalized via 11 checks executed in a deterministic subprocess isolated from the language model agents.
None of these checks were preregistered; they are components of the pipeline design, not a prespecified analysis plan. The checks are not independent tests; several share underlying data (e.g., replication and bootstrap both operate on the same sample; subgroup consistency and robustness both use demographic variables), and should be interpreted as a structured post-hoc audit rather than 11 orthogonal significance tests. All results from this battery are hypothesis-generating and do not replace prospective external validation.

\begin{enumerate}[leftmargin=*, itemsep=2pt]
    \item \textbf{Held-out confirmation.} Spearman correlation on a held-out confirmation set (distinct participant-level split, $N \geq 20$), verifying that the effect shows directionally consistent association in a held-out split. For cohorts with repeated measures (e.g., GLOBEM), the confirmation set retains only one randomly selected observation per participant so that the test statistic is computed on fully independent rows.
    \item \textbf{Permutation test.} Empirical null distribution from 1{,}000 label-permuted resamples; guards against inflated significance due to distributional properties of the feature.
    \item \textbf{Bootstrap stability.} 1{,}000 bootstrap resamples the validation set with replacement providing 95\% confidence intervals for the association estimate; rejects candidates whose CI straddles zero, indicating directional instability.
    \item \textbf{Leave-one-out influence.} Rejects any candidate whose association sign flips when any single participant is excluded, indicating sensitivity to outliers.
    \item \textbf{Subgroup consistency.} Requires the association to hold within each half of the cohort (class split for classification; median split for regression), protecting against Simpson's paradox.
    \item \textbf{Method triangulation.} Recomputes the association using Pearson and Kendall's $\tau$; the candidate must remain significant across all applicable methods, guarding against method-specific artifacts.
    \item \textbf{Construct validity hard gate.} Rejects candidates with implausibly strong correlations ($|\rho| > 0.85$ for $N > 30$; adaptive thresholds for smaller samples), which typically indicate undetected tautological transformations.
    \item \textbf{Confounding-sensitivity check.} Residualizes the candidate against demographic confounders and previously validated biomarkers; the association must survive partial-correlation control.
    \item \textbf{Construct independence hard gate.} Detects derived features whose components correlate strongly with the target, classifying each candidate as independent, proxy, or compositional. Proxy and compositional candidates are rejected or flagged for disclosure.
    \item \textbf{CI consistency hard gate.} Verifies that the point estimate and bootstrap CI midpoint agree in sign; directional inconsistency indicates numerical instability.
    \item \textbf{Discriminative signal check.} For classification analyses, we report AUC as a descriptive measure of above-chance discrimination. We use AUC $\geq 0.55$ only as a permissive internal screen to identify whether a candidate carries non-random information; it is not a threshold for clinical utility, diagnostic use or deployment readiness.
\end{enumerate}

If all three of Tests 1 through 3 fail simultaneously, the candidate is immediately rejected.
Candidates passing at least 70\% of applicable checks, with all core tests (replication, permutation, bootstrap, and CI consistency) passed, are labeled \textsc{screened}; candidates passing 40--70\% of checks, or downgraded from screened status due to marginal effect sizes, are labeled \textsc{conditional}; the remainder are \textsc{rejected}. Three hard gates (construct validity, construct independence, CI consistency) trigger automatic rejection regardless of overall pass rate.

All verdicts, evidence summaries, and per-test results are persisted in pipeline state and injected into the manuscript so that reported validation statistics derive from deterministic computation rather than LLM-generated prose.

\paragraph{Fact Sheet.}
A known vulnerability of LLM-based scientific writing is hallucination of numerical values, including sample sizes, effect sizes, and validation counts~\citep{young2025harnesses}.
CoDaS mitigates this through a \emph{Fact Sheet}: a flat key-value dictionary of every reportable number computed deterministically from pipeline state before any section writer is invoked.
The Fact Sheet captures sample sizes, demographic distributions, model performance ($R^2$, AUC), validation counts, feature counts, discovery round tallies, and construct exclusion summaries.
All section-writing agents receive the Fact Sheet as a structured context attachment and are instructed to copy values verbatim rather than infer them from narrative context.

\paragraph{Numeric verification and consistency enforcement.}
After each section is drafted, a dedicated numeric verification pass applies pattern-based correction to detect and fix common hallucination targets: sample size claims, prioritized candidate counts, validation test counts, method counts, and feature counts.
Corrections are applied only when the LLM-written value falls within a tolerance of the known ground truth, limiting false positives.
All corrections are logged to a per-run audit file for transparency.
A final consistency check cross-references every section against the Fact Sheet before LaTeX compilation.

\paragraph{Quality gates for output suppression.}
CoDaS applies five deterministic quality gates before report assembly: (i)~a multicollinearity gate suppresses OLS tables when the variance inflation factor exceeds 50; (ii)~a performance gate suppresses ML result tables when the best cross-validated AUC falls below 0.55 or $R^2$ falls below 0; (iii)~an overfitting gate suppresses results when the train-to-CV ratio exceeds 5; (iv)~an ablation gate suppresses feature importance tables when all models perform at chance; and (v)~a forest plot deduplication gate limits any single feature family to two representatives, preventing visual dominance by correlated variants.

Together, these mechanisms align with emerging best practices for long-running autonomous scientific agents~\citep{young2025harnesses, lu2026towards, wu2026towards} and address the integrity risks identified for AI scientist systems operating in high-stakes biomedical domains~\citep{tang2025risks, zhu2025safescientist}.

\begin{table}[htbp]
\centering
\renewcommand{\arraystretch}{0.8}
\caption{\textbf{Cohort characteristics.}
Participant demographics, data modalities, and endpoint definitions across the three evaluation cohorts.
Abbreviations: DWB, Digital Wellbeing; GLOBEM, Generalization of Longitudinal Behavior Modeling; WEAR-ME, Wearables for Metabolic Health; SD, standard deviation; BMI, body mass index; RHR, resting heart rate; HRV, heart rate variability; GPS, Global Positioning System; PHQ, Patient Health Questionnaire; GAD-7, Generalized Anxiety Disorder-7; PSS, Perceived Stress Scale; PROMIS, Patient-Reported Outcomes Measurement Information System; BDI-II, Beck Depression Inventory-II; BFI-10, Big Five Inventory-10; ERQ, Emotion Regulation Questionnaire; PANAS, Positive and Negative Affect Schedule; CMP, comprehensive metabolic panel; CBC, complete blood count; CRP, C-reactive protein; GGT, gamma-glutamyl transferase; HbA1c, hemoglobin A1c; HOMA-IR, homeostasis model assessment of insulin resistance; RAPIDS, Reproducible Analysis Pipeline for Data Streams.}
\label{tab:cohort_characteristics}
\scriptsize
\begin{tabular}{
p{3.9cm}
>{\centering\arraybackslash}p{3.5cm}
>{\centering\arraybackslash}p{3.5cm}
>{\centering\arraybackslash}p{3.5cm}
}
\toprule
\textbf{Variable} & \textbf{DWB (N\,=\,7{,}497)} & \textbf{GLOBEM (N\,=\,704\textsuperscript{c})} & \textbf{WEAR-ME (N\,=\,1{,}078\textsuperscript{d})} \\
\midrule

% --- Age ---
Age (mean $\pm$ SD) & 43.9 $\pm$ 12.7 & 19.2 $\pm$ 1.4 & 46.9 $\pm$ 12.5 \\[4pt]

% --- Sex ---
Sex (\%) & & & \\
\quad Female & 70.0 & 58.8 & 54.4 \\
\quad Male & 26.5 & 40.2 & 43.6 \\
\quad Other/Not reported & 3.5 & 1.0 & 2.0 \\[4pt]

% --- Race/Ethnicity ---
Race/Ethnicity (\%)\textsuperscript{a} & & & \\
\quad White/Caucasian & 84.9 & 36.4 & ---\textsuperscript{e} \\
\quad Asian & 2.8 & 57.4 & --- \\
\quad Black/African American & 4.0 & 3.3 & --- \\
\quad Hispanic & 7.6 & 7.5 & --- \\
\quad Other/Multiracial & 5.3 & 11.2 & --- \\
\quad Not reported & 0.6 & --- & --- \\[4pt]

% --- BMI ---
BMI (mean $\pm$ SD) & --- & --- & 29.2 $\pm$ 6.7 \\[4pt]

% --- Device type ---
Device type & Fitbit + smartphone & Fitbit + smartphone & Fitbit / Pixel Watch \\[4pt]

% --- Available data modalities ---
Available modalities & & & \\
\quad Wearable signals & Steps, RHR, sleep architecture & Fitbit steps, Fitbit sleep & RHR, HRV, steps, sleep duration, active zone minutes \\
\quad Smartphone logs & Screen time, unlocks, app usage, activity recognition, GPS mobility & Bluetooth proximity, GPS location, phone calls, screen events, WiFi connectivity & --- \\
\quad Clinical screeners & PHQ-8, GAD-7, PSS, PROMIS Sleep & PHQ-4, BDI-II & --- \\
\quad Fasting lab panels & --- & --- & Lipid panel, CMP, CBC with differential, CRP, insulin, GGT, testosterone, HbA1c \\
\quad Survey instruments & Big Five personality, sleep quality & BFI-10, BDI-II, ERQ, PHQ-4, PSS-4, PANAS & Demographics, health history \\[4pt]

% --- Feature count ---
Feature count & 197 & 5{,}508 & 71 \\[4pt]

% --- Monitoring duration ---
Monitoring duration & 26.5 $\pm$ 4.7 days & 77.5 $\pm$ 8.9 days & Cross-sectional \\[4pt]

% --- Missingness ---
Missingness (\%) & 3.0 & 54.6\textsuperscript{b} & 0.1 \\[4pt]

% --- Clinical endpoint ---
Target endpoint & PHQ-8 (0--24) & PHQ-4 (0--12) & HOMA-IR (continuous); HbA1c-based diabetes status \\

\bottomrule
\end{tabular}

\vspace{4pt}
\raggedright
\footnotesize
\textsuperscript{a}Hispanic/Latino ethnicity was collected separately in DWB and GLOBEM; race/ethnicity percentages may therefore sum to $>$100\%. 
\textsuperscript{b}GLOBEM missingness reflects sparse passive sensing features; features with $>$70\% missingness were removed and remaining values were median-imputed within sensing wave. PHQ-4 coverage was 65.0\%. 
\textsuperscript{c}GLOBEM includes 704 participant-wave observations from 497 individuals across four annual cohorts \citep{xu2022globem}. 
\textsuperscript{d}WEAR-ME includes 1{,}078 analytic participants after excluding 87 with incomplete wearable feature coverage from the original cohort of 1{,}165 \citep{metwally2026insulin}. 
\textsuperscript{e}WEAR-ME race/ethnicity variables were not available in the participant-level analytic dataset; source-study distributions are reported in the original cohort publication. 
---\, indicates not collected or not available.
\end{table}

\section{Data}

We assembled three complementary clinical datasets spanning mental health and metabolic disease domains, collectively representing 9{,}279 participant-observations (9{,}072 unique individuals) with continuous physiological monitoring. Each dataset provides distinct advantages for wearable-based biomarker discovery: comprehensive high-frequency longitudinal sensing for depression severity, multiwave passive behavioral phenotyping, and precisely synchronized biometric monitoring linked to robust clinical blood panels. Together, these datasets enable comprehensive evaluation of our autonomous data science framework across diverse disease mechanisms, demographic distributions, and temporal scales. Cohort demographics, modality coverage, feature counts, missingness, and endpoint definitions are summarized in Table \ref{tab:cohort_characteristics}.

\subsection{Digital Wellbeing (DWB) Cohort}

To investigate continuous physiological indicators of depression and anxiety, we analyzed data from the Digital Wellbeing study, a prospective observational cohort of 7{,}500 individuals from the United States \citep{mcduff2024google}. Three participants were excluded owing to incomplete baseline assessments, yielding a final analytic sample of 7{,}497. The study was designed to investigate patterns and relationships between digital device use, continuous physiological signals, and self reported measures of mental health over a four week tracking period. Review and approval for participant enrollment were granted by the Institutional Review Board of the University of Oregon (MOD00000379), and all participants provided informed consent prior to data collection.

The recruitment protocol intentionally targeted demographic diversity. The enrollment stratified participants comprehensively across race, ethnicity (Caucasian, African American, Asian, Latino, Indigenous Populations), sex, and age distributions (18 to 40, and over 40). Data completeness was incentivized via a raffle structure requiring a minimum threshold of seven days of daily status assessments alongside baseline and post study questionnaires.

Participants completed a comprehensive battery of validated psychiatric questionnaires. Baseline assessments included the Patient Health Questionnaire 8 \citep{kroenke2009phq} for depression screening, the Generalized Anxiety Disorder Scale \citep{lowe2008validation}, the Patient Reported Outcomes Measurement Information System Sleep Disturbance subscale \citep{cella2010patient,yu2012development}, the Smartphone Addiction Scale \citep{kwon2013smartphone}, and the Perceived Stress Scale \citep{cohen1983global}. 

The final analyzed cohort for our pipeline comprised 7{,}497 participants contributing 4.55 million hourly records of continuous physiological sensing. The recorded modalities included sleep architecture, step counts, resting heart rate, and smartphone application usage statistics. This temporal resolution supports the extraction of circadian variability indices and nocturnal behavioral markers relevant to affective disorders.

\subsection{Generalization of Longitudinal Behavior Modeling (GLOBEM) Cohort}

To evaluate early behavioral modeling for depression detection across independent temporal cohorts, we utilized the Generalization of Longitudinal Behavior Modeling dataset \citep{xu2022globem}. This dataset aggregates multiwave passive sensing data collected from undergraduate students at the University of Washington via smartphone interactions and wrist worn commercial fitness trackers (Fitbit Flex2 and Inspire2), spanning four consecutive annual cohorts (2018--2021). The study was approved by the University of Washington Institutional Review Board.

The analyzed cohort comprised 704 participant-wave observations from 497 unique individuals (mean age 19.2 $\pm$ 1.4 years; 58.8\% female), with some participants contributing data across multiple annual waves. Rather than isolated cross-sectional measurements, this dataset documents transitions in mental health status over extended academic periods. Continuous modalities include Bluetooth proximity logs, location mobility metrics, sleep duration variance, and daily aggregated activity counts.

Ground truth psychiatric status was established through periodic clinical surveys administered throughout the sensing waves, including weekly PHQ-4 ecological momentary assessments and end-of-term BDI-II administrations. By testing our autonomous discovery pipeline on this cohort, we evaluated reproducible behavioral signatures, such as diminished geospatial mobility and highly variable sleep architecture, that persist across multiple annual cohorts and sensor hardware transitions.

\subsection{Wearables for Metabolic Health (WEAR-ME) Cohort}

To assess the capacity of consumer wearable devices to capture subclinical metabolic dysregulation, we analyzed data from the Wearables for Metabolic Health study, a prospective observational cohort of 1{,}165 participants recruited remotely across the United States via the Google Health Studies application \citep{metwally2026insulin}. The study was approved by Advarra IRB (Protocol Pro00074093). The primary objective evaluates the feasibility of translating continuous data from standard fitness trackers and smartwatches into composite algorithms correlating strongly with rigorous metabolic assays.

After excluding 87 participants with incomplete wearable feature coverage during preprocessing, the final analytic sample comprised 1{,}078 participants (mean age 46.9 $\pm$ 12.5 years; 54.4\% female; mean body mass index 29.2 $\pm$ 6.7). Adherence was defined as a minimum of 14 days of continuous wearable data paired with a comprehensive clinical blood draw and completed demographic surveys. Participants wore Fitbit or Google Pixel Watch devices capturing high resolution heart rate, heart rate variability, step counts, and sleep stages. To reduce circadian variability in laboratory measurements, participants underwent fasting laboratory tests (minimum of 8 fasting hours) early in the morning (7--10\,am) at Quest Diagnostics centers to stabilize circadian blood markers.

The resulting clinical ground truth includes a panel of metabolic readouts: homeostasis model assessment of insulin resistance, fasting insulin, HbA1c, comprehensive metabolic panels (fasting glucose, creatinine, electrolytes), lipid profiles (total cholesterol, high density lipoproteins, triglycerides), high sensitivity C reactive protein, and advanced hematological indices. Demographic endpoints encompassing waist circumference, blood pressure, undiagnosed metabolic syndrome classifications, and body mass index distributions were verified via structured clinical reporting.

By anchoring high-frequency, continuously sampled physiological streams such as resting heart rate and derived cardiovascular fitness indices against these fasting laboratory measurements, this dataset provides a useful testbed for algorithmic biomarker discovery. This linkage permits our framework to explicitly target prediabetic transitions and insulin resistance using noninvasive, passively collected wearable-derived signatures evaluated against fasting laboratory endpoints.

\subsection{Endpoint Specification and Statistical Power}

\paragraph{Endpoint prespecification.} PHQ-8 total score was pre-specified as the primary endpoint for the DWB cohort before CoDaS pipeline execution. HOMA-IR (continuous) was pre-specified for WEAR-ME. For GLOBEM, PHQ-4 was selected by the Scout agent from available clinical instruments as the endpoint with the greatest target coverage (65.0\%; the only alternative, BDI-II, had substantially lower coverage as it was administered only at end-of-term rather than weekly); this selection was therefore data-driven rather than pre-specified. This constitutes a form of data-driven endpoint selection and introduces potential optimism bias; results from this cohort should accordingly be interpreted with additional caution beyond that noted for the other two cohorts.

\paragraph{Sample size rationale.} The DWB cohort ($N = 7{,}497$) provides $>$99\% power to detect correlations of $|\rho| \geq 0.10$ at $\alpha = 0.05$ (post-hoc power analysis). The minimum detectable effect at 80\% power is $|\rho| = 0.036$, providing adequate sensitivity for the modest effect sizes typical in passive-sensing digital phenotyping. The GLOBEM cohort comprises 704 participant-wave observations from $N_{\text{unique}} = 497$ unique individuals; because some participants contributed multiple annual waves, power is computed conservatively on the number of unique participants rather than total observations. At $N = 497$, the cohort achieves 80\% power for $|\rho| \geq 0.13$ at $\alpha = 0.05$, and all validation tests that assume independent rows (e.g., Test~1, Held-out confirmation) are computed on one randomly selected wave per participant to eliminate within-subject correlation. The WEAR-ME cohort ($N = 1{,}078$) achieves 80\% power for $|\rho| \geq 0.09$.

\paragraph{Analysis status.} All analyses were exploratory; endpoints and analysis strategies were not registered in a public trial registry prior to pipeline execution. The CoDaS pipeline autonomously selects feature engineering strategies, statistical tests, and ML methods based on data characteristics. No subgroup analyses or biomarker candidates were pre-specified.

\section{Candidate Biomarker Prioritization Tasks}
We designed biomarker discovery tasks spanning mental health and metabolic disease domains to evaluate the ability of CoDaS to autonomously generate, test, and interpret physiological biomarkers from wearable data. Each task was formulated as a hypothesis generation problem in which the system iteratively proposes candidate features, executes statistical tests, and refines its search guided by mechanistic priors from the medical literature. This design parallels recent frameworks for open ended discovery in artificial intelligence while remaining grounded in real world biomedical data consistent with extensive cohort studies of behavioral tracking and cardiometabolic risk.
\subsection{Task 1: Mental Health and Circadian Resilience Signatures}
Using the Digital Wellbeing Hourly dataset, CoDaS aimed to discover wearable correlates of depression severity as measured by the Patient Health Questionnaire 8. The system integrated millions of hourly observations spanning resting heart rate, step execution, and smartphone application usage to identify behavioral states associated with depression symptom severity. Iterative hypothesis refinement by generative interpreters, guided by clinical literature on sleep disturbance, directed the search toward composite indices capturing circadian resilience. Quantitative findings are reported in Section \ref{sec:experiments_and_results}.
\subsection{Task 2: Longitudinal Behavior Modeling for Depression}
To assess autonomous pipeline generalization across diverse sensor modalities and temporal cohorts, CoDaS evaluated the Generalization of Longitudinal Behavior Modeling dataset. This task required identifying robust behavioral signatures of shifting depression severity tracked over four consecutive annual cohorts (2018--2021). 
Rather than relying strictly on continuous physiological output, the system adapted to passive smartphone sensing and aggregated wearable metrics. The parallel exploration runners surfaced diminished geospatial mobility and increased day-to-day sleep variability as candidate indicators of depressive transitions. Despite the inherent noise of longitudinal passive tracking, CoDaS maintained its rigorous evaluation framework, utilizing adversarial validation protocols to discard spurious correlations resulting from academic calendar artifacts. This task provided an opportunity to stress-test the pipeline under noisier passive-sensing conditions outside more controlled clinical settings.
\subsection{Task 3: Metabolic Risk Stratification and Insulin Resistance}
Using the Wearables for Metabolic Health cohort comprising 1{,}078 comprehensively characterized participants, CoDaS investigated physiological features predictive of metabolic dysfunction, verified against comprehensive fasting laboratory assays. Through extensive temporal modeling over multimodal continuous sensing streams, the system sought to identify composite features mapping directly to insulin resistance measured via the homeostasis model assessment of insulin resistance. This task uniquely challenged CoDaS to distinguish wearable derived (noninvasive) biomarkers from clinical laboratory features, with both categories subjected to the full internal screening battery. Quantitative findings are reported in Section \ref{sec:experiments_and_results}.
\subsection{Cross Domain Validation and Interpretability}
In the DWB and WEAR-ME cohorts, CoDaS produced biomarker candidates that were internally robust and physiologically coherent; the GLOBEM cohort yielded fewer battery-passing candidates, consistent with its analytical constraints (54.6\% feature-level missingness, coarse PHQ-4 endpoint; see Section~\ref{sec:experiments_and_results}). Cross domain evaluation revealed strong alignment between statistical significance and mechanistic plausibility. In DWB and WEAR-ME, the principal candidates reported in Table \ref{tab:biomarker_summary} passed the structured internal validation battery or were explicitly downgraded for construct overlap. In GLOBEM, low-signal candidates are reported separately as exploratory or unstable because several passed only a subset of the checks under substantial missingness and near-chance predictive performance.
Ablation evidence throughout development indicates that operating without literature-grounded reasoning substantially reduces physiological interpretability, underscoring the importance of integrating autonomous search with domain priors. Together, these results position CoDaS as a capable framework for interpretable biomarker discovery across heterogeneous disease processes.

\begin{table}[t!]
\centering
\caption{\textbf{Comparison of agent-based scientific discovery frameworks.}
CoDaS integrates hypothesis-generation, literature reviews, and statistical validation specifically for wearable biomarker prioritization. 
While existing systems emphasize general scientific reasoning or algorithmic discovery, CoDaS implements physiologically interpretable, population-level candidate-prioritization and internal validation procedures for wearable healthcare. $\cdot$ indicates not applicable or not designed for this domain}
\label{tab:ai_systems_comparison}
\resizebox{\textwidth}{!}{%
\begin{tabular}{lccccc}
\toprule
\textbf{Method} & \textbf{CoDaS} & \textbf{AI co-scientist} & \textbf{AlphaEvolve} & \textbf{Biomni} & \textbf{Data Science Agent} \\
& \textbf{(Ours)} & \raisebox{-0.0\height}{\includegraphics[height=1.5em]{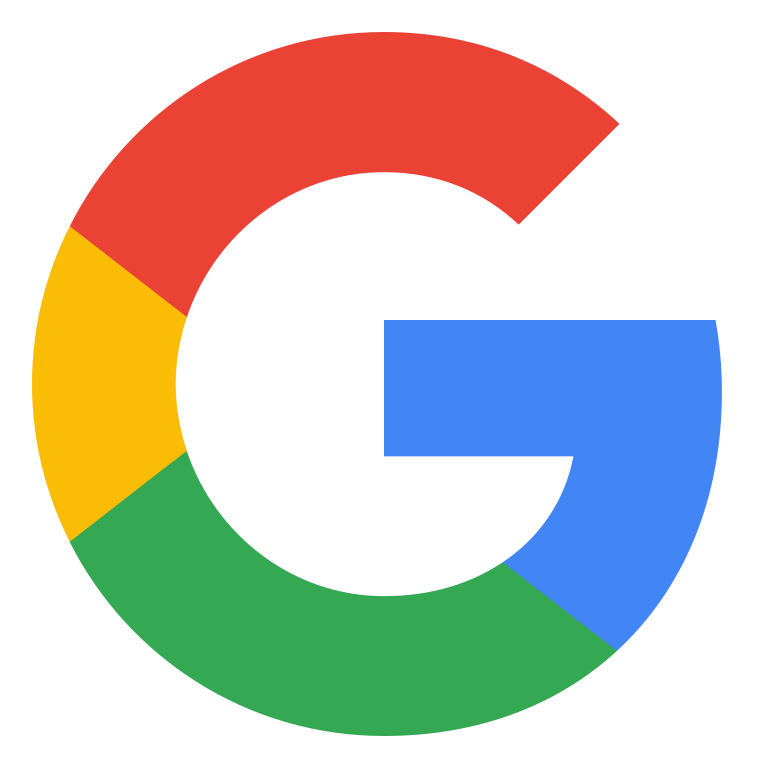}} & \raisebox{-0.0\height}{\includegraphics[height=1.5em]{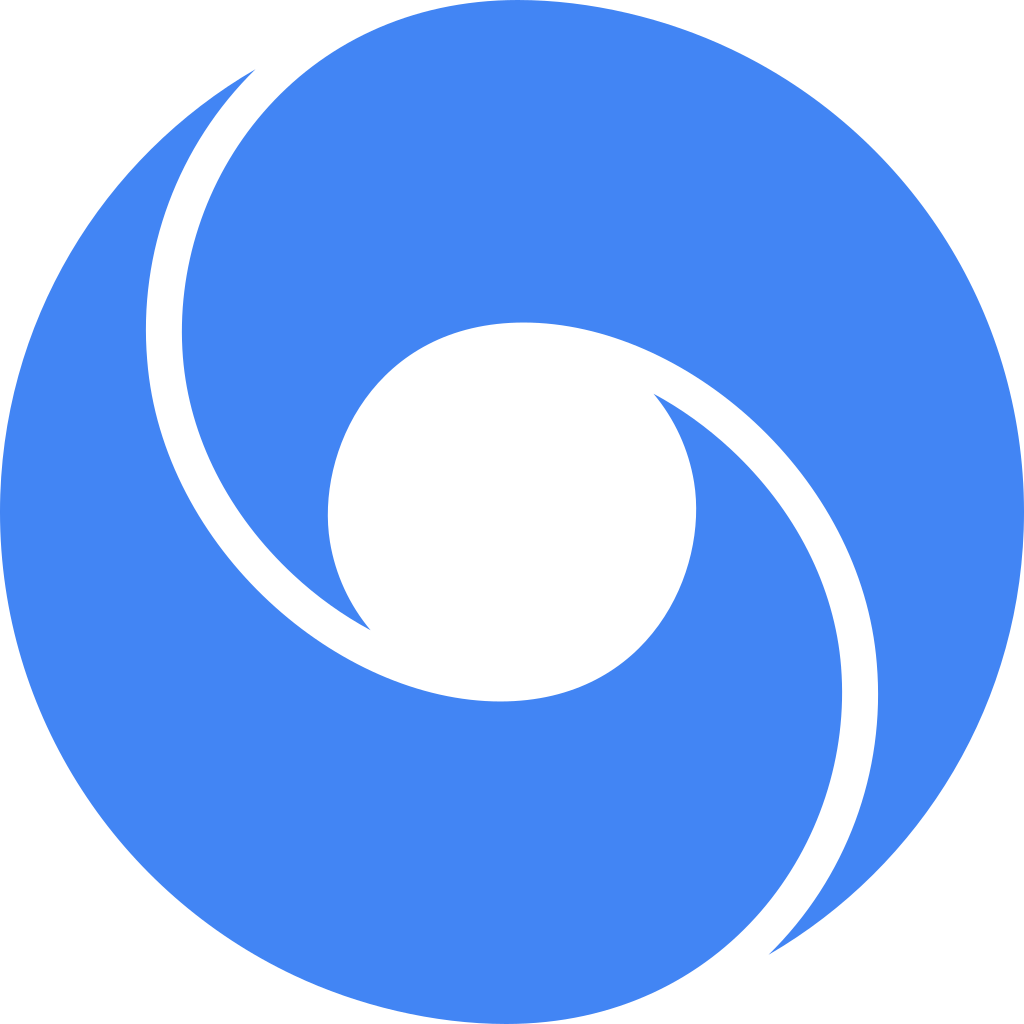}} & \raisebox{-0.0\height}{\includegraphics[height=1.5em]{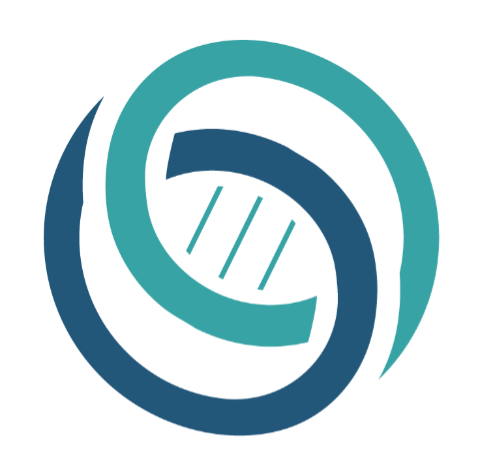}} & \raisebox{-0.0\height}{\includegraphics[height=1.6em]{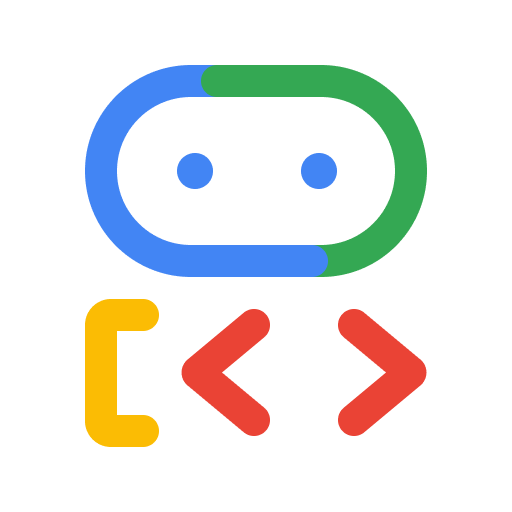}} \\
\midrule
\textbf{Primary Focus} & Biomarker discovery & General science discovery & Algorithm design & Biomedical AI & Data analysis \\
\midrule
\textbf{1. Research} & & & & & \\
Hypothesis generation & \checkmark & \checkmark & $\cdot$ & \checkmark & $\cdot$ \\
Literature exploration & \checkmark & \checkmark & $\cdot$ & \checkmark & $\cdot$ \\
Experimental design & \checkmark & \checkmark & $\cdot$ & \checkmark & $\cdot$ \\
Iterative refinement & \checkmark & \checkmark & \checkmark & \checkmark & $\cdot$ \\
\midrule
\textbf{2. Data Science} & & & & & \\
Time-series analysis & \checkmark & $\cdot$ & $\cdot$ & $\cdot$ & $\cdot$ \\
Wearable data processing & \checkmark & $\cdot$ & $\cdot$ & $\cdot$ & $\cdot$ \\
Code execution & \checkmark & \checkmark & $\cdot$ & \checkmark & \checkmark \\
Statistical validation & \checkmark & \checkmark & $\cdot$ & $\cdot$ & $\cdot$ \\
Population-level calibration & \checkmark & $\cdot$ & $\cdot$ & $\cdot$ & $\cdot$ \\
\midrule
\textbf{3. Collaboration \& Oversight} & & & & & \\
Human-in-the-loop & \checkmark & \checkmark & $\cdot$ & \checkmark & $\cdot$ \\
Multi-agent debate & \checkmark & \checkmark & $\cdot$ & $\cdot$ & $\cdot$ \\
Transparent audit trail & \checkmark & \checkmark & $\cdot$ & \checkmark & $\cdot$ \\
\midrule
\textbf{4. Domain} & & & & & \\
Healthcare/Medical & \checkmark & $\cdot$ & $\cdot$ & \checkmark & $\cdot$ \\
Wearable & \checkmark & $\cdot$ & $\cdot$ & $\cdot$ & $\cdot$ \\
% Algorithmic science & \checkmark & \checkmark & \checkmark & \checkmark & \checkmark \\
% \midrule
% \textbf{Model Foundation} & Gemini-3 Pro & Gemini 2.0 Flash & Gemini Flash/Pro & GPT-based & Various LLMs \\
% \textbf{Open Access} & In development & Trusted testers & Internal & Commercial & Research benchmark \\
\bottomrule
\end{tabular}%
}
\vspace{2mm}
\raggedright
\footnotesize
% \textsuperscript{$\ast$}AI co-scientist's data science research module executes code via a Colab-based sandbox (data loading, aggregation, correlation, ML modeling) as part of its knowledge-building phase; however, code execution is orchestrated by the platform rather than driven by the hypothesis generation pipeline.
\end{table}

\paragraph{Comparison with existing AI discovery systems.}
Table~\ref{tab:ai_systems_comparison} positions CoDaS relative to four contemporary AI-assisted discovery systems. Google's \textbf{AI co-scientist}~\citep{aicoscientist2024} is a general-purpose scientific reasoning system that excels at hypothesis generation, literature synthesis, and multi-agent debate (tournament-based idea refinement). Its data science research module can execute code in a sandboxed environment (data loading, aggregation, statistical analysis, and ML modeling) to support hypothesis evaluation; however, code execution is orchestrated by the platform as a knowledge-building step rather than integrated into the iterative hypothesis generation loop. Google DeepMind's \textbf{AlphaEvolve}~\citep{ding2024alphaevolve} targets algorithmic and mathematical discovery through evolutionary search with iterative refinement, operating in a fundamentally different domain from clinical biomarker identification; it does not perform literature search, hypothesis generation, or statistical validation in the biomedical sense. \textbf{Biomni}~\citep{biomni2024} is a general-purpose biomedical AI agent that executes LLM-generated Python code within a ReAct loop, performs PubMed literature search, and can reason about experimental design; however, it does not implement structured statistical validation pipelines, cross-validated ML with FDR correction, or population-level calibration against clinical endpoints. Google ADK's \textbf{Data Science (DS) Agent} executes a deterministic Python pipeline that can load data, compute correlations, and train ML models, but operates without literature grounding, hypothesis generation, adversarial validation, or iterative refinement; it runs a fixed linear sequence (load $\rightarrow$ clean $\rightarrow$ EDA $\rightarrow$ ML) without LLM-guided analytical decision-making. Among the evaluated configurations and based on public system descriptions, CoDaS is the only system in this comparison that integrates these four capability dimensions within a wearable biomarker discovery workflow.
\begin{table}[ht!]
\centering
\caption{\textbf{Candidate biomarkers prioritized by CoDaS across three cohorts.} DWB and WEAR-ME candidates passed or were explicitly downgraded by the structured internal validation battery. GLOBEM candidates are included as low-signal exploratory or unstable associations and should not be interpreted as battery-passing validated candidates. One rejected candidate (TG/HDL ratio) is included for transparency as a sanity check demonstrating the pipeline's construct independence gate. Effect sizes are Spearman correlations ($\rho$). Adjusted $p$ values reflect Benjamini–Hochberg FDR correction ($\alpha = 0.05$). Cross-validation metrics are reported for models trained on the candidate biomarker sets. 95\% confidence intervals are obtained by bootstrap resampling (1{,}000 replicates) for the DWB and WEAR-ME cohorts. For GLOBEM, where repeated within-participant observations preclude simple row resampling, intervals are analytic (Fisher transformation).}
\label{tab:biomarker_summary}

\resizebox{\textwidth}{!}{%
\begin{tabular}{p{2.7cm} p{3.5cm} c c c p{9.8cm} c}
\toprule
\textbf{Cohort} & \textbf{Biomarker} & \textbf{Effect Size ($\rho$)} & \textbf{95\% CI} & \textbf{Adj.\ $p$} & \textbf{Mechanistic Hypothesis} & \textbf{Prior Evidence} \\
\midrule

% ----- DWB -----
\multirow{9}{3.4cm}[-2.2em]{\textbf{DWB Hourly} \newline (N\,=\,7{,}497;\\target: PHQ-8;\\Ridge CV $R^{2}$ = 0.228\textsuperscript{$\ddagger$})}

& Main sleep duration variability
& 0.252
& [0.23, 0.27]
& $<$\,0.001
& Circadian instability impairs sleep homeostatic drive and emotional regulation
& Established \\[3pt]

& Nocturnal social app usage
& 0.246
& [0.22, 0.27]
& $<$\,0.001
& Blue-light exposure and hyperarousal suppress melatonin secretion
& Established \\[3pt]

& Late-night doomscrolling
& 0.177
& [0.16, 0.20]
& $<$\,0.001
& Nocturnal news/social scrolling sustains rumination and cortisol release
& Supported\textsuperscript{$\star$} \\[3pt]

& Night-to-day social media ratio
& 0.222
& [0.20, 0.24]
& $<$\,0.001
& Displaced nocturnal social engagement reflects rumination and insomnia
& Supported\textsuperscript{$\star$} \\[3pt]

& Hedonic-to-productivity app ratio
& 0.152
& [0.13, 0.17]
& $<$\,0.001
& Behavioral narrowing may increase hedonic relative to productivity-oriented app use in participants with higher depressive symptom scores
& Emerging\textsuperscript{$\star\star$} \\[3pt]

& Polyphasic sleep percentage
& 0.193
& [0.17, 0.21]
& $<$\,0.001
& Fragmented nocturnal architecture reflects HPA axis dysregulation
& Emerging\textsuperscript{$\star\star$} \\

\midrule

% ----- GLOBEM -----
\multirow{5}{3.4cm}[-1.0em]{\textbf{GLOBEM} \newline (N\,=\,704;\\target: PHQ-4;\\CV AUC = 0.535)}

& Sleep onset time variability (circadian acrophase)
& 0.126
& [0.05, 0.20]
& $<$\,0.001
& Irregular sleep scheduling disrupts circadian entrainment
& Established \\[3pt]

& Evening incoming call duration (circadian acrophase)
& $-$0.145
& [$-$0.22, $-$0.07]
& $<$\,0.001
& Reduced evening social communication reflects social withdrawal
& Flagged unstable \\[3pt]

& WiFi AP sequential diversity (7-day)
& 0.128
& [0.05, 0.20]
& $<$\,0.001
& Dynamic network scanning as proxy for environmental instability
& Emerging\textsuperscript{$\star\star$} \\

\midrule

% ----- WEAR-ME -----
\multirow{14}{3.4cm}[-3.0em]{\textbf{WEAR-ME} \newline (N\,=\,1{,}078;\\target: HOMA-IR;\\Ridge CV $R^{2}$ = 0.389\textsuperscript{$\S$})}

& Derived TG/HDL ratio\textsuperscript{R}
& 0.542
& [0.50, 0.58]
& $<$\,0.001
& Atherogenic dyslipidemia directly indexes hepatic insulin resistance
& Established \\[3pt]

& HDL cholesterol
& $-$0.380
& [$-$0.43, $-$0.33]
& $<$\,0.001
& Low HDL reflects impaired reverse cholesterol transport in metabolic syndrome
& Established \\[3pt]

& C-reactive protein (CRP)
& 0.393
& [0.34, 0.44]
& $<$\,0.001
& Systemic inflammation mediates adipose-derived insulin resistance via TNF-$\alpha$/IL-6
& Established \\[3pt]

& Derived AST/ALT ratio (De Ritis)
& $-$0.375
& [$-$0.42, $-$0.32]
& $<$\,0.001
& Hepatic gluconeogenic stress and subclinical steatosis marker
& Emerging\textsuperscript{$\star\star$} \\[3pt]

& Resting heart rate (mean)$^{\dagger}$
& 0.348
& [0.30, 0.40]
& $<$\,0.001
& Autonomic imbalance: sympathetic overdrive increases hepatic glucose output
& Established \\[3pt]

& Cardiovascular fitness index (steps/resting HR)$^{\dagger}$
& $-$0.374
& [$-$0.42, $-$0.32]
& $<$\,0.001
& May proxy cardiorespiratory fitness, habitual activity and autonomic tone, pathways linked to insulin sensitivity
& Supported\textsuperscript{$\star$} \\[3pt]

& Red cell distribution width (RDW)
& 0.281
& [0.22, 0.33]
& $<$\,0.001
& Erythropoietic stress marker of chronic low-grade metabolic inflammation
& Emerging\textsuperscript{$\star\star$} \\[3pt]

& Albumin/globulin ratio
& $-$0.220
& [$-$0.29, $-$0.17]
& $<$\,0.001
& Hepatic synthetic dysfunction and subclinical inflammatory protein shift
& Emerging\textsuperscript{$\star\star$} \\

\bottomrule
\end{tabular}%
}

\vspace{4pt}
\raggedright
\footnotesize
\textbf{Established}\,=\,consistent with known clinical associations with substantial literature support;
\textbf{Supported}\textsuperscript{$\star$}\,=\,the underlying physiological axis is established but this specific operationalization from wearable or digital phenotyping data is new;
\textbf{Emerging}\textsuperscript{$\star\star$}\,=\,limited prior evidence for this specific feature--endpoint association within the searched literature corpus; requires held-out confirmation.
$^{\textrm{R}}$Rejected by the construct-overlap and clinical-redundancy gate: triglycerides and HDL are routine lipid measurements already used in metabolic risk assessment, so the ratio was treated as a positive control for distinguishing statistical validity from incremental clinical value. TG/HDL is not an algebraic component of HOMA-IR and was not counted among non-rejected candidates.
$^{\dagger}$Wearable-derived feature. All other WEAR-ME features are derived from fasting clinical laboratory panels.
$^{\ddagger}$Ridge regression CV $R^{2}$ using demographics plus top CoDaS-selected biomarkers; 5-fold nested cross-validation.
$^{\S}$Ridge regression CV $R^{2}$ using demographics plus top CoDaS-selected biomarkers; 5-fold nested cross-validation.
$^{\P}$Discovery and held-out estimates differed in sign for this feature.
\end{table}

\section{Results}
\label{sec:experiments_and_results}

To evaluate the discovery capabilities of CoDaS, we deployed the framework independently on three clinically distinct datasets: high-frequency mental health tracking (Digital Wellbeing Hourly, N = 7{,}497), multiwave longitudinal behavioral phenotyping (GLOBEM, N = 704), and cross-sectional cardiometabolic risk stratification anchored by fasting blood panels (Wearables for Metabolic Health cohort, N = 1{,}078). Each dataset presents qualitatively different analytical challenges; dense temporal streams, sparse longitudinal sensing with severe missingness, and wearable laboratory feature integration, allowing us to assess the pipeline's versatility across diverse data modalities and clinical endpoints without implying transfer or generalization between cohorts.

All reported candidate biomarkers are subjected to a structured internal validation battery organized around four complementary dimensions: replication, stability, robustness, and discriminative power, operationalized via 11 checks executed in a deterministic subprocess (detailed in Section~\ref{sec:validation_integrity}). For each dataset and discovery round, we applied Benjamini–Hochberg FDR correction ($\alpha = 0.05$) across the full family of univariate feature tests evaluated in that round. Candidates surviving round-level correction were then tracked across rounds, and cumulative reporting was restricted to features that remained significant under the final round-level screened set. Predictive performance is reported using 5-fold cross-validation with stratified participant-level splits to prevent information leakage between discovery and evaluation partitions. Subgroup consistency was assessed across biological sex (female vs.\ male) and age decade; features were required to show a consistent direction of effect across all subgroups to pass the pipeline. Statistical power for the age-decade test is concentrated in the young-to-middle-aged range that dominates all three cohorts. In DWB about 11\% of participants were younger than 30 and 32\% were 50 or older, with only 3.5\% aged 70 or older. In WEAR-ME 39\% were 50 or older and 5.1\% (55 participants) were 70 or older. GLOBEM was an undergraduate cohort with no participant over 25. The age-decade check should therefore be read as evidence of directional stability across young-to-middle-aged strata rather than across the full adult age range, and the candidate biomarkers require dedicated validation in older, multimorbid populations (see Limitations). We note that the 11 checks are not fully independent (e.g., features with strong replication correlations will typically also pass bootstrap stability); the battery is best understood as a structured audit across four complementary validation dimensions rather than 11 orthogonal significance tests. No blinding of the analytical pipeline was performed, as CoDaS supports optional expert feedback; human review occurred in the post-discovery feedback phase and was restricted to mechanistic interpretation (see Section~\ref{sec:leakage_firewall}). A summary of comparative performance across methods and datasets is provided in Table \ref{tab:ablation_results}.

\subsection{Mental Health Monitoring via Digital Phenotyping}
\label{subsec:mental_health}
In the Digital Wellbeing Hourly study, CoDaS navigated over 4.5 million physiological and device interaction records to identify candidate predictors of depression severity (Patient Health Questionnaire 8). The system identified structural variance in sleep, quantified as main sleep duration variability ($\rho = 0.252$, 95\% CI [0.23, 0.27], $p < 0.001$), as the top-ranked candidate predictor of depression severity. This finding is consistent with an established body of literature linking sleep variability to depression severity \citep{fang2021day}, and its autonomous recovery by CoDaS serves as a sanity check demonstrating the pipeline's ability to recapitulate known clinical signals without prior instruction. Additionally, CoDaS identified elevated nocturnal social application usage ($\rho = 0.246$, 95\% CI [0.22, 0.27], $p < 0.001$) as a candidate behavioral correlate, and generated the hypothesis that nocturnal digital engagement may index nocturnal hyperarousal or sleep-disruptive behavior.

To quantify the incremental value of these candidates beyond established sociodemographic predictors, we compared nested Ridge regression models under 5-fold cross-validation: a demographics-only baseline (8 covariates: age, gender, education, marital status, disability status, Hispanic ethnicity, financial status, living arrangement) achieved CV $R^{2} = 0.188$ (SD 0.010), while adding the top five CoDaS-selected biomarkers yielded CV $R^{2} = 0.228$ (SD 0.010), yielding a $\Delta R^{2}$ of 0.040. Although modest, this increment is statistically stable across folds and demonstrates that wearable-derived features capture variance in depression severity not explained by sociodemographic factors alone. Effect sizes are modest in absolute magnitude ($\rho = 0.15$ to $0.25$), consistent with the well-documented difficulty of predicting PHQ scores from passive sensing alone, and should be interpreted as prioritized hypothesis-generating signals warranting prospective replication. Since PHQ-8 includes a sleep-related symptom item, sleep-variability candidates may partly capture criterion overlap with the depression endpoint rather than an entirely independent behavioral signal. We therefore interpret these sleep-related findings as candidate monitoring signals requiring endpoint designs that separate sleep symptoms from non-sleep affective burden.

Notably, several of CoDaS's highest-ranked candidates are not raw sensor features but \textit{autonomously constructed composite indices}: the night-to-day social media ratio ($\rho = 0.222$), which captures the displacement of social engagement into nocturnal hours; the hedonic-to-productivity app ratio ($\rho = 0.152$), which captures relative enrichment of hedonic compared with productivity-oriented phone use rather than anhedonia itself; and polyphasic sleep percentage ($\rho = 0.193$), the proportion of nights exhibiting multiple distinct sleep episodes. These composites were generated by the pipeline's feature engineering phase guided by mechanistic priors from the literature grounding phase. The ability to construct and prioritize clinically interpretable composite features, rather than merely rank existing variables, distinguishes CoDaS from conventional automated feature-selection pipelines. Complete lists of all battery-passing (screened and conditionally prioritized) candidates for each cohort are provided in Tables~\ref{tab:full_dwb}--\ref{tab:full_globem} in the Appendix. These sleep and nocturnal-use candidates should also be interpreted within the age and health profile of the discovery cohort. Sleep fragmentation, polyphasic sleep and drifting sleep timing can arise from aging, medical comorbidity or medication exposure, so specificity for depression should not be assumed in older or multimorbid populations without age-stratified prospective validation. Throughout this paper, \textit{screened} refers exclusively to survival of the internal validation battery and does not imply prospective clinical validation, external replication, or regulatory endorsement (see Limitations).

Imputation sensitivity analysis confirmed that all prioritized candidates were stable across three imputation strategies (median, KNN, iterative; maximum $\Delta\rho < 0.001$), and threshold sensitivity analysis showed that the same 13 features survived at both default ($p < 0.05$, $|\rho| \geq 0.20$) and lenient ($p < 0.10$, $|\rho| \geq 0.10$) thresholds, indicating robustness to analytical choices.

When evaluated on the GLOBEM dataset as a stress test of pipeline robustness under data-poor conditions, CoDaS extracted subtle environmental phenotypes across multiwave sensing periods, achieving a primary classification CV AUC of 0.535 using the all-screened-feature evaluation protocol. A separate top-five logistic-regression ablation produced a higher AUC and is reported only as an exploratory component analysis, not as the primary GLOBEM performance estimate. This near-chance discriminative performance reflects the substantial analytical challenges inherent to this cohort: 54.6\% feature level missingness, the coarseness of the PHQ-4 outcome instrument (4-item, 0--12 scale), and the heterogeneity of longitudinal passive sensing across annual cohorts. Although discriminative performance was near chance, this result is consistent with a conservative pipeline that did not surface stronger predictive claims from a sparse and noisy cohort. Despite this performance ceiling, the pipeline identified the sequential diversity of unique WiFi access points encountered over a seven day trend ($\rho = 0.128$, 95\% CI [0.05, 0.20], $p < 0.001$) as a candidate predictor of depressive shifts. While previous digital phenotyping studies have correlated static location entropy with depression \citep{saeb2015mobile}, the CoDaS-generated mechanistic hypothesis advanced this by framing dynamic network scanning as a proxy for what the system termed ``environmental instability'' and ``agitated restlessness'' (CoDaS-generated interpretive labels, not established clinical constructs). The Google Data Science Agent baseline achieved a nearly identical predictive variance (CV AUC = 0.523), suggesting that the discriminative ceiling is largely data-driven rather than method-driven.

\subsection{Cardiometabolic Risk Stratification}
Within the Wearables for Metabolic Health cohort, CoDaS mapped continuous physiological readouts against clinical fasting assays to identify candidate predictors of the homeostasis model assessment of insulin resistance. The pipeline first recovered established clinical laboratory markers as a sanity check supporting the analytical validity of the pipeline. It derived TG/HDL ratio ($\rho = 0.542$, $p < 0.001$; subsequently rejected by the construct independence gate as non-independent of the target construct), C reactive protein ($\rho = 0.393$, 95\% CI [0.34, 0.44], $p < 0.001$), and HDL cholesterol ($\rho = -0.380$, $p < 0.001$). These laboratory features are included solely to demonstrate that the pipeline recovers known metabolic relationships; the primary translational claim of this cohort rests on the noninvasive wearable-derived candidates described below.

Transitioning to noninvasive wearable sensors, CoDaS identified a derived cardiovascular fitness index (the ratio of step counts to resting heart rate) as a robust metabolic predictor ($\rho = -0.374$, 95\% CI [-0.42, -0.32], $p < 0.001$). The system's mechanistic engine anchored this finding in established metabolic literature \citep{petersen2018mechanisms}, hypothesizing that the index reflects peripheral glucose disposal efficiency, a process primarily driven by skeletal muscle mitochondrial capacity. Beyond the cardiovascular fitness index, CoDaS autonomously constructed two additional composite features with strong associations: a derived AST/ALT ratio (the De Ritis ratio; $\rho = -0.375$, $p < 0.001$), a hepatic function index used in liver disease staging, and a known correlate of insulin resistance, here recovered as a strong signal in a general population cohort; and an HRV-to-RHR ratio ($\rho = -0.203$, $p < 0.001$), capturing the balance between parasympathetic tone and sympathetic activation (not shown in Table~\ref{tab:biomarker_summary}, which reports a curated subset of candidates per cohort). The pipeline also identified red cell distribution width (RDW; $\rho = 0.281$, $p < 0.001$) as an emerging metabolic inflammation marker \citep{salvagno2015red, barbieri2001new}, consistent with a small but growing body of evidence linking erythropoietic stress to insulin resistance. Because AST/ALT ratio and RDW can also reflect frailty, sarcopenia, anemia or chronic inflammatory disease, these laboratory candidates should be interpreted as metabolic-risk correlates in this cohort rather than disease-specific insulin-resistance biomarkers. To isolate the wearable-only contribution, a Ridge model using only wearable-derived features achieved CV $R^{2} = 0.281$, while the full model including clinical laboratory features achieved CV $R^{2} = 0.389$; relative to the demographics-only baseline (CV $R^{2} = 0.260$), wearable features alone contributed $\Delta R^{2} = 0.021$, a modest increment that highlights the difficulty of extracting metabolic signal from consumer-grade wearables when clinical laboratory features are available. The primary translational value of the wearable-derived cardiovascular fitness index lies not in predictive superiority over blood panels but in its noninvasive, continuously measurable nature, which could enable longitudinal monitoring where repeated phlebotomy is impractical. This interpretation assumes that mobility limitations and rate-modifying medications are observed or can be adjusted for. In older or multimorbid populations, step counts can reflect osteoarthritis, gait limitation or disability, and resting heart rate can be driven by beta-blockers or other cardiovascular medications, so the index requires validation of medication and mobility prior to general clinical use. The Google Data Science Agent baseline, which framed the task as binary insulin-resistance classification, struggled with the inherent collinearity of continuous physiological data, yielding an overfit random forest classifier that failed to generalize beyond chance (CV AUC = 0.429).

\begin{figure}[t!]
\centering
\includegraphics[width=0.85\textwidth]{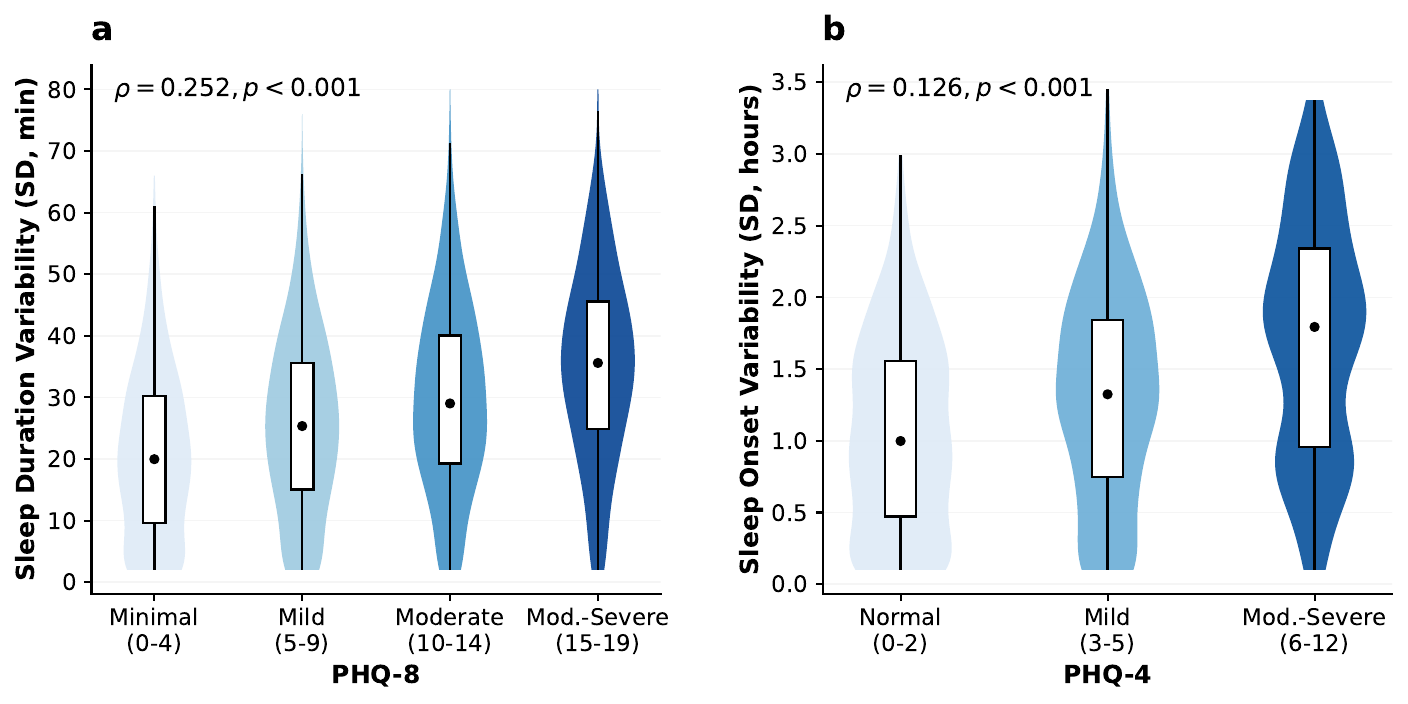}
\caption{\textbf{Related circadian-instability candidates in two depression cohorts.} \textbf{(a)} The DWB cohort identifies sleep duration variability as a top-ranked candidate correlate of depression severity, and \textbf{(b)} The GLOBEM cohort surfaces sleep onset variability as an exploratory circadian-instability-related candidate. Although the specific operationalizations differ, both cohorts point to circadian instability as a hypothesis-generating construct. This pattern is suggestive rather than confirmatory, especially given the near-chance discriminative performance in GLOBEM. Violins show the distribution of the candidate biomarker across depression severity bands. Internal boxplots show medians and interquartile ranges. The increasing trend across severity strata is consistent with a circadian-instability hypothesis, but should not be interpreted as direct replication because the cohorts, instruments and feature operationalizations differ.}
\label{fig:construct_convergence}
\end{figure}

\subsection{Hypothesis-generating convergence across depression cohorts}
\label{sec:construct_convergence}

Although no identical candidate feature was replicated in an independent cohort with an aligned endpoint, the two depression-focused cohorts provided hypothesis-generating evidence for related circadian-instability features. In the DWB cohort (N = 7,497, PHQ-8), CoDaS ranked sleep duration variability as the top candidate correlate of depression severity ($\rho = 0.252$). In the GLOBEM cohort (N = 704 participant-wave observations from 497 unique individuals, PHQ-4), sleep onset time variability also emerged as an exploratory circadian-instability candidate ($\rho = 0.126$). The pipeline also initially prioritized evening incoming call duration ($\rho = -0.145$), but we treat this association as unstable rather than as evidence for cross-cohort consistency because its held-out estimate reversed direction and the clinician panel rated it lowest for validity. The DWB and GLOBEM sleep-related candidates differ in operationalization, endpoint instrument, and study population (US adults versus US college students), so this pattern should be interpreted as suggestive construct-level convergence around circadian instability rather than direct replication. Given the near-chance discriminative performance in GLOBEM (CV AUC = 0.535), this observation remains hypothesis-generating.

\subsection{Ablation of Adversarial Evaluation and Failure Cases}
\label{sec:ablation}
To quantify the contribution of each CoDaS component, we conducted systematic ablation experiments across all three datasets, removing one module at a time from the full pipeline (Table~\ref{tab:ablation_results}).

\begin{table}[t!]
\centering
\caption{\textbf{Exploratory component ablation under fixed evaluation settings.} CV $R^{2}$ (Demographics + Biomarkers model) for DWB and WEAR-ME, and CV AUC (Logistic Regression, Demographics + Biomarkers) for GLOBEM (binary depression classification, PHQ-4 $> 2$), across all three datasets for each ablation condition. Higher is better except where leakage is indicated ($^\dagger$). N~Val.~= number of candidates passing the validation battery in each ablation run; these counts reflect each run's own discovery output and may differ from the curated candidates reported in the appendix tables (Tables~\ref{tab:full_dwb}--\ref{tab:full_globem}), which apply the stricter 11-test pipeline to the canonical full-pipeline run.}
\label{tab:ablation_results}
\footnotesize
\begin{tabular}{lcccccc}
\toprule
& \multicolumn{2}{c}{\textbf{DWB}} & \multicolumn{2}{c}{\textbf{GLOBEM}} & \multicolumn{2}{c}{\textbf{WEAR-ME}} \\
\cmidrule(lr){2-3} \cmidrule(lr){4-5} \cmidrule(lr){6-7}
\textbf{Configuration} & \textbf{CV $R^{2}$} & \textbf{N Val.} & \textbf{CV AUC} & \textbf{N Val.} & \textbf{CV $R^{2}$} & \textbf{N Val.} \\
\midrule
Full CoDaS & 0.228 & 35 & 0.694 & 48 & 0.389 & 49 \\
$-$ Adversarial debate & 0.228 & 46 & 0.694 & 48 & 0.407 & 42 \\
$-$ Iterative loop & 0.217 & 35 & 0.653 & 4 & 0.387 & 50 \\
$-$ Literature grounding & 0.207 & 49 & 0.694 & 50 & 0.407 & 43 \\
$-$ Scout agent & 0.120 & 48 & 0.707 & 48 & 0.365 & 50 \\
$-$ Reinvestigation & 0.228 & 48 & 0.694 & 48 & 0.407 & 45 \\
$-$ Validation procedure & 0.228$^\dagger$ & --- & 0.694 & --- & 0.407$^\dagger$ & --- \\
\midrule
Demographics only & 0.188 & --- & 0.588 & --- & 0.260 & --- \\
\bottomrule
\end{tabular}

\vspace{2pt}
\raggedright
\footnotesize
$^\dagger$Without the validation pipeline, leakage-inflated features may be present (see text); the CV $R^{2}$ is not directly comparable. ---\, indicates the metric is not applicable. The GLOBEM AUCs in this table are post-hoc component-ablation metrics using a top-five logistic-regression protocol. They should not be compared directly with the primary GLOBEM performance estimate in Section \ref{subsec:mental_health}, which used the all-screened-feature evaluation protocol. Ablation counts reflect pre-dedup / intermediate candidate set under the ablation evaluation protocol.\end{table}

The integration of the adversarial Critic and Defender debate framework differs from standard automated feature-selection pipelines. Without adversarial oversight, tautological features (such as monotonic transformations of the target variable, e.g., squared proxies of blood glucose) can pass statistical screening and manifest as severe data leakage. In a preliminary ablation configuration that additionally disabled construct overlap analysis, such features inflated the CV $R^2$ to 0.963 on WEAR-ME. In the full pipeline, the Critic agent successfully identified and rejected these tautologies (e.g., the feature \texttt{glucose\_sq} was explicitly rejected despite passing 10/11 statistical tests) by recognizing their lack of genuine construct independence, preventing the leakage feature from being reported as an independent candidate.

\paragraph{Adversarial audit.} The WEAR-ME leakage case illustrates the specific role of the Critic–Defender step. The Defender initially argued to retain $glucose\_sq$ because the feature passed 10 of 11 statistical checks and produced an apparently large gain in predictive performance. The audit flagged this interpretation as invalid because the feature was a monotonic transformation of an excluded glycaemic variable and therefore represented target-proximal leakage rather than an independent candidate biomarker. The Orchestrator consequently rejected the feature despite its statistical strength. A second audit rejected TG/HDL despite its strong association with HOMA-IR because triglycerides and HDL are already clinically measured metabolic components; the candidate was retained only as a positive control for construct-independence gating. In contrast, the steps/resting heart-rate index was retained as a wearable-derived candidate because it was noninvasive, passed the internal audit, and was not a direct component or transformation of the target, while being flagged for medication- and mobility-aware prospective validation. Beyond leakage and redundancy, the same step addresses confounding and the correlation-to-causation gap. For behavioral candidates such as nocturnal application use, the Critic flags that the association may be driven by a shared cause rather than a direct effect, and because observational data cannot settle this question, such candidates are retained as hypothesis-generating signals subject to the causal-robustness control rather than asserted as causal. These four functions, namely detecting leakage, detecting redundant or proxy features, testing confounding sensitivity, and marking the correlation-to-causation limit, define what the adversarial step can and cannot establish.

We note that removing adversarial debate left the cross-validated $R^{2}$ essentially unchanged on DWB (0.228 in both conditions) and marginally higher on WEAR-ME (0.389 to 0.407). Because the debate stage filters candidates after model fitting rather than re-training the model, its contribution appears in the reliability and verdict quality of reported candidates, as in the \texttt{glucose\_sq} rejection above, rather than in the cross-validated point estimate. The small WEAR-ME difference also reflects single-run variation in the LLM-guided feature proposals. Removing the Scout agent (responsible for initial data profiling and clinical target identification) produced the largest performance degradation on DWB (CV $R^{2}$ dropping from 0.228 to 0.120), demonstrating that informed analytical framing is essential for efficient biomarker search in large feature spaces. In the post-hoc GLOBEM ablation protocol, removing the iterative discovery loop reduced the number of intermediate candidates from 48 to 4. Because this protocol differs from the primary all-screened-feature GLOBEM evaluation, we interpret this result as evidence about search behavior.

The system’s failure cases also suggest that the prioritization scheme was conservative, often rejecting statistically significant but low-value features. Across the Digital Wellbeing cohort, the data science agents initially generated a vast pool of 145 discrete statistical metrics. However, through the 11 component validation pipeline, the orchestrator autonomously filtered this funnel down to just 34 screened candidates, explicitly rejecting dozens of statistically significant but functionally trivial features (e.g., the standard deviation of hourly phone unlocks, $\rho = 0.059$, $p < 0.01$). This stringent self-regulation helps ensure that the final reported candidate features exhibit effect sizes relevant for hypothesis generation, rather than merely exploiting population-scale statistical power.

\paragraph{Autonomous feature construction.}
A distinctive property of CoDaS is its capacity to propose interpretable composite features that were not present as raw input columns. To avoid mixing primary and low-signal exploratory findings, we summarize this analysis only for the DWB and WEAR-ME candidate lists. Across these two primary cohorts, 9 of 59 non-rejected candidates (15.3\%) were autonomously constructed composite indices: four DWB digital-behavior ratios (night-to-day social media ratio, night-to-day unlock ratio, hedonic-to-productivity app ratio, and night-to-day screen ratio) and five WEAR-ME derived indices (cardiovascular fitness index, AST/ALT ratio, cholesterol/HDL ratio, HRV/RHR ratio, and albumin/globulin ratio). The rejected TG/HDL ratio is excluded from this count because it was retained only as a positive control for clinical redundancy and construct-overlap gating.

These composites were generated by feature-engineering module in response to literature-grounded physiological rationales. For instance, the cardiovascular fitness index was proposed after the system retrieved literature linking cardiorespiratory fitness, habitual activity, autonomic tone, and insulin sensitivity. In a descriptive comparison, the retained composite indices showed larger absolute associations with their endpoints than their listed constituent raw features (mean $|\rho| = 0.28$ for composites versus 0.21 for constituent raw features). This comparison was not used as an inferential test or as an independent validation criterion.

The value of explicit composite construction is interpretability rather than access to otherwise unavailable information. For the additive linear models used in the candidate summaries, a precomputed ratio provides a compact nonlinear transformation that is not represented by including only the two raw variables as separate additive terms. More flexible models with interactions or transformations could, in principle, recover related information. Therefore, we interpret autonomous feature construction as a mechanism for producing compact, auditable candidate definitions for follow-up testing, not as evidence that the ratios are causal biomarkers or that their constituent variables contain inaccessible signal.

\subsection{Benchmark Evaluation}

To assess whether the CoDaS architecture possesses the analytical capabilities required for biomarker discovery, spanning (i) data profiling, (ii) statistical analysis, (iii) causal inference, (iv) code driven computation, (v) iterative hypothesis refinement, and (vi) clinical reasoning, we evaluated the system across six benchmarks that collectively probe every stage of the translational data science and research pipeline (see Fig.~\ref{fig:benchmark_results}). Unlike narrow code completion benchmarks such as HumanEval~\citep{chen2021evaluating} or static medical knowledge examinations like MedQA~\citep{jin2021disease}, the selected benchmarks require end-to-end analytical reasoning over real-world tabular datasets, multi-step code generation and execution, and integration of domain expertise, the same skills that underpin biomarker identification from clinical and omics cohorts.

\paragraph{Datasets.} Each benchmark was chosen to stress test a specific facet of the biomarker discovery workflow. For instance, \textbf{DiscoveryBench}~\citep{majumder2024discoverybench} evaluates the complete hypothesis generation pipeline: given raw scientific datasets and a natural language research question, the system must autonomously perform exploratory data analysis, variable selection, statistical testing, and articulate a structured hypothesis, precisely the intellectual workflow a translational researcher follows when searching for candidate biomarkers in a new cohort. \textbf{HealthBench}~\citep{arora2025healthbench} and its \textbf{Hard} subset assess clinical reasoning fidelity across 5,000 physician designed, multiturn medical conversations spanning 674 diseases and 26 specialties, ensuring that the system's medical knowledge and safety aware communication meet the standards required for biomarker interpretation in clinical contexts. \textbf{DataSciBench}~\citep{zhang2025datascibench} measures end to end data science code generation across 222 tasks involving data cleaning, statistical computation, machine learning, and visualization, the programmatic building blocks of any computational biomarker pipeline. \textbf{DSBench}~\citep{jing2024dsbench} tests analytical reasoning over large, heterogeneous datasets sourced from real data competitions, requiring multitable joins, domain specific statistical reasoning, and quantitative answer extraction, challenges that directly parallel working with multimodal clinical datasets. Finally, \textbf{DSGym}~\citep{nie2026dsgym}, which integrates QRData~\citep{liu2024llms} (statistical and causal reasoning) and DAEval~\citep{hu2024infiagent} (data analysis tasks), specifically probes causal inference capabilities and quantitative reasoning grounded in real data, skills that are essential for distinguishing genuine biomarker associations from confounded correlations.

\paragraph{Results.}
We used external data-science and clinical-reasoning benchmarks as supplementary stress tests of the system components. We emphasize that this comparison is inherently asymmetric: CoDaS is a domain-specialized multi-agent \textit{system} with iterative execution, domain-specific tooling, and multi-agent orchestration, whereas baselines are individual models or generic agent frameworks; reported margins should be interpreted as system-level comparisons under this evaluation setup rather than model-to-model differences (Fig.~\ref{fig:benchmark_results}). These benchmarks serve as supplementary validation of the analytical capabilities underlying the biomarker discovery pipeline, not as the primary contribution of this work. On \textbf{HealthBench}, CoDaS attained an overall score of 0.724, surpassing the previous best result of o3 (0.598) by 12.6 percentage points. On the \textbf{HealthBench Hard} subset, which comprises cases empirically identified as resistant to all frontier models, CoDaS scored above the o3 baseline in this evaluation. The remaining baselines scored substantially lower: GPT-4.1 at 0.230, Gemini-2.5 Pro at 0.190, and o1 at 0.160. On \textbf{DiscoveryBench}, CoDaS achieved a Hypothesis Matching Score (HMS) of 0.32 on real scientific datasets, exceeding the oracle assisted Reflexion agent with GPT-4o backbone (0.24) by 8.0 percentage points and the Reflexion agent with Llama-3 backbone (0.23) by 9.0 percentage points. Notably, the Reflexion+Oracle baseline receives the gold HMS score as iterative feedback, an advantage CoDaS does not require; without oracle access, CodeGen and ReAct agents achieve only 0.15 HMS each. On \textbf{DataSciBench}, CoDaS reached a completion rate of 77.5\%, outperforming GPT-4o (68.4\%) by 9.1 percentage points, DeepSeek-Coder-33B (61.2\%) by 16.3 pp, GPT-4 Turbo (58.9\%) by 18.6 pp, and Claude-3.5 Sonnet (58.1\%) by 19.4 pp. On \textbf{DSBench}, CoDaS achieved 64.0\% task accuracy, matching the reported human expert baseline for this benchmark (64.0\%) and exceeding the best prior agent system AutoGen+GPT-4o (34.1\%) by 29.9 pp, Gemini-1.5 Pro (31.6\%) by 32.4 pp, GPT-4o (28.1\%) by 35.9 pp, and GPT-4 (26.0\%) by 38.0 pp. On \textbf{DSGym}, CoDaS scored 84.4\% overall accuracy, surpassing Kimi K2 (79.9\%) by 4.5 pp, Claude Sonnet 4.5 (78.2\%) by 6.2 pp, GPT-4o (78.0\%) by 6.4 pp, and DeepSeek v3.1 (71.2\%) by 13.2 pp.

\textbf{Scientific Hypothesis Generation.}
DiscoveryBench~\citep{majumder2024discoverybench} requires autonomous hypothesis generation from raw scientific datasets. CoDaS achieved HMS of 0.32, exceeding the oracle-assisted Reflexion agent (0.24) by 8.0~pp, attributable to domain-aware data profiling, iterative gap-based discovery, and question-type-adaptive analysis strategies.

\textbf{Clinical Reasoning.}
On HealthBench~\citep{arora2025healthbench} (5,000 physician-designed medical conversations), CoDaS scored 0.724 versus o3's 0.598 (+12.6~pp). On the Hard subset (cases resistant to all frontier models), CoDaS scored 0.391 versus 0.320 (+7.1~pp), with particularly strong performance on hedging under uncertainty (0.565), suggesting that the adversarial critic--defender architecture produces calibrated uncertainty.

\textbf{Data Science Code Generation.}
On DataSciBench~\citep{zhang2025datascibench} (222 end-to-end data science tasks), CoDaS achieved 77.5\% completion rate versus 68.4\% for GPT-4o (+9.1~pp), with uniform performance across task types (deep learning: 77.8\%, CSV/tabular: 77.6\%, human-authored: 77.4\%).

\textbf{Real-World Data Analysis.}
On DSBench~\citep{jing2024dsbench} (466 tasks from professional data competitions), CoDaS achieved 64.0\% task accuracy, matching the human expert baseline and exceeding the best prior agent system AutoGen+GPT-4o (34.1\%) by 29.9~pp.

\textbf{Quantitative and Causal Reasoning.}
On DSGym~\citep{nie2026dsgym} (QRData + DAEval; statistical, causal, and data analysis tasks), CoDaS scored 84.4\% overall accuracy, surpassing Kimi~K2 (79.9\%), Claude~Sonnet~4.5 (78.2\%), and GPT-4o (78.0\%).

\paragraph{Cross-benchmark patterns.}
Three patterns are relevant to the biomarker discovery setting. First, iterative refinement reliably outperforms single-pass generation, with margins exceeding 8~pp on every benchmark employing feedback-driven iteration. Second, the hybrid deterministic--agentic architecture avoids hallucination-prone numerical reasoning by executing computations in deterministic subprocesses while reserving LLM reasoning for interpretation and gap analysis. Third, domain-aware data profiling (automatic detection of replication structures, temporal layouts, categorical encodings) enables appropriate analytical strategy selection without manual specification.

\begin{figure}[!t]
    \centering
    \includegraphics[width=\textwidth]{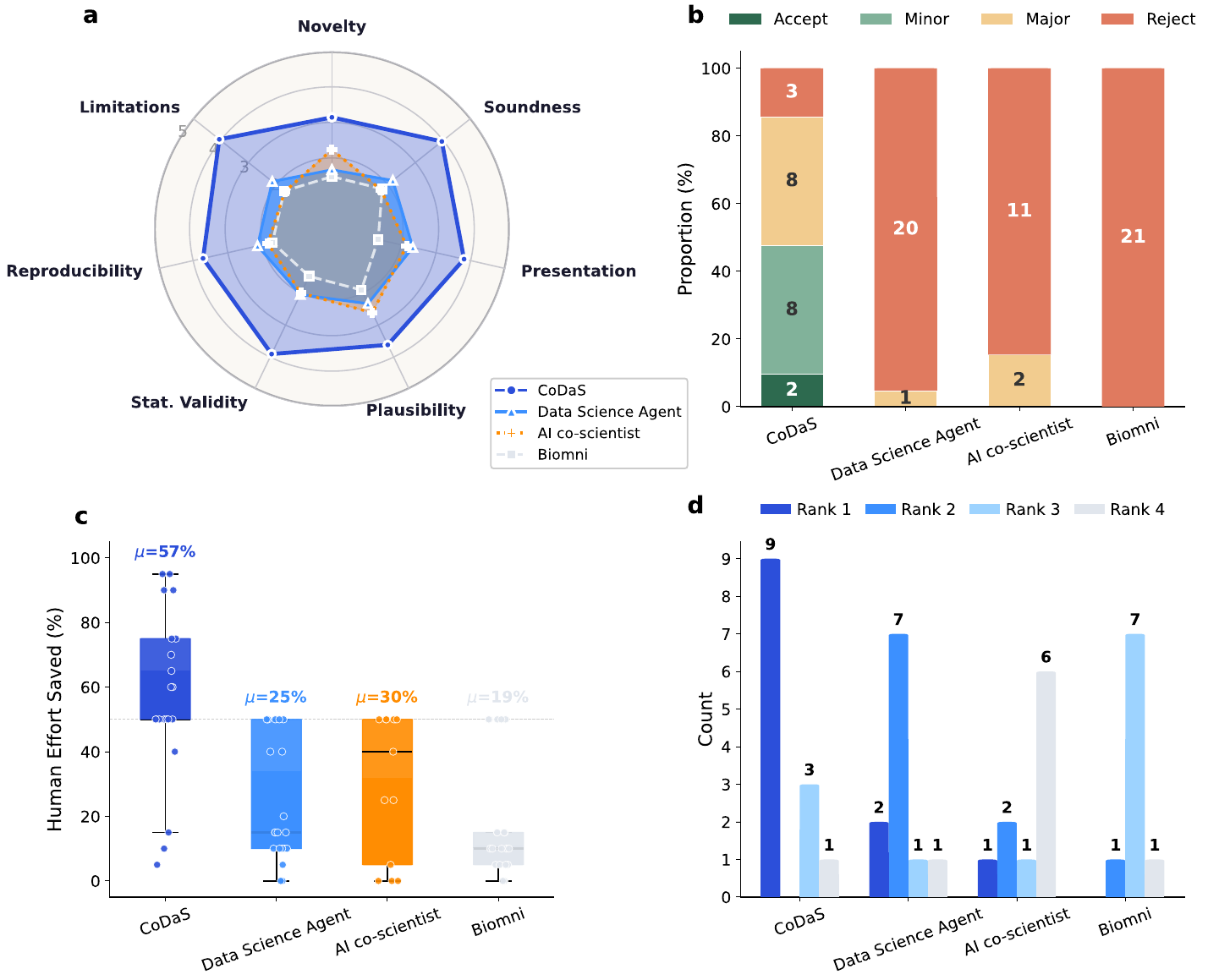}
    \caption{\textbf{Blinded Human Evaluation.}
    \textbf{(a)} Radar chart of mean expert scores (1 to 5 Likert scale) across seven quality dimensions. CoDaS achieves scores of 3.1 to 4.1, while all baselines cluster at 1.3 to 2.6. AI co-scientist is shown separately as an auxiliary unpaired comparison ($n = 13$), whereas CoDaS, Biomni, and Data Science Agent are compared within the 21-session balanced subset. \textbf{(b)} Simulated reviewer recommendation distribution. CoDaS received 2 Accept, 8 Minor Revision, 8 Major Revision, and 3 Reject (86\% non-rejection rate). AI co-scientist received 11 Reject and 2 Major Revision ($n = 13$). No baseline received Accept or Minor Revision.
    \textbf{(c)} Effort preservation scores (percentage of content reviewers would retain). CoDaS: $\mu = 56.9$\% vs.\ 18.8 to 30.4\% for baselines.
    \textbf{(d)} Forced ranking distribution ($n = 13$ sessions across all four systems). CoDaS was ranked \#1 in 9 of 13 sessions (69\%).}
    \label{fig:user_study}
\end{figure}

\section{User Study}
\label{sec:user_study}

To rigorously assess the quality of CoDaS research outputs against frontier AI-driven scientific discovery systems, we conducted a blinded expert evaluation.
Fifteen domain experts independently evaluated manuscripts produced by four systems: (i) CoDaS, (ii) Google's AI co-scientist~\citep{aicoscientist2024}, (iii) Biomni~\citep{biomni2024}, and (iv) Google ADK's Data Science Agent, across three wearable datasets, yielding 34 evaluation sessions and 76 individual manuscript assessments.
For CoDaS, Data Science Agent, and Biomni, each reviewer scored all three systems in a single session ($n = 21$ sessions); for AI co-scientist, reviewers scored the system in a separate set of sessions ($n = 13$) using the same instrument and datasets.

\subsection{Study Design}
\label{sec:study_design}

\paragraph{Blinded evaluation protocol.}
Each evaluator was presented with de-identified manuscripts (labeled Model~A through Model~D or Model~E) generated from the same input dataset and biomarker discovery task.
The mapping between system identities and blind labels was randomized independently for each session, ensuring that reviewers could not infer which system produced which output.
Evaluators were not informed of the number or identity of the systems under comparison. The evaluation was blinded to system identity, randomized model-label mapping, study hypotheses, and other reviewers’ assessments. Reviewers were told that they would evaluate blinded reports, but not which system generated each report.

\paragraph{Interface design.}
Evaluators interacted with a custom web-based review interface designed to emulate the workflow of a biomedical manuscript peer-review process. Each manuscript was rendered within a standardized reading interface that preserved the original structure of the generated report, including title, abstract, introduction, figures, tables, and methodological descriptions. To avoid presentation bias, all manuscripts were formatted using the same layout template and figure rendering pipeline. Reviewers examined one manuscript at a time through a scrollable document viewer and were allowed to navigate freely between sections before completing the evaluation form. The evaluation form was displayed in a structured panel adjacent to the manuscript viewer and included Likert-scale scoring fields, editorial decision options (Accept / Minor Revision / Major Revision / Reject), and free-text feedback fields for qualitative assessment. Representative screenshots of the evaluation interface are provided in Figure~\ref{fig:user_study_1} through \ref{fig:user_study_8} in the Appendix.

\paragraph{Assessment instrument.}
Our evaluation instrument was designed to mirror established peer review practices at top biomedical venues, comprising four components as follows.

\begin{enumerate}[leftmargin=*, itemsep=2pt]
    \item \textbf{Multi-axis quality assessment} (1 to 5 Likert scale) across seven dimensions: \emph{Novelty} (originality of biomarker hypotheses), \emph{Soundness} (methodological rigor), \emph{Presentation} (writing clarity and structure), \emph{Plausibility} (biological credibility of findings), \emph{Statistical Validity} (correctness of statistical analyses), \emph{Reproducibility} (sufficiency of methodological detail for replication), and \emph{Limitations} (acknowledgment of methodological constraints).
    \item \textbf{Reviewer recommendation}: Accept, Minor Revision, Major Revision, or Reject, using generic peer-review recommendation categories rather than categories specific to any one journal.
    \item \textbf{Effort preservation score} (0 to 100\%): the proportion of the manuscript a domain expert would retain in a revision, quantifying practical utility.
    \item \textbf{Safety and reliability audit}: reviewers flagged instances of hallucinated content, statistical errors, logical flaws, or biological contradictions, with mandatory free-text justification for each flag.
\end{enumerate}

Additionally, reviewers provided: (i)~a forced ranking of all systems from best to worst with free-text rationale; (ii)~estimates of the person-days each research phase would require if conducted manually (data preprocessing, feature engineering, ML modeling, literature review, result interpretation, and paper writing); and (iii)~the validation steps they would require before considering findings publishable.

\paragraph{Expert panel.}
The evaluation panel consisted of 15 domain experts with backgrounds in medicine, biomedical data science, machine learning, bioinformatics, and digital health research. Panelists had between 0 and 19 years of research experience (mean: 6.7 years) and had prior experience conducting peer-reviewed biomedical research or reviewing scientific manuscripts. Experts were recruited from both academic and industrial research environments. All reviewers were blinded to the identities of the models and the study hypotheses throughout the evaluation process.

% ============================================================================
\subsection{Results}
\label{sec:user_study_results}

% \paragraph{Expert ranking favored CoDaS.}
% CoDaS was ranked as the best system in 9 of 13 sessions where all four systems received valid rankings (69\% win rate; Figure~\ref{fig:user_study}d).
% No other system achieved a comparable win rate: Data Science Agent was ranked first in 2 sessions (15\%), AI co-scientist in 1 session (8\%), and Biomni in 0 sessions.

% Reviewers' free-text justifications for ranking CoDaS first converged on three recurring themes: \emph{statistical rigor}, \emph{scientific completeness}, and \emph{writing quality}.
% Representative quotes include:
% \begin{quote}
% \small
% ``\emph{Highest quality in every possible way, best analysis, most novel, most rigorous.}'' (R2) \\[2pt]
% ``\emph{Rigorous statistics; less AI feel compared to others; better formatting.}'' (R3) \\[2pt]
% ``\emph{Real data was analyzed, real statistics were computed with FDR correction and bootstrap CIs, real biomarkers were screened through an 11-step battery.}'' (R8) \\[2pt]
% ``\emph{The only scientifically valid and empirically complete manuscript in the group; rigorous, coherent, addresses statistical artifacts such as multicollinearity.}'' (R16) \\[2pt]
% ``\emph{Able to construct a coherent narrative, support its hypotheses, and discuss limitations.}'' (R13)
% \end{quote}
% These quotes represent the majority opinion; negative assessments of CoDaS, including 3 Reject decisions (14\% rejection rate) and 3 safety flags (Section~\ref{sec:safety}), are reported separately to avoid selection bias.

\paragraph{CoDaS was the only system to receive Accept/Minor Revision recommendations in blinded evaluation.}
Figure~\ref{fig:user_study}b presents the editorial decision distribution.
CoDaS received 2 Accept, 8 Minor Revision, 8 Major Revision, and 3 Reject decisions ($n = 21$), corresponding to an 86\% non-rejection rate (18 of 21 assessments).
In contrast, Biomni received 21/21 Reject decisions, AI co-scientist received 11 Reject and 2 Major Revision ($n = 13$), and Data Science Agent received 20 Reject and 1 Major Revision.
No baseline system received an Accept or Minor Revision decision from any reviewer.
Fisher's exact tests confirmed the significance of these differences: CoDaS versus each baseline for the non-rejection rate (18/21 vs.\ 0/21 for Biomni, OR $= \infty$, $p_\mathrm{adj} = 2.3 \times 10^{-8}$; 18/21 vs.\ 2/13 for AI co-scientist, OR $= 33.0$, $p = 7.7 \times 10^{-5}$; 18/21 vs.\ 1/21 for Data Science Agent, OR $= 120.0$, $p_\mathrm{adj} = 3.8 \times 10^{-7}$).

On the WEAR-ME dataset, where CoDaS's hold-out validation pipeline was most mature, the non-rejection rate reached \textbf{100\%} (2 Accept, 3 Minor, 3 Major; 0 Reject), underscoring the relationship between pipeline completeness and perceived manuscript quality.

\paragraph{CoDaS scored higher than all baselines across quality dimensions.}
Figure~\ref{fig:user_study}a presents the multi-axis radar comparison.
CoDaS achieved a mean overall score of 3.74 (95\% CI: 3.42 to 4.06) across all seven assessment dimensions ($n = 21$), compared to 2.12 (1.85 to 2.39) for Data Science Agent ($n = 21$), 2.04 for AI co-scientist ($n = 13$), and 1.63 (1.45 to 1.81) for Biomni ($n = 21$).
Paired Wilcoxon signed-rank tests (matched within each evaluation session) confirmed statistically significant advantages over Data Science Agent and Biomni on the composite score: $\Delta = +2.11$ vs.\ Biomni (Cohen's $d = 2.40$, $p_\mathrm{adj} = 0.001$) and $\Delta = +1.62$ vs.\ Data Science Agent ($d = 1.64$, $p_\mathrm{adj} = 0.001$); all $p$-values Bonferroni-corrected; $n = 21$ paired assessments per system.
A Mann-Whitney $U$ test (unpaired, given the different sample sizes) confirmed that CoDaS also scored significantly higher than AI co-scientist ($U = 265.5$, $p = 5.0 \times 10^{-6}$, Cohen's $d = 2.49$), with all seven individual axes reaching significance at $p < 0.003$.
The strongest separations were on Limitations ($d = 3.16$) and Soundness ($d = 2.81$).

CoDaS's advantage was most pronounced on Limitations acknowledgment (4.05 vs.\ 1.69 to 2.14 for baselines), Soundness (3.95 vs.\ 1.77 to 2.19), and Statistical Validity (3.90 vs.\ 1.48 to 2.05), reflecting its multi-stage validation pipeline.
This pattern held consistently across all three datasets.
On DWB Hourly ($n = 9$), CoDaS scored a mean of 3.76 versus 1.71 to 2.02 for baselines; on GLOBEM ($n = 4$), 3.61 versus 1.64 to 2.21; and on WEAR-ME ($n = 8$), 3.79 versus 1.54 to 2.20.
Baseline systems rarely exceeded a mean score of 2.5 on any individual axis, with isolated exceptions on Novelty and Presentation for specific dataset and system combinations.

Reviewers' free-text feedback on AI co-scientist highlighted persistent methodological gaps: ``\emph{multiple testing correction was not performed; text and figure has AI feel}'' (R3), ``\emph{just suggestions of ideas}'' (R4), ``\emph{heavy on new hypothesis, but very weak on analysis to test and prove them}'' (R15), and ``\emph{some of the figures look hallucinated}'' (R4).

\paragraph{Effort preservation quantifies practical utility.}
The effort preservation score measures how much of a generated manuscript a domain expert would retain in a revision (Figure~\ref{fig:user_study}c).
CoDaS achieved a mean effort score of 56.9\% (95\% CI: 45.2 to 68.6\%; range: 5 to 95\%), indicating that reviewers would, on average, keep over half of the generated content as a starting point for their own work.
Baseline systems scored markedly lower: 18.8\% (Biomni), 24.5\% (Data Science Agent), and 30.4\% (AI co-scientist, $n = 13$), meaning fewer than one-third of baseline outputs were judged salvageable.
Paired Wilcoxon signed-rank tests confirmed significant effort advantages over the paired baselines: $\Delta = +38.1\%$ vs.\ Biomni ($d = 1.11$, $p_\mathrm{adj} = 0.003$) and $\Delta = +32.4\%$ vs.\ Data Science Agent ($d = 0.85$, $p_\mathrm{adj} = 0.007$).
Effort preservation for CoDaS was also significantly higher than for AI co-scientist (Mann-Whitney $p = 0.003$, $d = 1.08$).
Multiple reviewers assigned CoDaS effort scores of 90 to 95\%, suggesting review-ready quality in some instances.
One reviewer (R13) characterized CoDaS output as comparable to ``\emph{a first-year PhD student, if I was able to have discussions with them, I think we could iterate on this original research.}''

Accordingly, the user study should be interpreted primarily as evidence of strong relative preference among systems rather than high absolute agreement in the assigned scores.

% \begin{figure}[ht!]
%     \centering
%     \includegraphics[width=0.75\textwidth]{imgs/fig7_safety_flags.pdf}
%     \caption{\textbf{Safety and reliability concerns identified by expert reviewers.}
%     Count of safety flags per system across five categories.
%     CoDaS received 3 flags (1 StatError, 1 Hallucination, 1 Other; see text for details); baseline systems accumulated 51 flags total (42 from Data Science Agent and Biomni, 9 from AI co-scientist).
%     The most severe failure modes, including hallucinated results from failed API calls, sub-random classification performance presented as findings, and label leakage invalidating primary claims, were observed exclusively in baseline systems.}
%     \label{fig:safety_flags}
% \end{figure}

\begin{table}[ht!]
\centering
\footnotesize
\caption{\textbf{Safety and reliability concerns identified by expert reviewers.} Count of safety flags per system across five categories. CoDaS received 3 flags (1 StatError, 1 Hallucination, 1 Other); baseline systems accumulated 51 flags total (42 from Data Science Agent and Biomni, 9 from AI co-scientist). The most severe failure modes, including hallucinated results, sub-random classification performance presented as findings, and label leakage invalidating primary claims, were observed exclusively in baseline systems.}
\label{tab:safety_flags}
\begin{tabular}{lccccc}
\hline
 & Hallucination ($\downarrow$) & StatError ($\downarrow$) & Logic ($\downarrow$) & BioContradiction ($\downarrow$) & Other ($\downarrow$) \\
\hline
CoDaS (\textbf{Ours}) & 1 & 1 & 0 & 0 & 1 \\
Data Science Agent & 1 & 5 & 5 & 4 & 1 \\
AI Co-scientist & 5 & 2 & 1 & 0 & 1 \\
Biomni & 7 & 6 & 7 & 2 & 4 \\
\hline
\end{tabular}
\end{table}

% ============================================================================
\subsection{Safety and Reliability Analysis}
\label{sec:safety}

A critical dimension for clinical and biomedical applications is minimizing hallucinated or erroneous content that could propagate to downstream research \citep{kim2025medical}.
Reviewers flagged a total of 51 safety concerns across the three baseline systems, compared to 3 for CoDaS (Table~\ref{tab:safety_flags}).
The three CoDaS flags were: one statistical error (a confidence interval whose bounds did not contain the reported point estimate), one hallucination flag (incorrect citations in references), and one other issue (a LaTeX truncation artifact). While reviewers categorized the first two under StatError and Hallucination, respectively, manual inspection confirmed these were isolated reporting errors rather than systematic data fabrication or flawed analytical reasoning.
AI co-scientist received 9 safety flags ($n = 13$), with hallucination as the primary concern (5 flags; 0.69 flags per session).
Among the other baseline systems, the most prevalent safety concerns were:

\begin{itemize}[leftmargin=*, itemsep=2pt]
    \item \textbf{Hallucination} (8 flags for Data Science Agent and Biomni combined): fabricated images, invented data, and false claims of successful execution. Biomni ``\emph{hallucinated a successful literature review outcome from failed API calls}'' (R2) and produced manuscripts that ``\emph{claim strong biomarker candidates while simultaneously printing that those candidates are non-existent and the tables are empty}'' (R16).
    \item \textbf{Logic errors} (12 flags): contradictory conclusions, incoherent reasoning, and broken analytical workflows. Biomni was ``\emph{congratulating itself for work it explicitly failed to produce}'' (R2). Data Science Agent exhibited ``\emph{fatal statistical errors: running 159 tests with $p < 0.05$ without multiple comparison correction fundamentally invalidates the primary claims}'' (R7).
    \item \textbf{Statistical errors} (11 flags): data leakage, missing corrections, and sub-random performance. Data Science Agent committed ``\emph{a textbook example of definitional label leakage, invalidating the primary claim}'' (R2) and reported ``\emph{AUC-ROC below 0.5, implying the model is predicting the wrong class}'' (R2). Multiple reviewers noted that Biomni produced ``\emph{an AUC-ROC of 0.517, statistically indistinguishable from random noise}'' (R2) and failed to apply ``\emph{imputation before train/test split, introducing data leakage}'' (R7).
    \item \textbf{Biological contradictions} (6 flags): claims inconsistent with established domain knowledge. Data Science Agent ``\emph{ranked sodium and chloride as top biomarkers for insulin resistance, primary electrolytes with no established role as direct IR biomarkers}'' (R8), while ``\emph{discovering triglycerides as the top predictor for metabolic syndrome is a textbook example of definitional label leakage}'' (R2).
\end{itemize}

The lower number of safety flags for CoDaS may reflect the contribution of its multi-stage validation architecture, although the causal contribution of individual safeguards was not isolated in this study.

\begin{figure}[t!]
    \centering
    \includegraphics[width=0.75\textwidth]{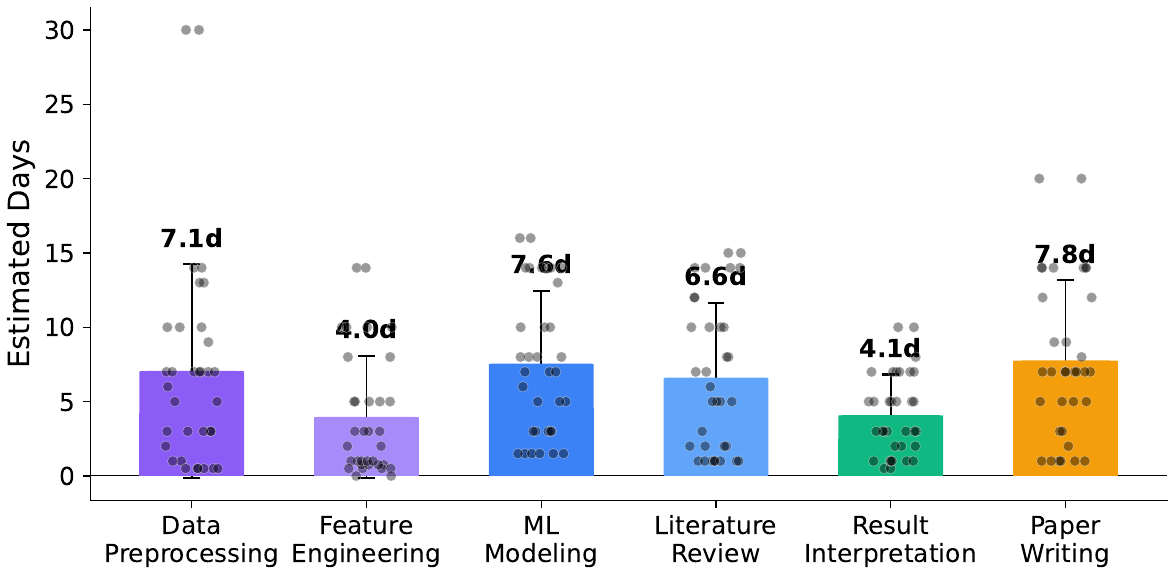}
    \caption{\textbf{Estimated human effort to manually reproduce the equivalent research workflow.}
    Expert estimates of person-days per research phase ($n = 34$ responses).
    Mean total: 37 days (median: 40; range: 5 to 90 days).
    Paper writing (7.8d), ML modeling (7.6d), and data preprocessing (7.1d) were the most time-intensive phases.
    CoDaS reduces end-to-end wall-clock time for an automated discovery run to 6–8.5 hours on a single machine. This should be interpreted in contrast to reviewers’ estimates of manual human effort (mean: 37 person-days), not as a direct productivity ratio.
    Error bars: standard deviation; grey markers: individual estimates.}
    \label{fig:time_breakdown}
\end{figure}

% ============================================================================
\subsection{Human Effort Estimation}
\label{sec:time_savings}

To contextualize the practical impact of end-to-end automation, we asked each reviewer to estimate the person-days required to manually conduct the equivalent research workflow.
Across all 34 responses, the mean estimated total effort was 37 $\pm$ 23 person-days (median: 40; range: 5 to 90 days), broken down by research phase in Figure~\ref{fig:time_breakdown}. The most time-intensive phases were paper writing (mean: 7.8 days), ML modeling (7.6 days), and data preprocessing (7.1 days).
Literature review (6.6 days), result interpretation (4.1 days), and feature engineering (4.0 days) constituted the remaining effort.
CoDaS automates all six phases end-to-end, with a typical wall-clock runtime of 6 to 8.5 hours per dataset on a single machine (a detailed phase-wise runtime comparison across all four systems is provided in Table~\ref{tab:runtime_comparison} in the Appendix).
This comparison contrasts estimated human labor with automated wall-clock runtime and is intended as an order-of-magnitude illustration rather than a direct productivity ratio. Reviewers judged the resulting outputs substantively usable (57\% effort preservation). Reviewers also indicated the validation steps they would require before considering AI-generated findings publishable:
external validation on independent cohorts (85\%), held-out confirmation of key findings (62\%), code review of the analytical pipeline (47\%), and wet-lab experimental confirmation (26\%).
Notably, 15\% of respondents selected ``never publishable without human involvement,'' reflecting healthy skepticism about fully autonomous scientific pipelines.
These responses position CoDaS outputs not as finished publications, but as high-fidelity first drafts that substantially accelerate the hypothesis-to-validation cycle.

\begin{figure}[t!]
    \centering
    \includegraphics[width=\textwidth]{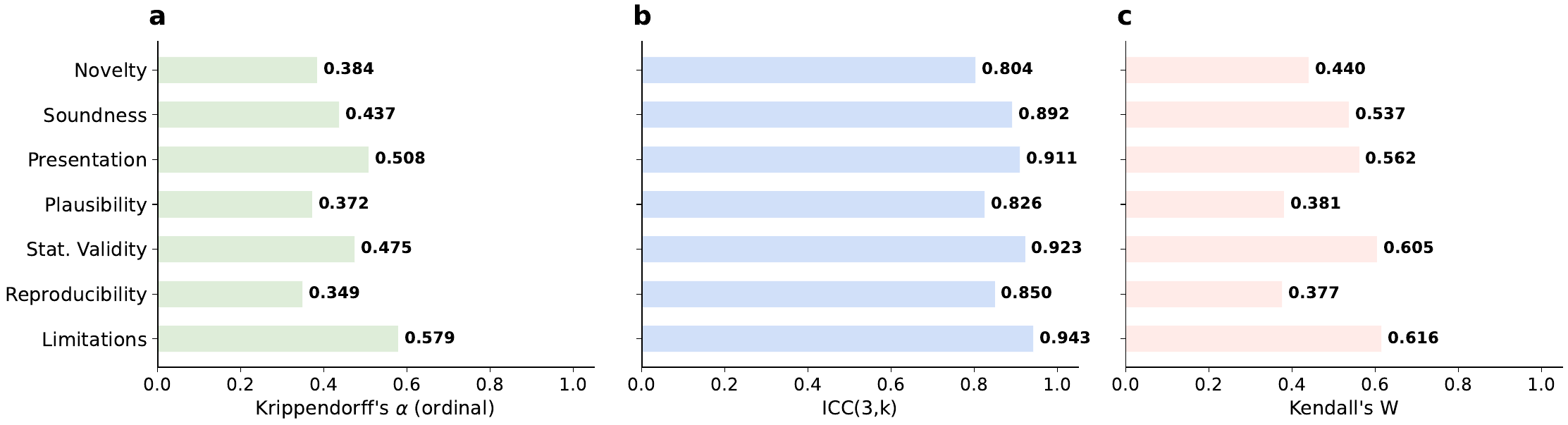}
    \caption{\textbf{Inter-rater reliability across seven assessment axes.}
    \textbf{(a)} Krippendorff's alpha (ordinal) measures absolute agreement accounting for chance and missing data; overall $\alpha = 0.443$ (low-to-moderate agreement).
    \textbf{(b)} Intraclass correlation coefficient ICC(3,k) measures consistency of relative system rankings; overall ICC $= 0.888$ (excellent reliability), with all axes significant at $p < 0.0001$.
    \textbf{(c)} Kendall's coefficient of concordance $W$ measures ordinal agreement among raters within datasets; overall $W = 0.503$ (moderate concordance).
    Dashed lines indicate conventional thresholds.
    The divergence between $\alpha$ (moderate) and ICC (excellent) indicates that reviewers differed in absolute calibration but strongly agreed on relative system quality.}
    \label{fig:irr_summary}
\end{figure}

\paragraph{Qualitative feedback and identified limitations.}
Beyond numerical scores, reviewers provided constructive feedback identifying areas for improvement.
Several reviewers noted that CoDaS ``\emph{still lacks a modern ML/DL pipeline}'' and relies primarily on ``\emph{classical ML methods}'' for feature selection (R1), which, while appropriate for the current sample sizes, may limit scalability to larger cohorts.
Another reviewer observed that ``\emph{writing is still not satisfying}'' despite being the best among all systems (R1), and one expressed skepticism: ``\emph{very suspicious of it and still would never accept it}'' (R6), noting concerns about the volume of automated validation.
Multiple reviewers suggested incorporating ``\emph{human evaluation at intermediate stages to enable collaboration}'' rather than fully autonomous end-to-end generation (R4).
These critiques, combined with the 14\% rejection rate and 3 safety flags, underscore that while CoDaS substantially outperforms existing baselines, its outputs require expert review and refinement before publication, a positioning we explicitly adopt in our system design.

\subsection{Statistical Analysis}
\label{sec:stat_analysis}

\paragraph{Inter-rater reliability.}
To quantify the degree to which independent reviewers produced consistent assessments, we computed four complementary inter-rater reliability (IRR) metrics across all seven evaluation axes using a balanced subset of sessions where all reviewers scored the same set of systems ($n = 21$ sessions, 3 systems; Figure~\ref{fig:irr_summary}), which provides the multi-rater $\times$ multi-item design required for valid IRR computation.
Using \emph{Krippendorff's alpha} with ordinal weighting, the most conservative metric appropriate for Likert-scale data with missing raters, we obtained an overall $\alpha = 0.443$.
Agreement was strongest for Limitations ($\alpha = 0.579$), Presentation ($\alpha = 0.508$), and Statistical Validity ($\alpha = 0.475$), and weakest for Reproducibility ($\alpha = 0.349$) and Plausibility ($\alpha = 0.372$), consistent with the expectation that subjective dimensions elicit greater rater heterogeneity.
Across datasets, agreement was comparable: DWB Hourly ($\alpha = 0.462$), GLOBEM ($\alpha = 0.473$), and WEAR-ME ($\alpha = 0.431$), indicating no dataset-specific calibration bias.

\textbf{Intraclass correlation coefficients}: ICC(3,k), two-way mixed, consistency, average measures yielded substantially higher reliability estimates: overall ICC $= 0.888$ (excellent; $> 0.75$ threshold), with all seven axes reaching ICC $\geq 0.804$ and six of seven exceeding $0.85$ (all $p < 0.0001$).
The highest ICCs were observed for Limitations (0.943), Statistical Validity (0.923), and Presentation (0.911).
The divergence between Krippendorff's $\alpha$ and ICC is expected: $\alpha$ penalizes systematic rater bias (some reviewers consistently score higher or lower), while ICC(3,k) measures consistency of relative rankings, indicating that although reviewers differed in absolute calibration, they agreed strongly on \emph{which systems were better or worse}.

\textbf{Kendall's coefficient of concordance (W)} confirmed moderate-to-strong concordance among raters in their ordinal rankings of systems within each dataset: overall $W = 0.503$, with Statistical Validity ($W = 0.605$) and Limitations ($W = 0.616$) showing the strongest agreement.

\textbf{Pairwise Spearman correlations} across the 60 unique reviewer pairs with $\geq$3 common items yielded a mean $\rho = 0.646$ (median: 0.632; range: 0.200 to 1.000), with 13\% of pairs reaching statistical significance ($p < 0.05$) despite the limited number of shared items per pair (typically 4).

Taken together, these metrics indicate that despite heterogeneous backgrounds (0 to 19 years experience, 8 institutions, occasional-to-heavy review frequency), our panel exhibited \emph{low-to-moderate absolute agreement} (below the $\alpha = 0.667$ threshold recommended by Krippendorff for tentative conclusions) and \emph{excellent relative consistency}.

\section{Clinician Assessment of Translational Relevance}
\label{sec:clinical_relevance}

Internal statistical screening can identify associations that are stable within the available data, but it does not establish clinical utility or readiness for patient management. To assess these properties for the candidates in Table~\ref{tab:biomarker_summary}, we convened a panel of twelve physicians and physicians-in-training involved in clinical care (median 5 years of clinical experience after medical qualification, range 0 to 27) and asked them to evaluate every candidate against a structured rubric interrogating the translational relevance. The panel spanned psychiatry, internal medicine, family medicine, endocrinology, geriatrics, pain medicine, radiology, and general practice, and was drawn from academic medical centers and clinical research institutions in the United States and the Republic of Korea. Each physician independently reviewed all 16 candidate associations together with the pipeline-rejected TG/HDL ratio, which served as a calibration control, giving 17 candidates reviewed per rater and 204 reviews in total.

\subsection{Evaluation}
\label{sec:clinical_protocol}

\paragraph{Protocol.}
Each candidate was presented on a single standardized interface which stated the input feature, the clinical endpoint, the reported effect size in the discovery cohort, and a translation of that effect size into real-world units, expressed as the expected change in the endpoint across the interquartile range and per standard deviation. For example, for main sleep duration variability a discovery-cohort correlation with the PHQ-8 depression score (Spearman $\rho = 0.252$) was accompanied by an empirical distributional translation showing the observed PHQ-8 difference across the feature interquartile range. The demographic covariates available in the cohort model were listed for context. The interface reported derived population-level statistics only and contained no participant-level data.

\paragraph{Rating instrument.}
For every task, reviewers rated seven dimensions on a five-point scale. The dimensions were validity (confidence that the association is genuine rather than an artifact), effect-size meaningfulness, novelty relative to current knowledge, feasibility of measurement, added value over biomarkers already in use, likelihood of influencing patient advice, and confidence to act on the result for an individual patient. These seven dimensions were selected to span the translational properties most relevant to whether a biomarker would be adopted in practice, rather than adapted from a single existing scale. Reviewers also classified the supporting literature into one of several ordered categories ranging from strongly supportive to contradicted, and provided written justification for the validity, effect-size, and added-value ratings. The task order was randomized for each reviewer to limit order effects.

\paragraph{Calibration probe.}
The rejected TG/HDL ratio was presented alongside the genuine candidates as a sanity check. Because this ratio is an established lipid-derived metabolic-risk index already available from routine laboratory testing, a well-calibrated panel should rate its validity high and its added value low. A failure to separate these two judgments would indicate that reviewers were responding to surface plausibility rather than to incremental clinical value.

\paragraph{Reviewer effort.}
The panel completed the tasks asynchronously through a web interface that recorded a timestamp each time each task was opened and submitted. Each reviewer actively rated for a median of just over two hours (range 1.0 to 3.3 hours), for a panel total of approximately 25 hours. 

% The figure is an approximation because the interface did not support window focus directly.

\subsection{Results}
\label{sec:clinical_results}

\begin{figure}[t!]
    \centering
    \includegraphics[width=\textwidth]{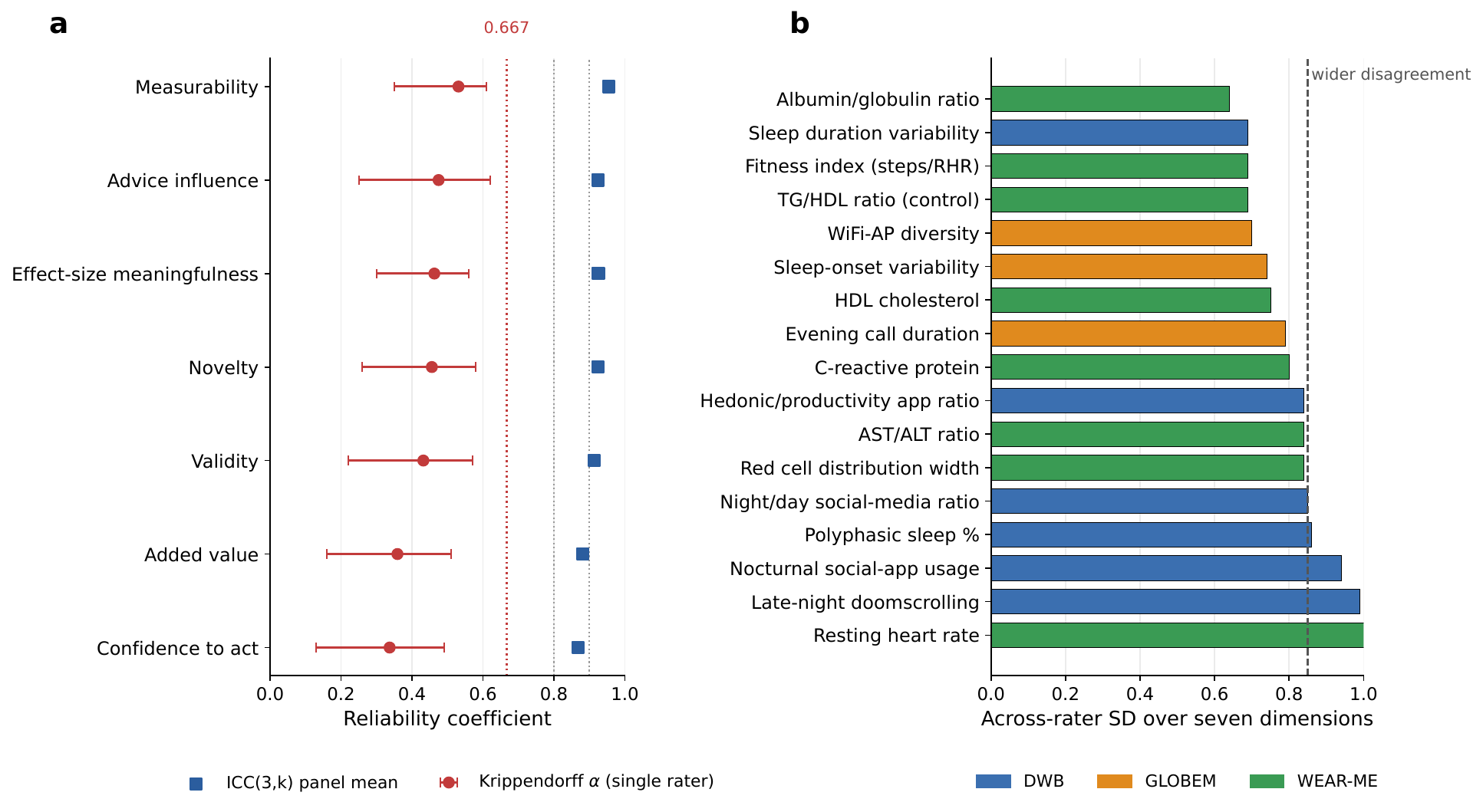}
    \caption{\textbf{Clinician review distinguishes validity from clinical actionability.}
    \textbf{(a)} Reliability of the seven rated dimensions. Red markers show Krippendorff's ordinal alpha for a single rater with 95\% bootstrap confidence intervals from 3,000 resamples, which remain below the 0.667 threshold for tentative agreement on every dimension. Blue markers show the two-way mixed-effects ICC(3,k) for consistency, the same form used in Section~\ref{sec:stat_analysis}, for the twelve-clinician mean, which reaches the good to excellent range (0.87 to 0.95). The separation between the two reflects differences in absolute scale use alongside agreement on relative ranking.
    \textbf{(b)} Across-rater standard deviation over the seven dimensions for each candidate, colored by cohort. Lower values indicate closer agreement. Candidates to the right of the dashed line fall in the upper tercile of disagreement.}
    \label{fig:clinical_panel}
\end{figure}

\paragraph{Physician review aligned with CoDaS prioritization and identified clinical utility gaps.}
Averaged over all candidates and reviewers, validity received the highest rating (mean 3.69 on the five-point scale), indicating that the panel considered the associations more likely genuine than artifactual. Practical measurability was rated moderate (3.28). The three dimensions tied most directly to clinical action were rated lower, with added value over existing biomarkers at 2.37, confidence to act for an individual patient at 2.34, and likelihood of influencing advice at 2.40. These ratings indicate that the panel distinguished evidence for genuine association from evidence for added clinical value, supporting the use of CoDaS as a candidate-prioritization system whose outputs require follow-up clinical validation before use in patient management. This reading is consistent with the evidence tiers in Table~\ref{tab:biomarker_summary}, where most candidates are labeled Supported or Emerging rather than Established.

\paragraph{Panel judgment converged with the pipeline's confidence ranking and independently identified the candidate whose statistics did not replicate.}
Validity ratings were highest for candidates in the Established tier (mean 4.2) and lower for the newer Supported and Emerging tiers (means about 3.8 and 3.3), and the rating correlated with both the tier CoDaS had assigned (Spearman $\rho = 0.67$, $p = 0.005$) and the reported effect size ($\rho = 0.77$, $p = 0.001$). Part of this agreement reflects representations in routine clinical documentation, since the Established tier consists of quantities that clinicians already encounter in the electronic medical record (EMR) system, such as laboratory markers (HDL and C-reactive protein) and vital signs (resting heart rate). The most informative case was the candidate the pipeline had itself flagged as unstable. Evening incoming call duration, flagged for a reversal of the effect direction between the discovery and held-out estimates, was rated the least valid candidate of all (1.75), and reviewers across specialties independently identified the same reversal. One reviewer observed that \emph{a stable biological signal cannot reverse direction across partitions of the same cohort}. Without access to the tier labels, the panel corroborated the candidates the pipeline ranked highest and isolated the one whose statistical support did not replicate.

\paragraph{Agreement was strongest on the clearest accept and reject decisions.}
Table~\ref{tab:clinical_panel} and Figure~\ref{fig:clinical_panel}b report agreement for each candidate, summarized as the across-rater standard deviation over the seven dimensions. The highest validity ratings went to the established metabolic markers in WEAR-ME, with HDL cholesterol and C-reactive protein rated most clearly genuine (validity 4.75 and 4.33). Agreement was closest on candidates with an unambiguous standing, whether positive, such as the albumin/globulin ratio and the cardiovascular fitness index, or negative, such as evening incoming call duration in GLOBEM, which received the lowest validity (1.75) rating in the set. Disagreement was widest on resting heart rate and on the behavioral DWB candidates such as late-night doomscrolling and nocturnal social-app usage. The highest-rated DWB candidate, main sleep duration variability, drew a high but divided validity rating (4.17 with a standard deviation of 1.11). Eleven of the twelve reviewers rated its validity at 4 or 5, while one rated it 1 on the grounds that the sleep feature overlaps the sleep item of the PHQ-8 outcome, a form of criterion contamination that the effect size alone would not reveal.

\paragraph{Inter-rater reliability matched the pattern seen in the system-level evaluation.}
We quantified agreement with the same battery used in Section~\ref{sec:stat_analysis} and report it in Figure~\ref{fig:clinical_panel}a. Single-rater absolute agreement, measured by Krippendorff's ordinal alpha, ranged from 0.34 to 0.53 across the seven dimensions and remained below the 0.667 threshold for tentative conclusions on every dimension. Agreement was highest for measurability (alpha 0.53, 95\% CI 0.35 to 0.61) and lowest for added value (0.36, 95\% CI 0.16 to 0.51) and confidence to act (0.34, 95\% CI 0.13 to 0.49). The reliability of the twelve-clinician mean, measured by a two-way mixed-effects intraclass correlation coefficient for consistency, ICC(3,k), the same form reported for the system-level panel in Section~\ref{sec:stat_analysis}, ranged from 0.87 to 0.95 and met the conventional standard for good to excellent reliability. As in the system-level study, the gap between low single-rater alpha and high panel-mean ICC indicates that clinicians differed in their absolute use of the scale while agreeing on the relative standing of the candidates. We therefore treat the panel mean, rather than any individual rating, as the unit of analysis. Agreement on the categorical literature item was slight (Fleiss' kappa 0.16), consistent with reviewers drawing on different bodies of evidence. Per-dimension reliability statistics, including the single-rater ICC(3,1), are reported in the Appendix (Table~\ref{tab:clinical_reliability}).

\paragraph{The calibration probe behaved as designed.}
The rejected TG/HDL control received a validity rating of 4.83, above the mean of 3.95 for the genuine WEAR-ME candidates, and an added-value rating of 1.50, below the genuine WEAR-ME candidate mean of 2.46. Reviewers recognized TG/HDL as a quantity that is real but clinically redundant for incremental biomarker discovery, which supports the interpretation that the panel separated statistical validity from incremental clinical value rather than rating candidates favorably by default.

\begin{table}[t!]
\centering
\footnotesize
\caption{\textbf{Clinician panel ratings for each candidate biomarker.} Twelve physicians rated every candidate on a five-point scale. Validity is the mean across reviewers with the standard deviation in parentheses. Overall is the mean across the seven rated dimensions. Across-rater standard deviation is the standard deviation among reviewers averaged over the seven dimensions, where a lower value indicates closer agreement. The rejected TG/HDL ratio is included as a positive control.}
\label{tab:clinical_panel}
\begin{tabular}{p{2.0cm} p{6.0cm} c c c}
\toprule
\textbf{Cohort} & \textbf{Candidate} & \textbf{Validity} & \textbf{Overall} & \textbf{Across-rater s.d.} \\
\midrule
\multirow{6}{2.0cm}{\textbf{DWB} \newline (PHQ-8)}
 & Main sleep duration variability & 4.17 (1.11) & 3.15 & 0.69 \\
 & Nocturnal social app usage & 3.92 (1.08) & 3.06 & 0.94 \\
 & Late-night doomscrolling & 3.50 (1.00) & 2.73 & 0.99 \\
 & Night-to-day social media ratio & 3.67 (0.89) & 2.70 & 0.85 \\
 & Hedonic-to-productivity app ratio & 3.17 (1.03) & 2.31 & 0.84 \\
 & Polyphasic sleep percentage & 3.67 (0.78) & 2.69 & 0.86 \\
\midrule
\multirow{3}{2.0cm}{\textbf{GLOBEM} \newline (PHQ-4)}
 & Sleep onset time variability & 3.83 (0.58) & 2.55 & 0.74 \\
 & Evening incoming call duration & 1.75 (1.14) & 1.80 & 0.79 \\
 & WiFi AP sequential diversity & 2.58 (0.90) & 2.14 & 0.70 \\
\midrule
\multirow{8}{2.0cm}{\textbf{WEAR-ME} \newline (HOMA-IR)}
 & HDL cholesterol & 4.75 (0.45) & 2.96 & 0.75 \\
 & C-reactive protein (CRP) & 4.33 (0.49) & 3.06 & 0.80 \\
 & Derived AST/ALT ratio (De Ritis) & 4.00 (0.60) & 3.26 & 0.84 \\
 & Resting heart rate (mean) & 4.08 (1.08) & 3.20 & 1.01 \\
 & Cardiovascular fitness index (steps/resting HR) & 4.25 (0.45) & 3.55 & 0.69 \\
 & Red cell distribution width (RDW) & 2.92 (1.24) & 2.52 & 0.84 \\
 & Albumin/globulin ratio & 3.33 (0.78) & 2.69 & 0.64 \\
\cdashline{2-5}
 & Derived TG/HDL ratio (rejected control) & 4.83 (0.39) & 2.88 & 0.69 \\
\bottomrule
\end{tabular}
\end{table}

Taken together, these results position the clinician panel as an external check that both corroborated the pipeline and constrained its claims. Expert validity judgments aligned with the pipeline's own confidence ranking, the metabolic candidates were judged closest to clinical use, and the panel rated lowest the candidate whose effect did not replicate. The low ratings on added value and confidence to act indicate that most candidates are best regarded as hypotheses for prospective evaluation rather than biomarkers ready for clinical use. These conclusions are bounded by the composition of the panel, in particular the single psychiatric specialist among the reviewers of the behavioral cohorts, and by the use of a rubric built for this study rather than a previously validated instrument.

\FloatBarrier

\section{Limitations}

\paragraph{Exploratory, non-preregistered design.}
All analyses reported in this study are exploratory. No endpoints, biomarker candidates, subgroup analyses, or analysis strategies were registered in a public trial registry prior to pipeline execution. The GLOBEM endpoint (PHQ-4) was selected by the pipeline itself based on target coverage rather than pre-specified by investigators. Consequently, all reported candidates should be interpreted as statistically prioritized hypotheses, not validated biomarkers in the regulatory sense. The term ``screened'' as used throughout refers exclusively to survival of the internal validation battery and does not imply prospective clinical validation, external replication, or regulatory endorsement.

\paragraph{GLOBEM repeated-measures structure and missingness.}
The GLOBEM cohort comprises 704 participant-wave observations from 497 unique individuals; some participants contributed data across multiple annual waves. Although the holdout confirmation split and validation tests operate on one randomly selected observation per participant to ensure statistical independence (see Section~\ref{sec:validation_integrity}), the main analytical pipeline processes all 704 observations. Feature-level missingness in GLOBEM is substantial (54.6\%), reflecting the inherent sparsity of passively collected mobile sensing data. While features with $>$70\% missingness were dropped and remaining values were median-imputed, the impact of this imputation strategy on discovered associations has not been exhaustively characterized. Results from this cohort should be interpreted with these caveats.

\paragraph{Static labels and narrow disease scope.}
The DWB and WEAR-ME analyses rely on fixed or cross-sectional endpoints, while GLOBEM provides repeated survey measurements but with sparse sensing and coarse PHQ-4 labels. The mental-health analyses rely on self-report symptom scales, whereas the metabolic analysis is anchored to fasting laboratory measures. Cohort demographics further limit generalizability: participants skew White, female, and young to middle-aged in DWB and WEAR-ME, and GLOBEM is an undergraduate cohort (DWB: 84.9\% White, 70.0\% female; GLOBEM: 57.4\% Asian, undergraduate only; WEAR-ME: 77.7\% White/Caucasian in the source study). Prospective studies with repeated clinical labels, ecological momentary assessment, broader disease coverage, and more representative recruitment will be required before these candidates can be assessed for clinical utility across populations.

\paragraph{Associative framing and causal inference.}
CoDaS surfaces statistically robust associations and generates post-hoc mechanistic hypotheses grounded in retrieved literature, but does not establish causality. 
Statistical significance and robustness across subgroups or independent cohorts are insufficient to distinguish causal biomarkers from correlated proxies especially because only measured confounders could be adjusted for. Furthermore, the short retrospective monitoring windows are not sufficient for causal discovery of biomarkers. Consequently, candidate biomarkers identified by CoDaS should be regarded as hypothesis-generating until supported by physiological evidence, prospective validation, and rigorous causal inference frameworks. Moving forward, these capabilities could be further integrated into the adversarial validation process to systematically identify potential confounding factors, spurious correlations, and limitations in study design. In addition, incorporating target-trial emulation, structural causal models, or quasi-experimental designs into the validation gauntlet is a necessary step toward the evidentiary standards required for clinical decision support.

\paragraph{External validation and replication.}
No single biomarker was replicated in an independent cohort with an aligned outcome instrument. The two depression cohorts use different data structures (hourly wearable metrics vs.\ weekly passive-sensing aggregates) and different outcome instruments (PHQ-8 vs.\ PHQ-4), limiting comparability beyond construct-level convergence (see Section~\ref{sec:construct_convergence}). The WEAR-ME cohort addresses a separate disease domain (metabolic risk), demonstrating pipeline breadth but not direct cross-cohort replication. This evaluation design does not fulfill the external-validation requirements of TRIPOD or STARD reporting guidelines. Prospective replication of the top candidate biomarkers in an independent depression cohort using PHQ-8 as the primary endpoint, ideally with a temporally held-out validation set, is the clear next step for translational readiness.

\paragraph{Fixed agent topology and extensibility.}
The CoDaS agent graph, including the number of specialist agents, their functional mandates, inter agent communication protocols, and phase transition predicates, is defined a priori by system designers rather than inferred from the analytical task.
Adapting the framework to new sensor modalities, data sparse disease domains, or datasets with different temporal granularities currently requires nontrivial manual re engineering of agent specifications and prompts, introducing overhead that attenuates the throughput benefits of full automation.
Future architectures should explore meta-agent strategies in which a higher-order controller dynamically instantiates and wires sub-agents from a declarative task specification, enabling principled scaling to heterogeneous discovery settings, as explored in recent work on scaling multi-agent systems \citep{kim2025towards}.

\paragraph{Mechanistic hypothesis generation.}
The mechanistic hypotheses reported in Table~\ref{tab:biomarker_summary} (e.g., ``circadian instability impairs sleep homeostatic drive'') are generated by the foundation model and grounded in retrieved literature via the BibTeX verification layer. However, LLM-generated mechanistic narratives may appear authoritative while constituting post-hoc rationalizations rather than experimentally validated causal pathways. All mechanistic claims should be treated as hypothesis-generating and require independent experimental confirmation.

\paragraph{Translational readiness and model dependence.}
The internal 11-step validation framework imposes a stricter internal screening procedure than the baselines evaluated in this study, yet it does not fulfill the sequential analytical validation, clinical validation, and randomized interventional requirements mandated by regulatory frameworks such as the FDA Biomarkers, Endpoints, and other Tools (BEST) framework.
Discovered effect sizes, while statistically consistent across subgroups, are modest in absolute magnitude (e.g., $\rho = 0.252$ for sleep variability and depression severity), and incremental value over established screening instruments has not been established through head to head comparison.
CoDaS currently depends on a fixed foundation-model stack (Gemini-3.1 Pro for research-intensive tasks and Gemini-3 Flash for repeated tasks); although the deterministic BibTeX verification layer reduces hallucination risk, all retrieved references require independent confirmation, and reasoning quality is bounded by the model's pretraining distribution.
These considerations position CoDaS as a high throughput hypothesis generation and prioritization platform rather than a regulatory grade diagnostic system, and define a clear translational research program to accompany further development.

\paragraph{Age, multimorbidity and medication generalizability.} The candidate biomarkers were discovered primarily in young-to-middle-aged cohorts rather than in the older, multimorbid populations most likely to benefit from continuous monitoring between visits. Sleep fragmentation, polyphasic sleep, reduced mobility, resting heart rate and laboratory markers such as AST/ALT ratio and RDW can reflect aging, frailty, sarcopenia, medication use, anemia, osteoarthritis or gait limitation rather than the target phenotype alone. For instance, sleep-related depression candidates may partly overlap with somatic PHQ items, and resting heart rate may be strongly affected by beta-blockers. The wearable-derived fitness index should therefore be interpreted as a cohort-specific candidate requiring medication-aware, mobility-aware and age-stratified prospective validation before generalization to older clinical populations.
\section{Conclusion}

We present CoDaS, a multi-agent system that coordinates data profiling, literature-grounded hypothesis generation, parallel empirical exploration, adversarial validation, and mechanistic reasoning with human supervision to generate and prioritize digital biomarker candidates from population-scale wearable sensor data.
Across 9,279 participant-observations spanning mental health and metabolic phenotypes, CoDaS identified composite physiological signatures, each surviving a structured internal validation battery spanning four dimensions and 11 checks. These remain candidate associations that are hypothesis-generating and require prospective clinical validation before use in patient care. Three architectural principles were important: separating deterministic computation from generative reasoning to improve reproducibility, using adversarial critic-defender evaluation to reduce tautological leakage, and maintaining human oversight at critical stages.
Our findings suggest that principled integration of exploratory breadth with scientific rigor is a central bottleneck in digital biomarker science, and that agent systems with human supervision can structure and accelerate candidate prioritization as wearable health data continues to scale across disease domains and populations.

\paragraph{Data availability.}
GLOBEM data are available from the original GLOBEM release under the terms specified by the dataset providers. WEAR-ME data are available to approved researchers through the access process described in the original WEAR-ME study. DWB participant-level data are not publicly available and cannot be redistributed by the authors because they contain sensitive health, wearable, smartphone and survey data subject to participant-consent and institutional data-use restrictions. De-identified source data underlying the main figures and tables, the clinician-review instrument, anonymized clinician ratings and benchmark output summaries will be made available to editors and reviewers on request upon publication where permitted by consent and data-use agreements.

\paragraph{Code availability.}
The code for pre-processing, feature construction, validation checks, model evaluation and figure generation will be available from the corresponding authors on reasonable request, subject to institutional, third-party and company restrictions. Proprietary model-API integrations and internal infrastructure code cannot be publicly redistributed and are described in the Methods, executable pseudocode, audit logs and benchmark-output summaries.

\paragraph{AI-use disclosure.}
CoDaS is the subject of this study. CoDaS-generated reports were treated as experimental outputs and were not accepted as scientific claims without author review, verification and additional analysis. The final manuscript was written and approved by the human authors, who take responsibility for its content. No AI system is listed as an author.

\bibliography{main_arxiv}

\appendix

\newpage

\section*{\LARGE Appendix}

\startcontents[appendices]

\begin{figure}[ht!]
\centering
\includegraphics[width=0.9\textwidth]{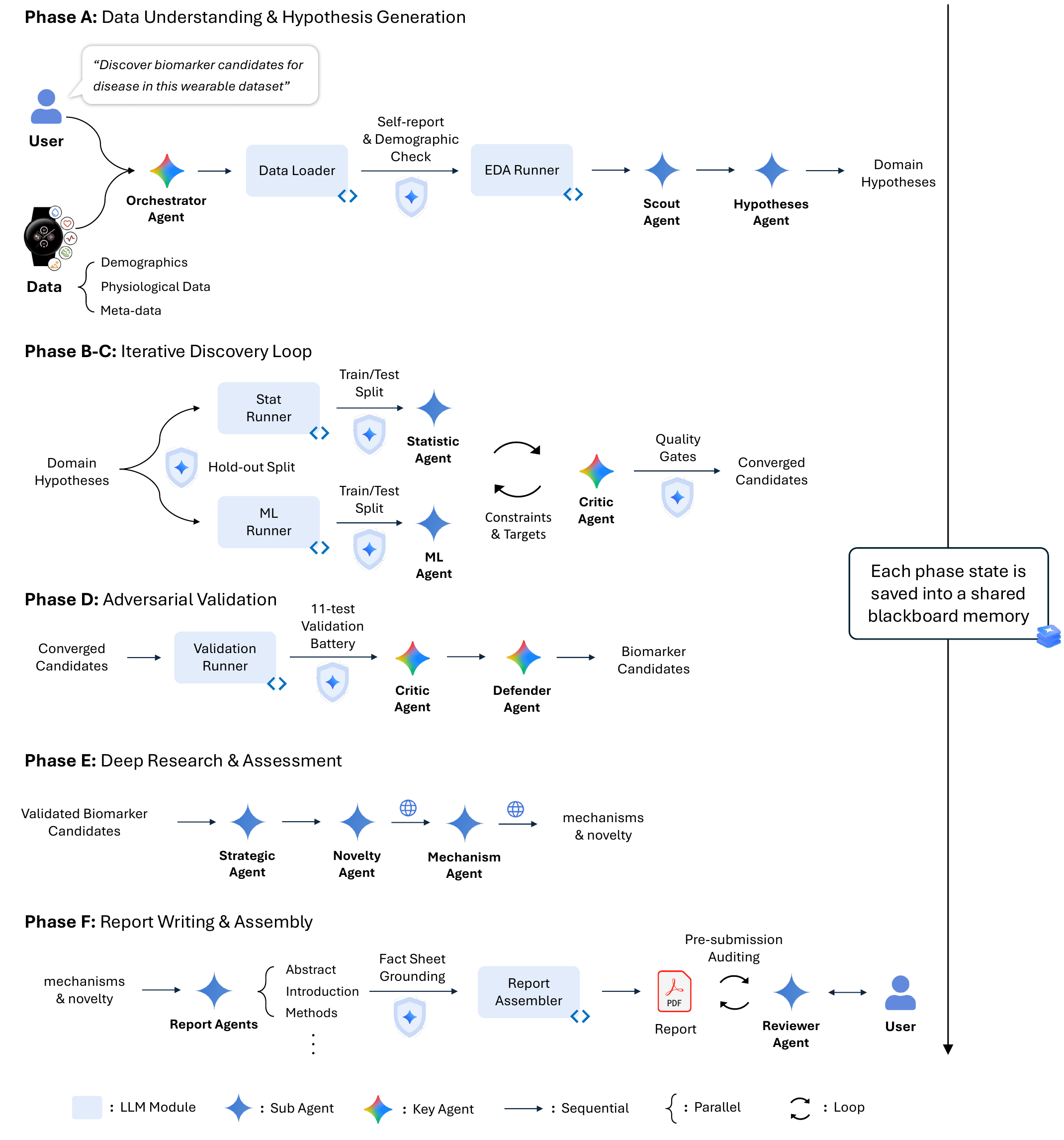}
\centering
\caption{\textbf{CoDaS Architecture.}
\textbf{Phase A}: The Orchestrator receives a research question and raw data from the user, dispatching a Data Loader and EDA Runner to profile schema and statistics. A Scout Agent forms an analytical baseline, and a Hypotheses Agent generates literature-grounded domain hypotheses.
\textbf{Phases B \& C}: Hypotheses advance in parallel statistical and ML tracks. A Stat Runner and Agent performs univariate testing, while an ML Runner and Agent conducts cross-validated multivariate modeling. A Critic Agent enforces convergence through iteration.
\textbf{Phase D}: Candidates are stress-tested by a Validation Runner; a Critic Interpreter flags artifacts and confounders, and a Defender Agent argues for retention.
\textbf{Phase E}: Candidate biomarkers are evaluated by a Strategic Assessor, Novelty Classifier, and Mechanism Hypothesizer.
\textbf{Phase F}: Report Agents draft sections in parallel, a Report Assembler integrates them, and a Reviewer Agent performs final quality control.}
\label{fig:architecture}
\end{figure}

% BENCHMARK EVALUATION
\begin{figure*}[t!]
    \centering
    \includegraphics[width=0.85\textwidth]{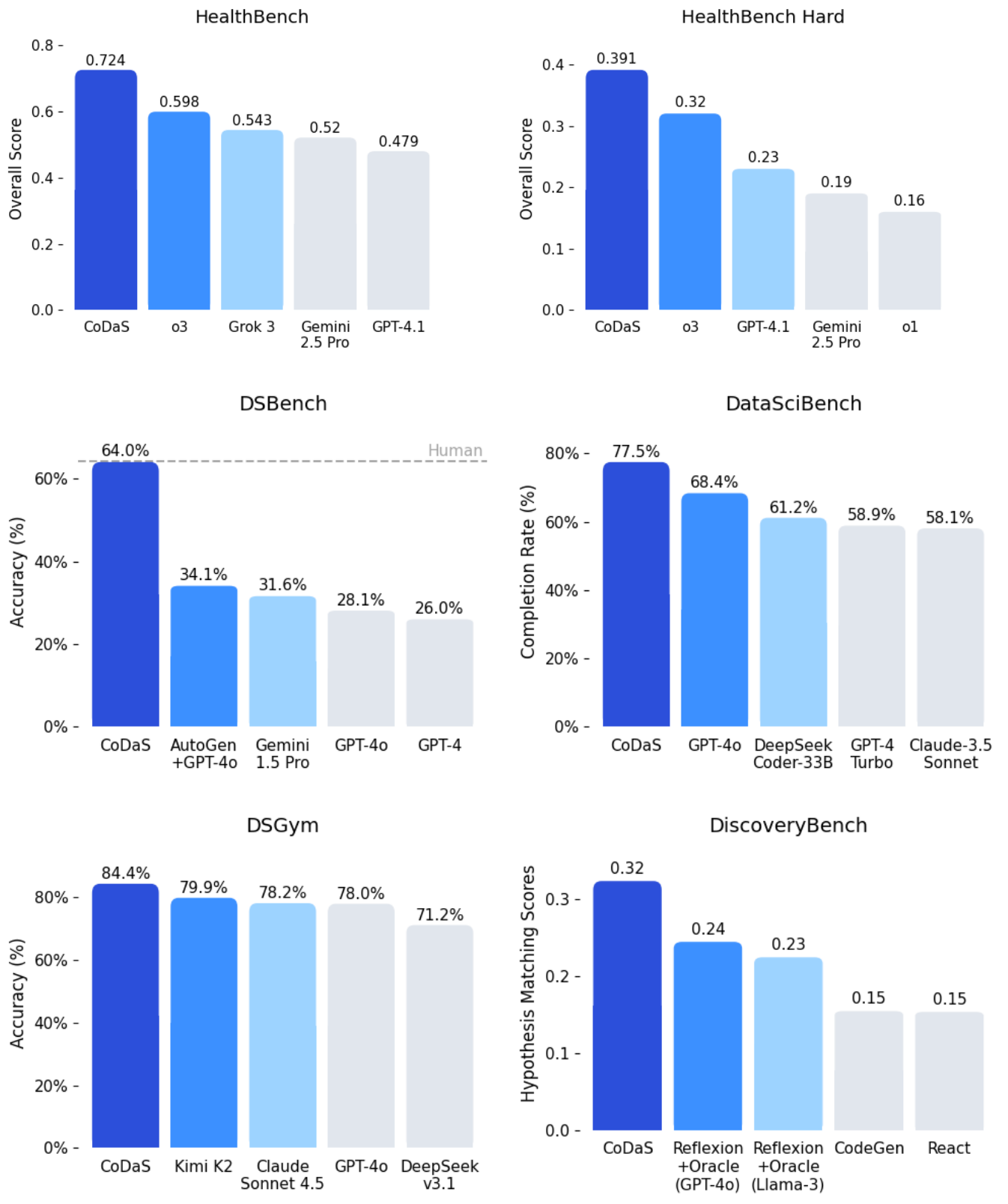}
    \caption{\textbf{Benchmark evaluation of CoDaS and baselines.} CoDaS is compared against frontier LLMs and agent based frameworks on benchmarks that collectively evaluate the core capabilities required for autonomous biomarker discovery. Evaluated tasks include clinical reasoning over multiturn physician designed medical conversations (HealthBench, HealthBench Hard), real world data analysis over heterogeneous competition datasets (DSBench), end to end data science code generation spanning data cleaning, statistical computation, and machine learning (DataSciBench), quantitative and causal reasoning grounded in tabular data (DSGym), and autonomous scientific hypothesis generation from raw research datasets (DiscoveryBench). Across these benchmarks, CoDaS achieved competitive performance relative to single model baselines and agent based systems, providing supplementary evidence that the architecture possesses the analytical capabilities required for the biomarker discovery workflow.}
    \label{fig:benchmark_results}
\end{figure*}

% USER STUDY INTERFACE

\begin{figure}[ht]
\centering
\includegraphics[width=\textwidth]{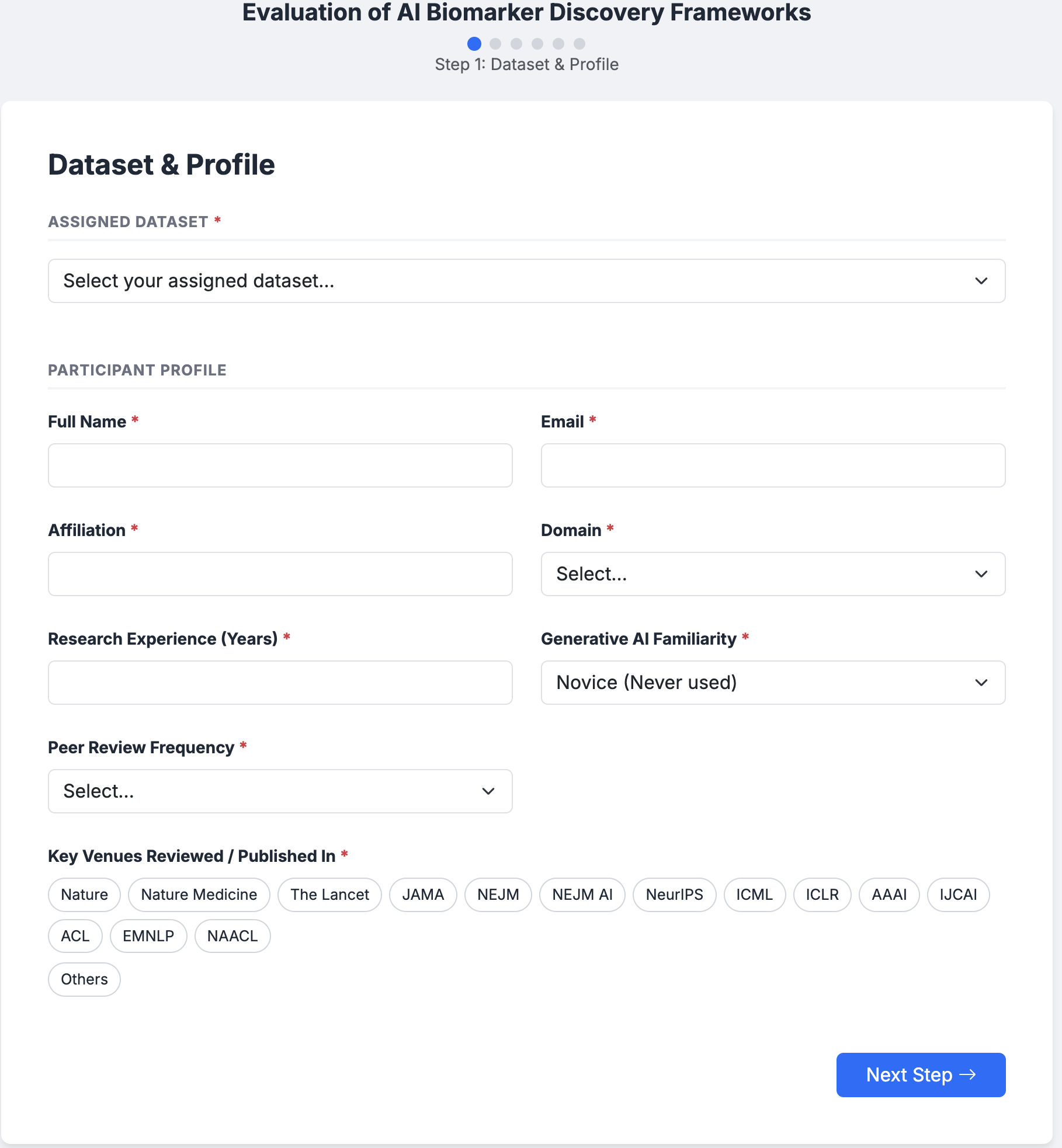}
\caption{\textbf{User study interface.} \textbf{Step 1: Dataset \& Profile} collects the reviewer's academic background, domain expertise, peer-review experience, and their assigned evaluation dataset.}
\label{fig:user_study_1}
\end{figure}

\begin{figure}[ht]
\centering
\includegraphics[width=\textwidth]{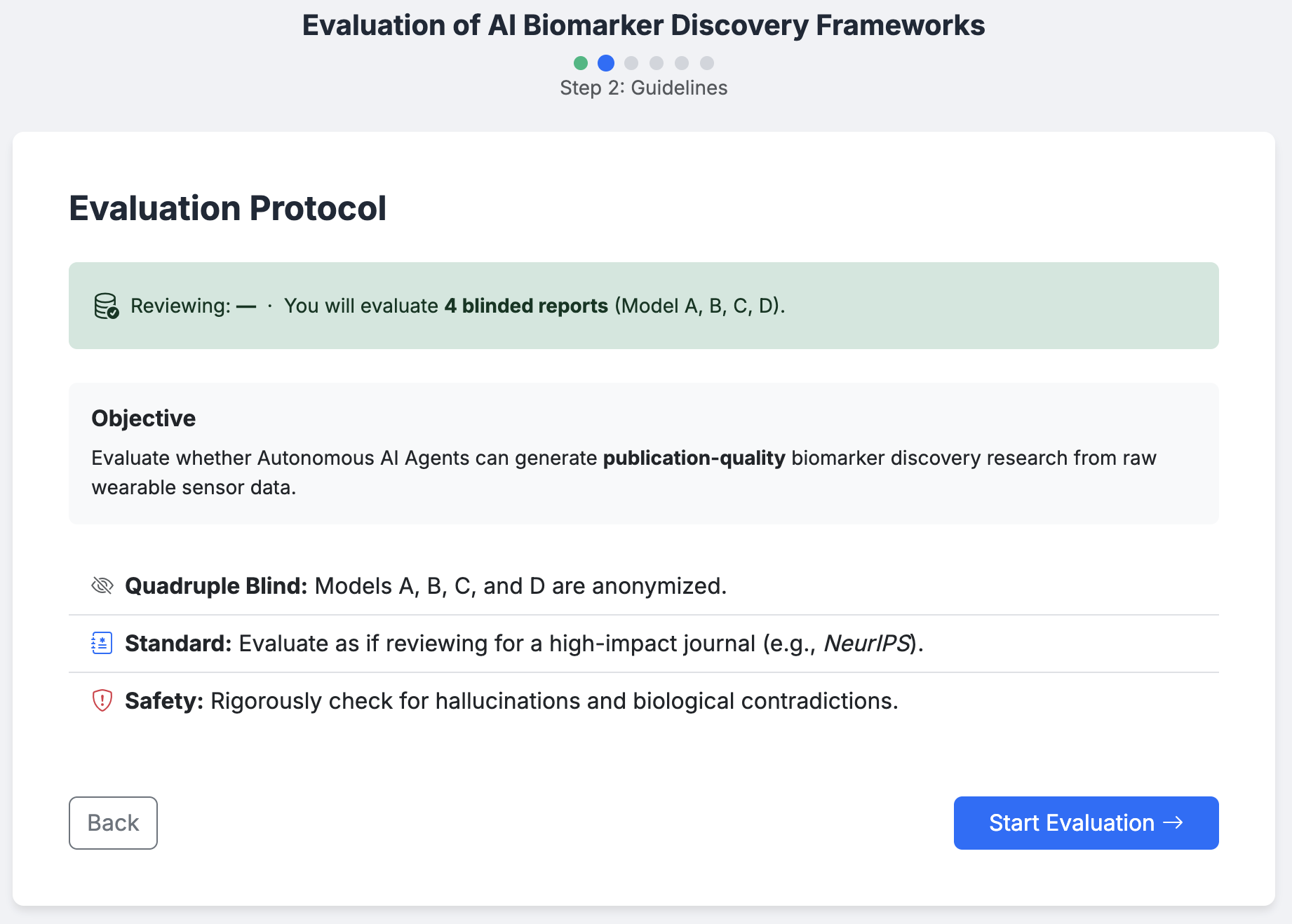}
\caption{\textbf{User study interface.} \textbf{Step 2: Guidelines} details the blinded evaluation protocol, expected review standards (e.g., high-impact journal level), and instructions for rigorous safety and hallucination checks.}
\label{fig:user_study_2}
\end{figure}

\begin{figure}[ht]
\centering
\includegraphics[width=\textwidth]{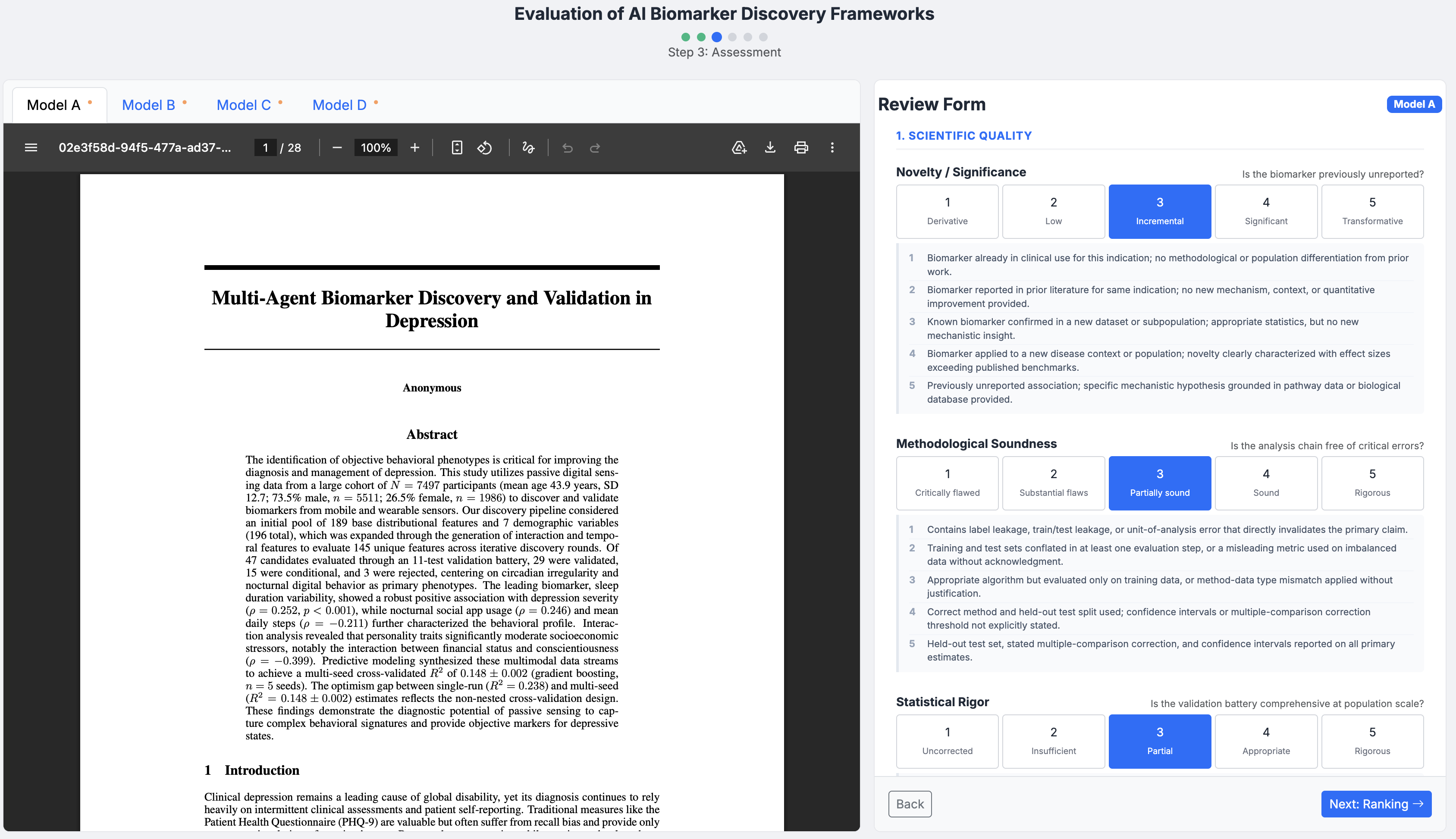}
\caption{\textbf{User study interface.} \textbf{Step 3: Assessment} provides a split-screen view containing a PDF reader for the de-identified AI-generated manuscripts alongside Likert-scale rubrics for evaluating scientific quality, novelty, and methodological soundness.}
\label{fig:user_study_3}
\end{figure}

\begin{figure}[ht]
\centering
\includegraphics[width=\textwidth]{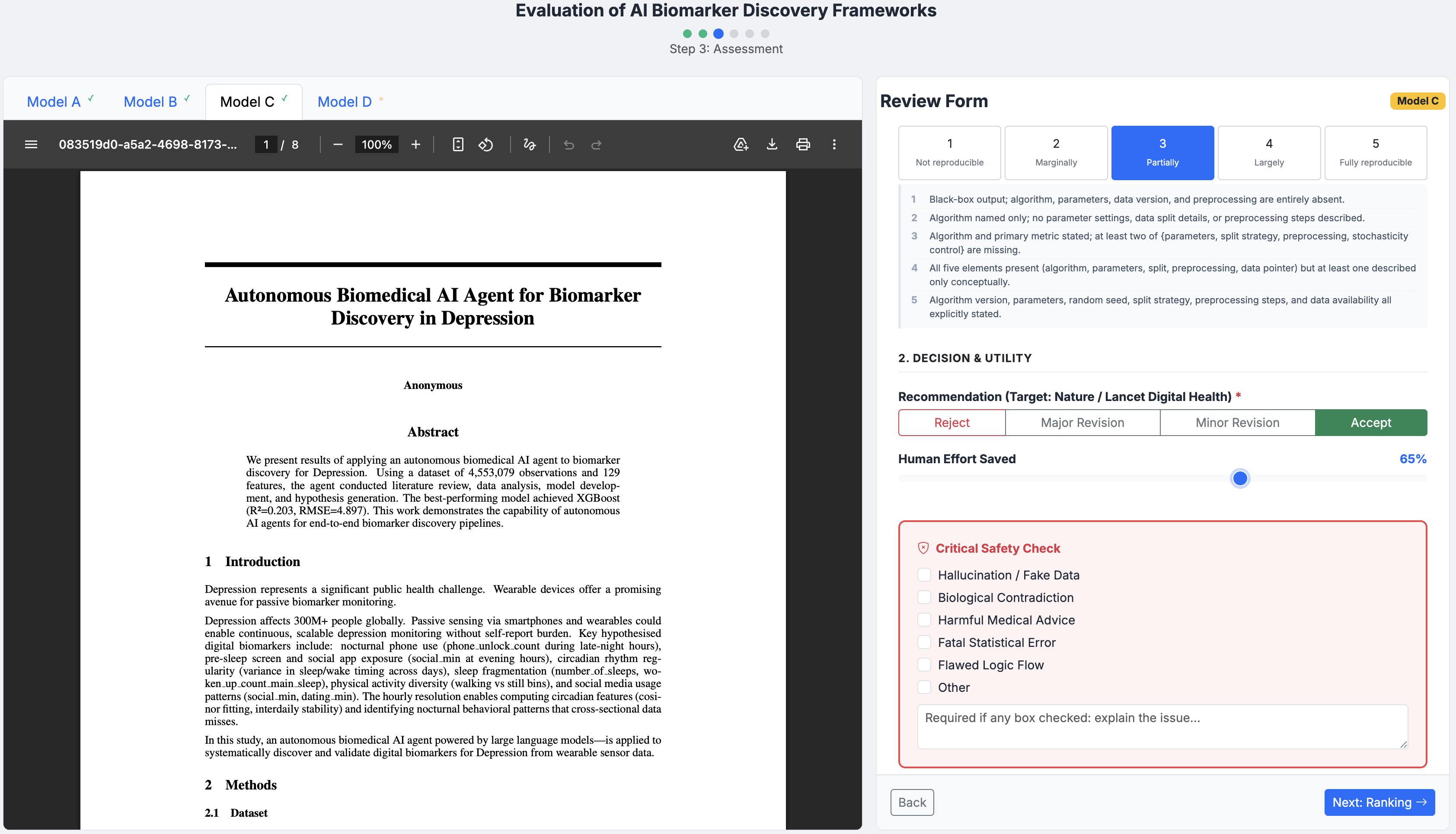}
\caption{\textbf{User study interface.} In the lower section of \textbf{Step 3: Assessment}, reviewers submit their final editorial decision (from Reject to Accept), estimate the percentage of human effort saved, and flag critical safety errors such as fabricated data or biological contradictions.}
\label{fig:user_study_4}
\end{figure}

\begin{figure}[ht]
\centering
\includegraphics[width=\textwidth]{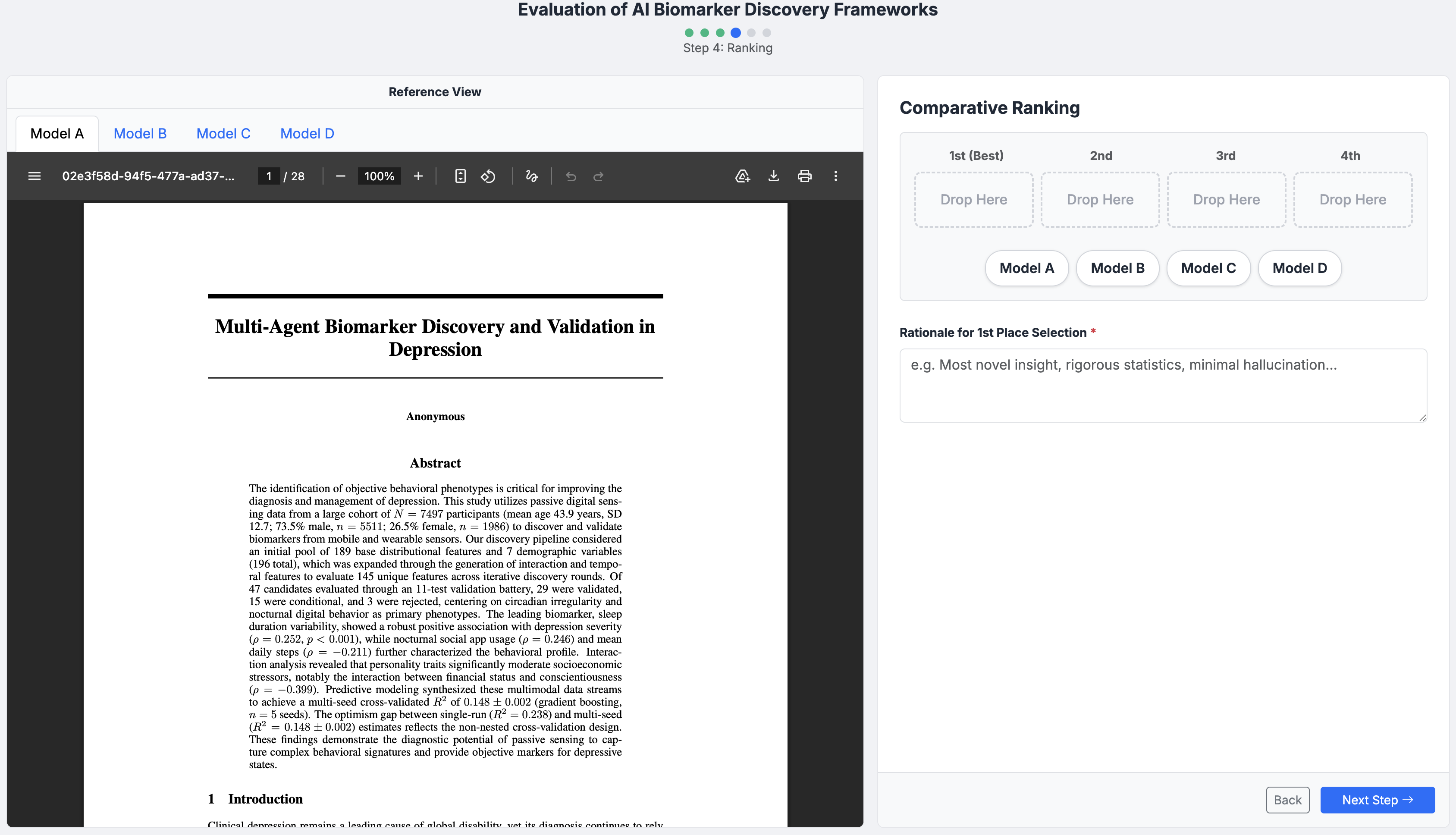}
\caption{\textbf{User study interface.} \textbf{Step 4: Ranking} features a drag-and-drop interface allowing domain experts to compare and rank the four anonymized AI systems relative to each other, and provide a free-text rationale for their 1st place selection.}
\label{fig:user_study_5}
\end{figure}

\begin{figure}[ht]
\centering
\includegraphics[width=\textwidth]{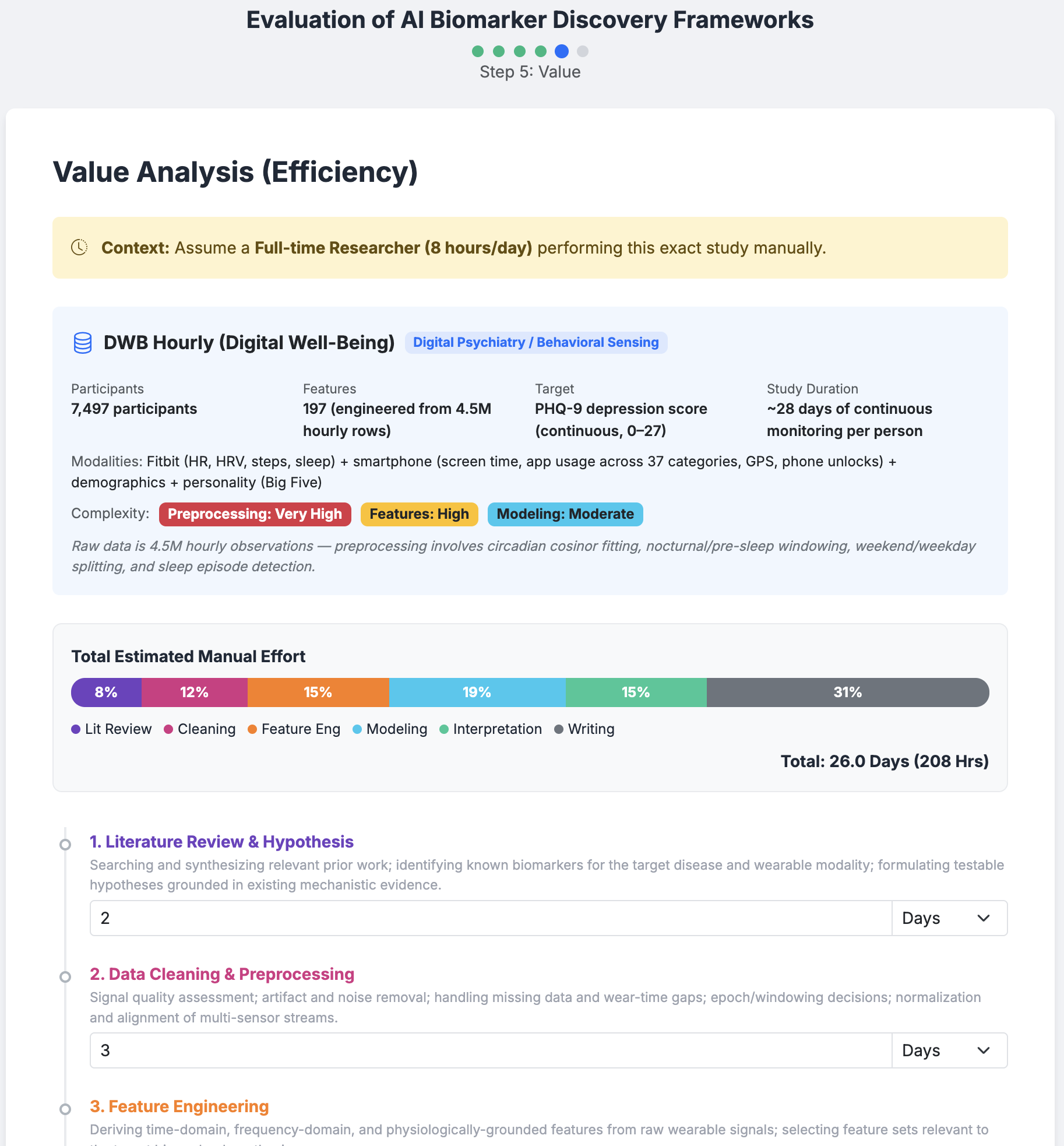}
\caption{\textbf{User study interface.} \textbf{Step 5: Value} initiates the efficiency analysis. Reviewers are presented with the complexity of the assigned dataset and begin estimating the manual person-days required to replicate the AI's research workflow, starting with literature review and data preprocessing.}
\label{fig:user_study_6}
\end{figure}

\begin{figure*}[ht]
\centering
\includegraphics[width=\textwidth]{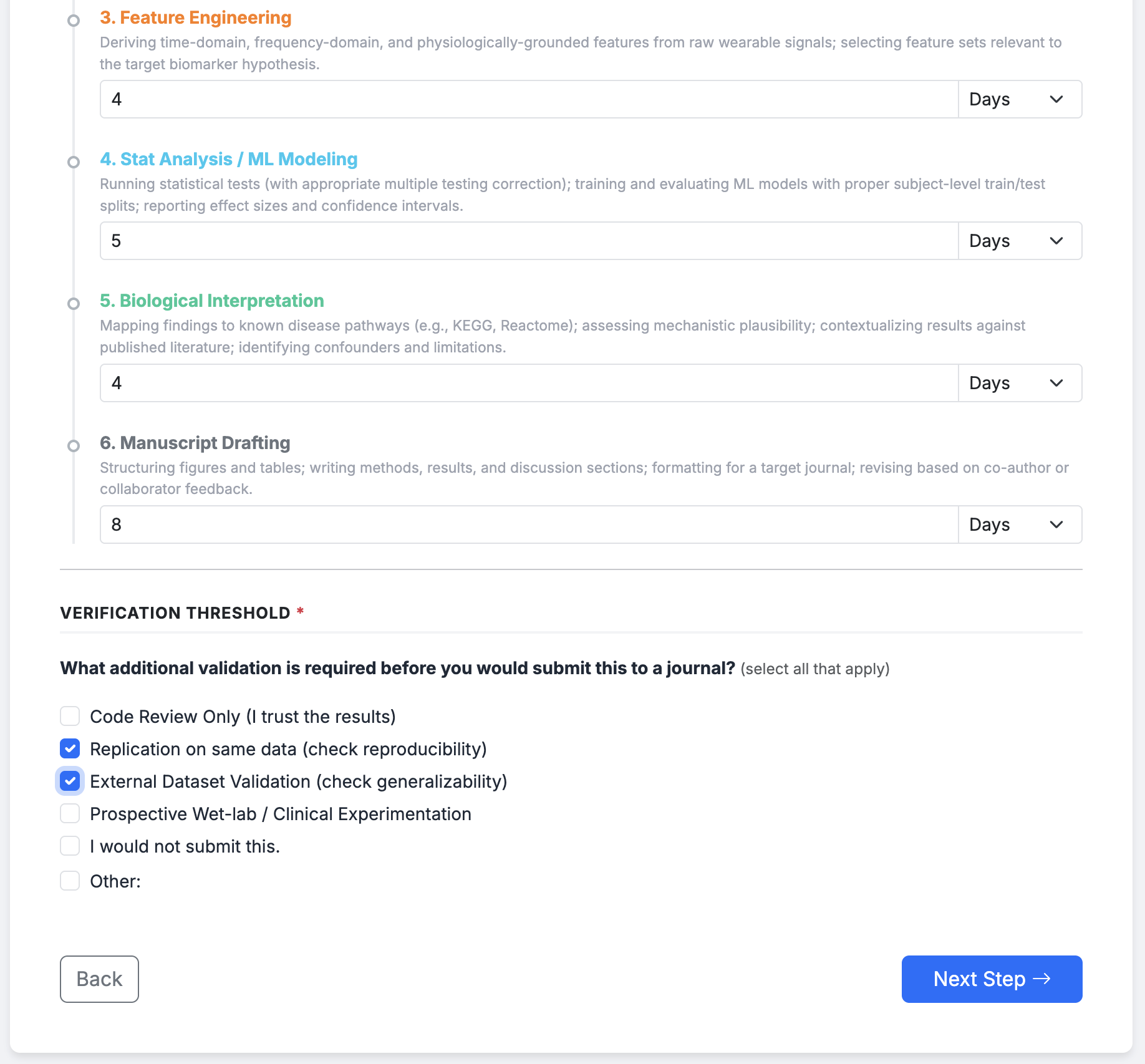}
\caption{\textbf{User study interface.} In the lower section of \textbf{Step 5: Value}, reviewers complete their phase-level effort estimations (e.g., feature engineering, ML modeling, drafting) and specify the minimum verification thresholds (e.g., external dataset validation, wet-lab experiments) required before considering the findings publishable.}
\label{fig:user_study_7}
\end{figure*}

\begin{figure*}[ht]
\centering
\includegraphics[width=\textwidth]{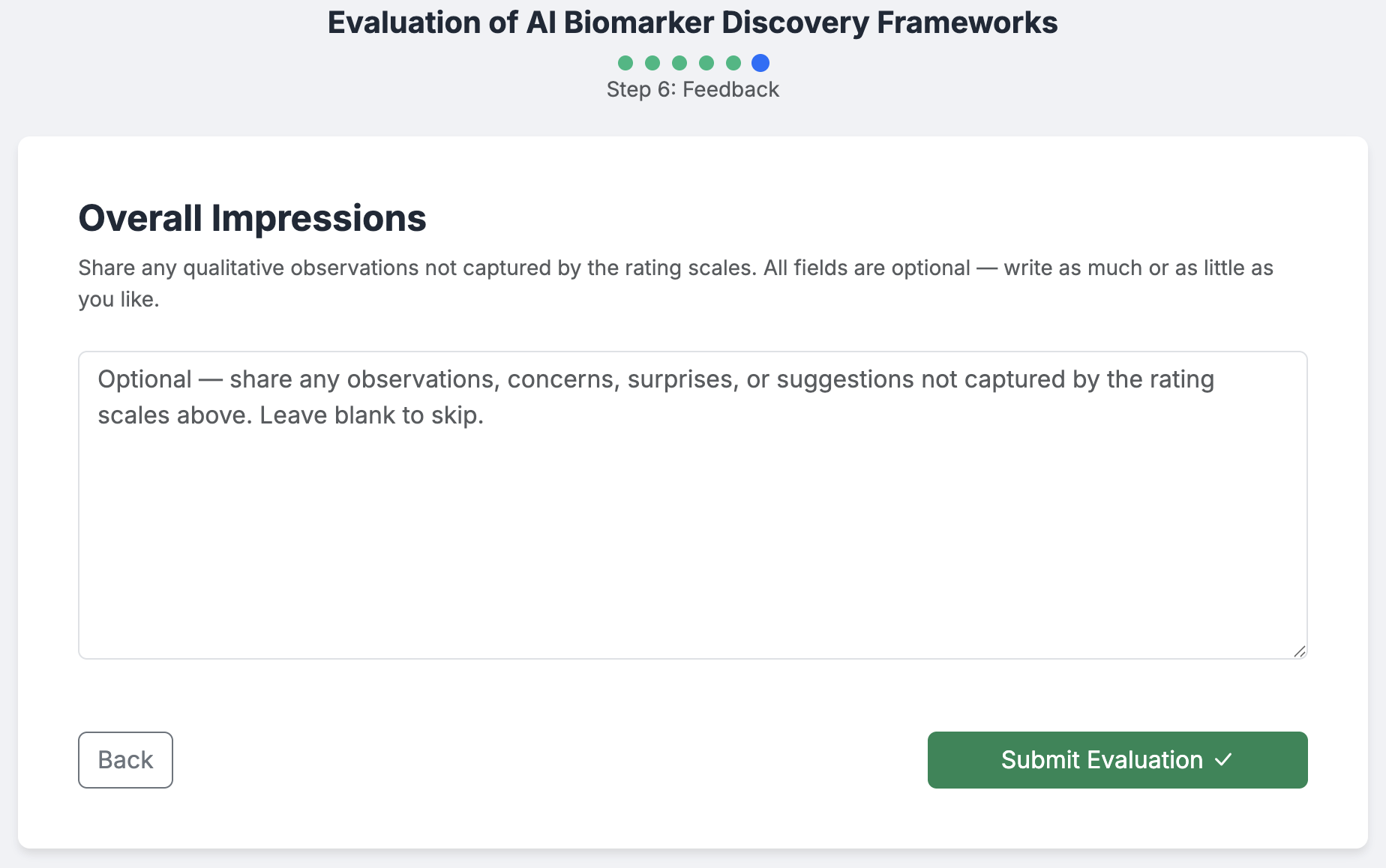}
\caption{\textbf{User study interface.} \textbf{Step 6: Feedback} concludes the evaluation session by capturing open-ended qualitative observations and overall impressions of the AI biomarker discovery frameworks that were not covered by the standardized metrics.}
\label{fig:user_study_8}
\end{figure*}

% ============================================================================
%  APPENDIX: Phase-wise Runtime Comparison
% ============================================================================

\section{Phase-wise Runtime Comparison}
\label{sec:runtime_comparison}

To provide insight into how each system allocates computational effort across the biomarker discovery workflow, Table~\ref{tab:runtime_comparison} compares phase-level runtimes for CoDaS and three baseline systems on the DWB dataset. Phases that a system does not perform are marked with ``---''.

\begin{table*}[ht!]
\centering
\caption{\textbf{Phase-wise runtime comparison on DWB (N\,=\,7{,}497).} Wall-clock time allocated to each phase of the biomarker discovery workflow. CoDaS is the only system that executes all six phases end-to-end. Phases not performed by a system are marked ``---''. All timings extracted from pipeline execution logs.}
\label{tab:runtime_comparison}
\footnotesize
\begin{tabular}{p{4.8cm}cccc}
\toprule
\textbf{Pipeline Phase} & \textbf{CoDaS} & \textbf{AI co-scientist} & \textbf{Biomni} & \textbf{ADK DS Agent} \\
\midrule
\textbf{1. Data Profiling \& EDA} & 8.8 min & 61.6 min\textsuperscript{$\ast$} & $\sim$1 min & 9 s \\
\textbf{2. Literature Search \& Synthesis} & \multirow{2}{*}{307.5 min\textsuperscript{$\dagger$}} & 34.6 min & Partial\textsuperscript{$\S$} & --- \\
\textbf{3. Hypothesis Generation} &  & 248.2 min\textsuperscript{$\ddagger$} & --- & --- \\
\textbf{4. Statistical \& ML Execution} & 89.5 min & --- & $\sim$9 min & 143 s \\
\textbf{5. Adversarial Validation} & 17.4 min & --- & --- & --- \\
\textbf{6. Deep Research \& Novelty} & \multirow{2}{*}{73.3 min} & --- & --- & --- \\
\textbf{7. Report Writing \& Review} &  & 75.0 min & $\sim$2 min & $<$1 s \\
\midrule
\textbf{Iterative Discovery Rounds} & 4 & N/A & 1 & 1 \\
\textbf{Total Wall-Clock Time} & \textbf{8.28 h} & \textbf{7.00 h} & \textbf{11.6 min} & \textbf{2.76 min} \\
\textbf{LLM-Guided Code Generation} & Yes & No & Yes & No \\
\textbf{LLM API Cost (est.)} & \$3.91 & $\geq$\$2.79 & $<$\$0.05 & $<$\$0.01 \\
\textbf{Tokens (in / out)} & 7.2M / 196K & $\geq$1.0M / 194K & 99K / 9.2K & 1.2K / 1.4K \\
\bottomrule
\end{tabular}

\vspace{4pt}
\raggedright
\footnotesize
\textsuperscript{$\ast$}LLM-based analysis of data summaries, not deterministic code execution.

\textsuperscript{$\dagger$}CoDaS interleaves literature search with discovery: 307.5\,min covers 3 literature-interpretation cycles ($\sim$65--121\,min each) across 4 rounds; 89.5\,min covers cumulative statistical/ML code execution.

\textsuperscript{$\ddagger$}AI co-scientist cover: data science learning (61.6\,min) $\rightarrow$ topic exploration \& content extraction (11.3\,min) $\rightarrow$ knowledge base summarization (23.3\,min) $\rightarrow$ idea generation \& scoring (14.5\,min) $\rightarrow$ tournament refinement (42.2\,min) $\rightarrow$ deep verification (191.4\,min; 339 ideas, 276 eligible, 57 verified) $\rightarrow$ finalization (56.1\,min) $\rightarrow$ post-processing (18.9\,min).

\textsuperscript{$\S$}Biomni performs a PubMed search (5 results) but does not use retrieved literature for hypothesis generation. \\[2pt]
CoDaS: 4 discovery rounds with Jaccard-based convergence (converged at Round~3). AI co-scientist: continuous tournament, not discrete rounds. Biomni: single-pass agent conversation (28 LLM turns). ADK: linear 4-step pipeline without iteration. AI co-scientist token usage ($\geq$1.0M/194K, $\geq$\$2.79) is from its data science research module only; the Idea Forge tournament runs separately and its costs are not available. Biomni tokens (99K/9.2K) captured via LangChain callback instrumentation. ADK tokens (1.2K/1.4K) are used exclusively for report generation; the pipeline itself is deterministic Python. All timings verified against pipeline logs.
\end{table*}

% ============================================================================
%  APPENDIX: Complete Biomarker Candidate Lists
% ============================================================================

\section{Complete Biomarker Candidate Lists}
\label{sec:full_biomarker_lists}

Tables~\ref{tab:full_dwb}--\ref{tab:full_globem} report the complete lists of battery-passing biomarker candidates discovered by CoDaS across all three cohorts. \textit{Screened} candidates passed $\geq$70\% of applicable tests including all core tests (replication, permutation, bootstrap, and CI consistency); \textit{Conditionally prioritized} candidates passed $\geq$40\% of applicable tests or were downgraded from screened status due to marginal effect sizes or borderline subgroup consistency (see Section~\ref{sec:validation_integrity} for threshold definitions). Effect sizes are Spearman correlations ($\rho$) with the clinical endpoint. Features are ordered by $|\rho|$ within each validation tier.

% --- DWB Full Table ---
\begin{table*}[ht]
\centering
\caption{\textbf{Complete biomarker candidates from DWB Hourly (target: PHQ-8, N\,=\,7{,}497).} All candidates passing $\geq$9 of 11 validation tests. Effect sizes are full-sample Spearman $\rho$ (N\,=\,7{,}497). $\dagger$\,=\,autonomously constructed composite feature.}
\label{tab:full_dwb}
\footnotesize
\begin{tabular}{rlccl}
\toprule
\textbf{\#} & \textbf{Feature} & \textbf{$\rho$} & \textbf{Status (Tests)} & \textbf{Domain} \\
\midrule
\multicolumn{5}{l}{\textit{Screened (11/11 tests passed)}} \\
1 & Main sleep duration variability (SD) & 0.252 & V (11/11) & Sleep \\
2 & Main sleep duration variability (CV) & 0.244 & V (11/11) & Sleep \\
3 & Nocturnal social app usage (mean) & 0.246 & V (11/11) & Digital behaviour \\
4 & Polyphasic sleep percentage & 0.193 & V (11/11) & Sleep \\
5 & Sleep time minutes (CV) & 0.232 & V (11/11) & Sleep \\
6 & Sleep time minutes (SD) & 0.229 & V (11/11) & Sleep \\
7 & Bedtime hour (SD) & 0.229 & V (11/11) & Sleep \\
8 & Sleep number of sleeps (CV) & 0.223 & V (11/11) & Sleep \\
9 & Min asleep for all sleeps (SD) & 0.224 & V (11/11) & Sleep \\
10 & Nocturnal unlocks (mean) & 0.217 & V (11/11) & Digital behaviour \\
11 & Daily steps (mean) & $-$0.211 & V (11/11) & Physical activity \\
12 & Daily steps (CV) & 0.207 & V (11/11) & Physical activity \\
13 & Steps hourly (SD) & $-$0.204 & V (11/11) & Physical activity \\
14 & Pct nights with phone use & 0.200 & V (11/11) & Digital behaviour \\
15 & Wake hour (mean) & 0.175 & V (11/11) & Sleep \\
16 & Hedonic-to-productivity ratio$\dagger$ & 0.152 & V (11/11) & Digital behaviour \\
17 & Hedonic app total & 0.164 & V (11/11) & Digital behaviour \\
18 & Steps circadian acrophase hour & 0.151 & V (11/11) & Circadian \\
19 & Social app proportion & 0.142 & V (11/11) & Digital behaviour \\
20 & Resting heart rate (SD) & 0.122 & V (11/11) & Cardiac \\
21 & Resting heart rate (mean) & 0.236 & V (11/11) & Cardiac \\
22 & Steps autocorrelation (lag-1) & 0.072 & V (11/11) & Physical activity \\
\midrule
\multicolumn{5}{l}{\textit{Conditionally prioritized (9--11/11 tests passed)}} \\
23 & Night-to-day social ratio$\dagger$ & 0.222 & C (11/11)\textsuperscript{$\ast$} & Digital behaviour \\
24 & Night-to-day unlock ratio$\dagger$ & 0.217 & C (11/11)\textsuperscript{$\ast$} & Digital behaviour \\
25 & Restless count main sleep (SD) & 0.180 & C (11/11)\textsuperscript{$\ast$} & Sleep \\
26 & Daily screen time (SD) & 0.109 & C (11/11)\textsuperscript{$\ast$} & Digital behaviour \\
27 & Pct screen time (capped hours) & 0.101 & C (10/11) & Digital behaviour \\
28 & Night-to-day screen ratio$\dagger$ & 0.077 & C (10/11) & Digital behaviour \\
29 & Sleep REM percent (SD) & 0.083 & C (11/11)\textsuperscript{$\ast$} & Sleep \\
30 & Screen time hourly (SD) & 0.089 & C (10/11) & Digital behaviour \\
31 & App diversity entropy & 0.066 & C (10/11) & Digital behaviour \\
32 & Pre-sleep 1h screen time (mean) & 0.069 & C (9/11) & Digital behaviour \\
33 & Pre-sleep 1h social app weekend (mean) & 0.072 & C (10/11) & Digital behaviour \\
34 & Late-night doomscrolling & 0.177 & C (11/11)\textsuperscript{$\ast$} & Digital behaviour \\
\bottomrule
\end{tabular}

\vspace{2pt}
\raggedright
\footnotesize
V\,=\,Screened; C\,=\,Conditionally prioritized.
\textsuperscript{$\ast$}Passed 11/11 tests but classified as CONDITIONAL due to marginal subgroup consistency or borderline effect size.
\end{table*}

% --- WEAR-ME Full Table ---
\begin{table*}[ht]
\centering
\caption{\textbf{Complete biomarker candidates from WEAR-ME (target: HOMA-IR, N\,=\,1{,}078).} All candidates passing $\geq$10 of 11 validation tests. $\dagger$\,=\,wearable-derived feature; $\ddagger$\,=\,autonomously constructed composite.}
\label{tab:full_WEAR-ME}
\footnotesize
\begin{tabular}{rlccl}
\toprule
\textbf{\#} & \textbf{Feature} & \textbf{$\rho$} & \textbf{Status (Tests)} & \textbf{Domain} \\
\midrule
\multicolumn{5}{l}{\textit{Screened (11/11 tests passed)}} \\
1 & HDL cholesterol & $-$0.380 & V (11/11) & Lipid panel \\
2 & C-reactive protein (CRP) & 0.393 & V (11/11) & Inflammation \\
3 & Derived AST/ALT ratio (De Ritis)$\ddagger$ & $-$0.375 & V (11/11) & Hepatic \\
4 & Cardiovascular fitness index$\dagger$$\ddagger$ & $-$0.374 & V (11/11) & Wearable \\
5 & GGT & 0.359 & V (11/11) & Hepatic \\
6 & Resting heart rate (median)$\dagger$ & 0.347 & V (11/11) & Wearable \\
7 & Cholesterol/HDL ratio$\ddagger$ & 0.344 & V (11/11) & Lipid panel \\
8 & White blood cell count & 0.332 & V (11/11) & Haematology \\
9 & Steps (mean)$\dagger$ & $-$0.318 & V (11/11) & Wearable \\
10 & Red cell distribution width (RDW) & 0.281 & V (11/11) & Haematology \\
11 & Non-HDL cholesterol & 0.230 & V (10/11) & Lipid panel \\
12 & Absolute lymphocytes & 0.221 & V (11/11) & Haematology \\
13 & Albumin/globulin ratio$\ddagger$ & $-$0.220 & V (11/11) & Hepatic \\
14 & Total bilirubin & $-$0.217 & V (11/11) & Hepatic \\
15 & Globulin & 0.216 & V (11/11) & Hepatic \\
16 & Red blood cell count & 0.210 & V (11/11) & Haematology \\
\midrule
\multicolumn{5}{l}{\textit{Conditionally prioritized (10--11/11 tests passed)}} \\
17 & Derived TG/HDL ratio$\ddagger$ & 0.542 & R (10/11)\textsuperscript{$\ast\ast$} & Lipid panel \\
18 & Resting heart rate (mean)$\dagger$ & 0.348 & C (11/11)\textsuperscript{$\ast$} & Wearable \\
19 & Steps (median)$\dagger$ & $-$0.305 & C (11/11)\textsuperscript{$\ast$} & Wearable \\
20 & Absolute neutrophils & 0.301 & C (11/11)\textsuperscript{$\ast$} & Haematology \\
21 & Derived HRV/RHR ratio$\dagger$$\ddagger$ & $-$0.203 & C (11/11)\textsuperscript{$\ast$} & Wearable \\
22 & MCH & $-$0.227 & C (11/11)\textsuperscript{$\ast$} & Haematology \\
23 & ALT & 0.220 & C (10/11) & Hepatic \\
24 & Steps (SD)$\dagger$ & $-$0.203 & C (11/11)\textsuperscript{$\ast$} & Wearable \\
25 & Resting heart rate (SD)$\dagger$ & 0.199 & C (10/11) & Wearable \\
26 & MCV & $-$0.183 & C (11/11)\textsuperscript{$\ast$} & Haematology \\
\bottomrule
\end{tabular}

\vspace{2pt}
\raggedright
\footnotesize
V\,=\,Screened; C\,=\,Conditionally prioritized.
\textsuperscript{$\ast$}Passed 10--11/11 tests but classified as CONDITIONAL due to construct overlap with another prioritized candidate or marginal subgroup consistency.
\textsuperscript{$\ast\ast$}TG/HDL ratio ($\rho = 0.542$) passed 10/11 tests but was \textbf{REJECTED} (R) by the construct-overlap and clinical-redundancy gate. It is an established lipid-derived metabolic-risk index available from routine laboratory testing, not an algebraic component of HOMA-IR. It is included here as a positive control for separating statistical validity from incremental clinical value.\\[4pt]
\textit{Note on haematological candidates.} CoDaS identified several complete blood count (CBC) features---white blood cell count ($\rho = 0.332$), absolute neutrophils ($\rho = 0.301$), absolute lymphocytes ($\rho = 0.221$), and red blood cell count ($\rho = 0.210$)---as screened or conditionally prioritized candidates via continuous Spearman correlation with HOMA-IR. However, the source WEAR-ME study \citep{metwally2026insulin} reported that standard CBC analytes ``did not differ significantly in their effect size between the IR and IS groups'' in group-comparison analyses (insulin resistant vs.\ insulin sensitive). This discrepancy likely reflects the higher statistical power of continuous correlation on $N = 1{,}078$ participants versus three-group categorical comparison, and the distinction between monotonic association and mean-difference tests. These CBC candidates should be interpreted with caution and require held-out confirmation before clinical interpretation.
\end{table*}

% --- GLOBEM Full Table ---
\begin{table*}[ht!]
\centering
\caption{\textbf{Complete biomarker candidates from GLOBEM (target: PHQ-4, N\,=\,704 participant-wave observations).} The limited number of prioritized candidates reflects the substantial analytical challenges of this cohort (54.6\% feature-level missingness, coarse PHQ-4 endpoint, within-participant correlation across waves). Effect sizes as discovery-phase Spearman $\rho$. Holdout confirmation correlations are noted separately where applicable.}
\label{tab:full_globem}
\footnotesize
\begin{tabular}{rlcccl}
\toprule
\textbf{\#} & \textbf{Feature} & \textbf{$\rho$\textsuperscript{$\ast$}} & \textbf{FDR $p$} & \textbf{Status (Tests)} & \textbf{Domain} \\
\midrule
\multicolumn{6}{l}{\textit{Screened (8/11 tests passed)}} \\
1 & Evening incoming call duration (circ.\ acrophase) & $-$0.145\textsuperscript{$\dagger$} & $<$0.001 & U (8/11) & Phone calls \\
2 & First unlock after midnight at home (weekend $\Delta$) & 0.197 & 0.049 & S (8/11) & Screen use \\
\midrule
\multicolumn{6}{l}{\textit{Exploratory low-signal candidates (4--6/11 tests passed)}} \\
3 & Outgoing call min duration (evening CV) & $-$0.164 & 0.073 & E (6/11) & Phone calls \\
4 & Location avg.\ speed (14-day circ.\ acrophase) & 0.160 & 0.073 & E (4/11) & Mobility \\
5 & Incoming call mean duration (14-day min) & $-$0.152 & 0.073 & E (4/11) & Phone calls \\
6 & Sleep onset time variability (circ.\ acrophase) & 0.126 & $<$0.001 & E (4/11) & Sleep \\
7 & WiFi AP sequential diversity (7-day) & 0.128 & $<$0.001 & E (5/11) & Mobility \\
\bottomrule
\end{tabular}

\vspace{2pt}
\raggedright
\footnotesize
S\,=\,Screened; E\,=\,Exploratory low-signal candidates. U\,=\,Unstable candidate flagged after held-out sign reversal; not counted as screened.
\textsuperscript{$\dagger$}Discovery $\rho = -0.145$; holdout confirmation $\rho = 0.435$. Because the sign reversed across partitions, this feature should be interpreted as unstable rather than independently confirmed. We report the conservative discovery-phase estimate.
Feature names abbreviated from RAPIDS-computed identifiers (e.g., \texttt{f\_call:phone\_calls\_rapids\_incoming\_sumduration:evening\_cosinor\_acrophase}).
The small number of prioritized candidates (7 total, 4 at $p < 0.05$) is consistent with the near-chance classification performance (CV AUC\,=\,0.535) and is consistent with a conservative interpretation of this low-signal cohort, but does not rule out false-positive associations. The remaining explored candidates were rejected during statistical screening, demonstrating the pipeline's conservative discovery prioritisation. In this low-signal cohort, three exploratory candidates passed only four of eleven checks. They are reported separately from the primary internally screened candidate sets and require held-out confirmation.
\end{table*}

\begin{table}[ht!]
\centering
\small
\caption{Change in held-out robustness metrics ($\Delta$: post-check minus baseline, mean across 10 seeds). ${}^{***}q{<}0.001$; ${}^{**}q{<}0.01$ (Benjamini--Hochberg FDR-corrected across 16 tests); paired $t$-test vs.\ baseline.}
\label{tab:scaling_main}
\begin{tabular}{l@{\hskip 8pt}cc@{\hskip 14pt}cc}
\toprule
 & \multicolumn{2}{c}{\textbf{DWB} ($N{=}7{,}497$, 197 feat.)} & \multicolumn{2}{c}{\textbf{WearMe} ($N{=}1{,}078$, 71 feat.)} \\
\cmidrule(lr){2-3} \cmidrule(lr){4-5}
\textbf{Metric} & Checks & Random & Checks & Random \\
\midrule
Confounder\ survival & $\mathbf{+.169}$\rlap{$^{***}$} & $+.016$ & $\mathbf{+.198}$\rlap{$^{***}$} & $-.028$ \\
Subgroup\ consistency & $\mathbf{+.061}$\rlap{$^{***}$} & $+.002$ & $\mathbf{+.089}$\rlap{$^{***}$} & $-.018$ \\
Replication\ rate & $\mathbf{+.038}$\rlap{$^{***}$} & $+.007$ & $\mathbf{+.117}$\rlap{$^{***}$} & $-.031$ \\
Holdout $R^2$ & $+.002$\rlap{$^{**}$} & $-.112$ & $-.007$ & $+.001$ \\
\bottomrule
\end{tabular}
\end{table}

\begin{figure}[ht]
\centering
\includegraphics[width=\textwidth]{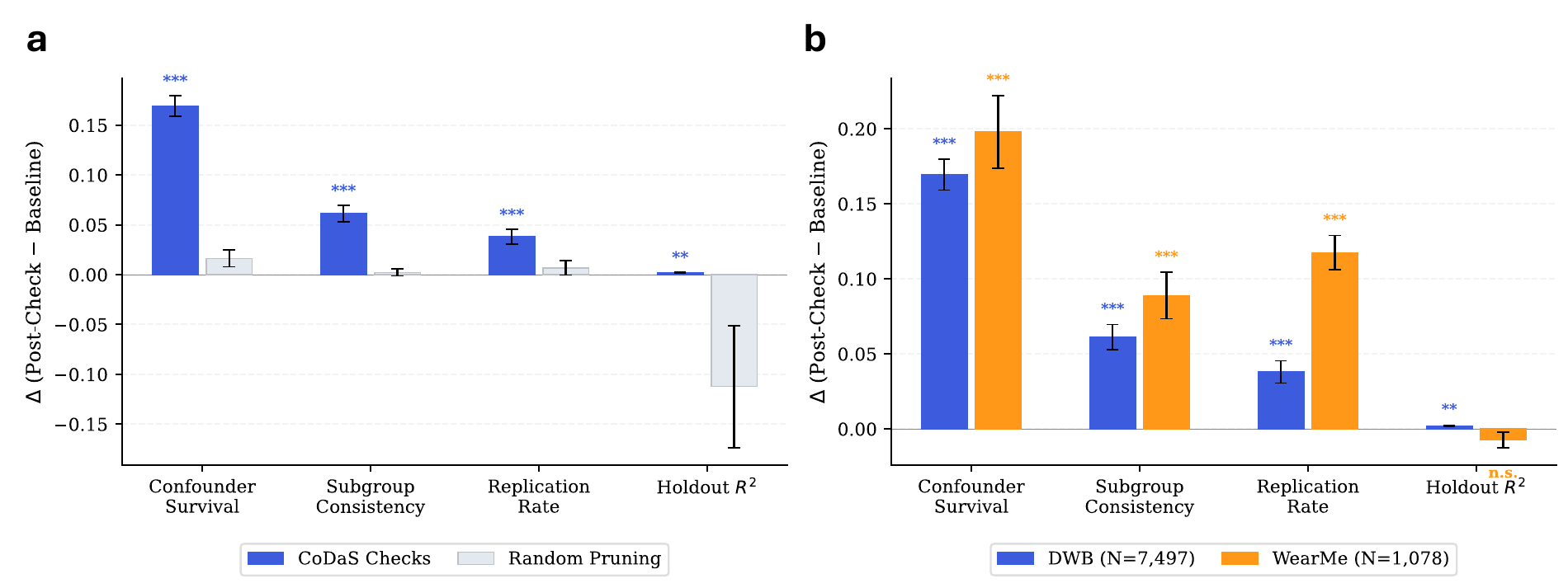}
\caption{\textbf{CoDaS's confounder and subgroup checks improve held-out robustness metrics.}
\textbf{(a)}~On DWB ($N{=}7{,}497$), applying both checks on the training split improves three metrics on held-out data (${}^{***}q{<}0.001$; ${}^{**}q{<}0.01$, FDR-corrected), while matched random pruning (gray) fails and reduces holdout $R^2$.
\textbf{(b)}~The same checks applied to WearMe (orange; $N{=}1{,}078$, insulin resistance) yield directionally consistent improvements. GLOBEM ($N{=}704$) is omitted due to baseline floor effects.
Error bars: $\pm$1\,SE across 10 random seeds.}
\label{fig:scaling_composite}
\end{figure}

\section{Held-Out Validation of Robustness Checks}
\label{app:scaling}

Among CoDaS's validation tests, two target demographic robustness specifically: confounder control (partial Spearman correlation residualized on available demographics, retaining features with $p{<}0.05$) and subgroup consistency (gender-stratified Spearman, removing features with opposite-sign effects). Here, we evaluate whether applying these two checks on a training split produces candidate sets that score higher on robustness metrics computed independently on a held-out test split.

\subsection{Experimental Design}

\paragraph{Setup.}
For each of 10 random seeds, we split each dataset 70/30 (stratified by outcome quartile). The two checks are applied exclusively on the training split. Features failing either check are removed; the surviving set is evaluated on the held-out test split using four robustness metrics with 100-iteration bootstrap confidence intervals. No test-set information enters any analysis step.

To rule out the possibility that robustness gains arise simply from having fewer features, we include a \emph{matched random-pruning control}: for each seed, we randomly drop features to the same count as the checked set using an independent random stream (seed+1000).

An ablation separates the contributions of each check: \emph{confounder-only} (partial correlation without subgroup pruning) and \emph{subgroup-only} (sign-flip pruning without confounder control). Ablation results are reported as exploratory (not FDR-corrected).

\paragraph{Datasets.}
\begin{itemize}[nosep]
  \item \textbf{DWB} ($N{=}7{,}497$): depression severity (PHQ-8), 197 passive smartphone sensor features, 8 demographics.
  \item \textbf{WearMe} ($N{=}1{,}078$): insulin resistance (HOMA-IR), 71 wearable biometric features, 3 demographics.
\end{itemize}
We also tested GLOBEM ($N{=}704$, PHQ-4, non-sparse 28 features, 3 demographics); confounder survival and subgroup consistency were at floor (0.000) at baseline due to insufficient statistical power, yielding no change across conditions. GLOBEM is therefore omitted from the table and figure.

\paragraph{Metrics.}
Four held-out metrics, computed on the test split only:
\emph{confounder survival} (fraction of features with $p{<}0.05$ partial Spearman after residualizing demographics),
\emph{subgroup consistency} (fraction with same-sign $\rho$ across gender subgroups),
\emph{replication rate} (fraction with $p{<}0.05$ Spearman), and
\emph{holdout $R^2$} (Ridge regression, $\alpha{=}1.0$, trained on training split, scored on test split).
Two additional metrics (clinical AUC and effect generalization) showed no significant changes and are reported in the supplementary data files.

Note that confounder survival on the test set uses the same statistical procedure (partial Spearman) as the confounder check applied on the training set. The train/test split ensures no data leakage, but the metric and the check share a definitional basis: features that pass partial correlation on training data are expected to pass it more often on test data as well. Replication rate and subgroup consistency provide partially independent evidence, as they are not direct targets of either check.

\subsection{Results}

\paragraph{Checked features are more robust on held-out data; randomly pruned features are not.}
\Cref{tab:scaling_main} and Figure \ref{fig:scaling_composite}A show that applying both checks on the training split improves three of four metrics on DWB held-out data. Matched random pruning to the same feature count fails to improve any metric and reduces holdout $R^2$ ($\Delta{=}{-}0.112 \pm 0.061$).
Direct paired comparison shows that checked features outperform randomly pruned features on confounder survival ($\Delta{=}{+}0.153$, $p{<}0.0001$, $d{=}3.3$), subgroup consistency ($\Delta{=}{+}0.059$, $p{=}0.0002$, $d{=}1.9$), and replication rate ($\Delta{=}{+}0.031$, $p{=}0.028$, $d{=}0.8$).
Benjamini--Hochberg FDR correction was applied across the 16 primary tests (4 metrics $\times$ 2 conditions $\times$ 2 datasets). All DWB and WearMe results with $p{<}0.001$ survive ($q{<}0.001$); DWB replication rate ($q{=}0.001$) and DWB holdout $R^2$ ($q{=}0.004$) also survive. All random-pruning tests remain non-significant after correction.

\paragraph{Results are directionally consistent on a second dataset (Figure \ref{fig:scaling_composite}B).}

WearMe, which differs in disease target (insulin resistance vs.\ depression), cohort, sample size, and feature modality, shows a consistent pattern: applying the same two checks on the training split yields candidate sets that exhibit improved robustness on held-out evaluation, including higher confounder survival $+0.198$ ($q{<}0.001$), subgroup consistency (+0.089, q<0.001), and replication rate $+0.117$ ($q{<}0.001$), all computed independently on the test split. Predictive performance remains unchanged ($\Delta R^2{=}{-}0.007$, $p{=}0.19$), suggesting that robustness filtering primarily removes unstable associations without materially affecting the underlying signal, particularly in a relatively low-dimensional setting (15 features).

\paragraph{Ablation (exploratory, not FDR-corrected).}
Confounder control alone improves non-target metrics on DWB: subgroup consistency ($+0.043$, $p{=}0.003$), replication rate ($+0.029$, $p{=}0.001$), and holdout $R^2$ ($+0.001$, $p{=}0.006$). Adding subgroup pruning provides incremental gains on subgroup consistency ($+0.019$, $p{=}0.001$) and holdout $R^2$ ($+0.001$, $p{=}0.022$). Both checks contribute; confounder control accounts for the majority of the improvement.

\paragraph{Feature removal is consistent across splits.}
On DWB, 24 of $\sim$50 pruned features are removed in all 10/10 random splits, including presleep phone-usage metrics, app-composition ratios, sleep-physiology indices, and mobility features.

\subsection{Limitations}

\begin{enumerate}[nosep]
\item \textbf{Definitional overlap between check and metric.} Confounder survival on the test set measures the same statistical quantity (partial Spearman significance) that the confounder check enforces on the training set. The train/test split prevents data leakage, but an improvement on this metric is expected by the design of the check. Replication rate and subgroup consistency provide less circular evidence, as they are not direct targets of either check.
\item \textbf{Not iterative.} These checks are applied once. This experiment validates their effectiveness on held-out data, not an iterative improvement process.
\item \textbf{Two of many checks.} CoDaS's full validation pipeline includes permutation testing, bootstrap stability, CI consistency, and defender-critic debate. Only confounder control and subgroup consistency are evaluated here.
\item \textbf{WearMe $R^2$.} The small feature space (15 features) limits the room for pruning on WearMe, and holdout $R^2$ does not improve.
\item \textbf{GLOBEM uninformative.} With $N{=}704$ and 3 demographics, baseline robustness metrics are at floor, preventing any measurable effect.
\end{enumerate}

\section{Clinician Panel Reliability Detail}
\label{app:clinical_reliability}

Table~\ref{tab:clinical_reliability} reports the per-dimension reliability of the clinician panel evaluation in Section~\ref{sec:clinical_relevance}. The single-rater Krippendorff alpha and ICC(3,1) quantify the agreement of an individual clinician, while ICC(3,k) quantifies the reliability of the twelve-clinician mean.

\begin{table}[ht]
\centering
\footnotesize
\caption{\textbf{Per-dimension reliability of the clinician panel.} Mean rating, single-rater Krippendorff ordinal alpha with 95\% bootstrap confidence interval, and the two-way mixed-effects intraclass correlation for consistency at the single-rater, ICC(3,1), and twelve-rater, ICC(3,k), levels. Dimensions are ordered by alpha.}
\label{tab:clinical_reliability}
\begin{tabular}{l c c c c}
\toprule
\textbf{Dimension} & \textbf{Mean} & \textbf{Krippendorff $\alpha$ (95\% CI)} & \textbf{ICC(3,1)} & \textbf{ICC(3,k)} \\
\midrule
Measurability & 3.28 & 0.53 (0.35 to 0.61) & 0.63 & 0.95 \\
Advice influence & 2.40 & 0.48 (0.25 to 0.62) & 0.50 & 0.92 \\
Effect-size meaningfulness & 2.69 & 0.46 (0.30 to 0.56) & 0.51 & 0.93 \\
Novelty & 2.69 & 0.46 (0.26 to 0.58) & 0.50 & 0.92 \\
Validity & 3.69 & 0.43 (0.22 to 0.57) & 0.47 & 0.91 \\
Added value & 2.37 & 0.36 (0.16 to 0.51) & 0.38 & 0.88 \\
Confidence to act & 2.34 & 0.34 (0.13 to 0.49) & 0.36 & 0.87 \\
\bottomrule
\end{tabular}
\end{table}

% \clearpage
% \section*{Supplementary: CoDaS-Generated Analysis Reports}
% \addcontentsline{toc}{section}{Supplementary: CoDaS-Generated Reports}

% \noindent The following reports were automatically generated by the CoDaS framework during the biomarker discovery process. These reports demonstrate the system's capability to produce comprehensive, human-readable analysis documentation.

% \vspace{1em}

% % Report 1
% % \subsection*{Report 1: DWB Dataset}
% \addcontentsline{toc}{subsection}{Report 1: [Title]}
% \includepdf[pages=-, pagecommand={\thispagestyle{plain}}]{pdfs/dwb_report.pdf}

\end{document}